\documentclass[10pt,twocolumn,letterpaper]{article}

\usepackage[pagenumbers]{cvpr} 

\usepackage{graphicx}
\usepackage{amsmath}
\usepackage{amssymb}
\usepackage{booktabs}
\usepackage[table]{xcolor}
\usepackage{comment}
\usepackage{enumitem}

\usepackage[pagebackref,breaklinks,colorlinks]{hyperref}

\usepackage[capitalize]{cleveref}
\crefname{section}{Sec.}{Secs.}
\Crefname{section}{Section}{Sections}
\Crefname{table}{Table}{Tables}
\crefname{table}{Tab.}{Tabs.}

\usepackage[linesnumbered,ruled,vlined,nofillcomment]{algorithm2e}
\SetKwInput{KwInput}{Input}
\SetKwInput{KwOutput}{Output}
\SetKwInput{KwInit}{Init}
\SetKwFor{Find}{find}{do}{}
\SetKwFunction{KwShap}{Shapely}
\SetKwProg{Fn}{Function}{:}{\KwRet}

\def\cvprPaperID{7668} 
\def\confName{CVPR}
\def\confYear{2022}

\begin{document}

\title{HINT: Hierarchical Neuron Concept Explainer}

\author{Andong Wang, Wei-Ning Lee, Xiaojuan Qi\\
The University of Hong Kong\\
{\tt\small wangad@connect.hku.hk, \tt\small wnlee@eee.hku.hk, \tt\small xjqi@eee.hku.hk}
}
\maketitle

\begin{abstract}


To interpret deep networks, one main approach is to associate neurons with human-understandable concepts.
However, existing methods often ignore the inherent relationships of different concepts (\eg, dog and cat both belong to animals), and thus lose the chance to explain neurons responsible for higher-level concepts (\eg, animal).  
In this paper, we study hierarchical concepts inspired by the hierarchical cognition process of human beings.
To this end, we propose HIerarchical Neuron concepT explainer (\textbf{HINT}) to effectively build bidirectional associations between neurons and hierarchical concepts in a low-cost and scalable manner.
HINT enables us to systematically and quantitatively study whether and how the implicit hierarchical relationships of concepts are embedded into neurons, such as identifying collaborative neurons responsible to one concept and multimodal neurons for different concepts, at different semantic levels from concrete concepts (\eg, dog) to more abstract ones (\eg, animal).
Finally, we verify the faithfulness of the associations using Weakly Supervised Object Localization, and demonstrate its applicability in various tasks such as discovering saliency regions and explaining adversarial attacks.
Code is available on \url{https://github.com/AntonotnaWang/HINT}.


\end{abstract}


\section{Introduction}
\label{sec:intro}


Deep neural networks have attained remarkable success in many computer vision and machine learning tasks.
However, it is still challenging to interpret the hidden neurons in a human-understandable manner which is of great significance in uncovering the reasoning process of deep networks and increasing the trustworthiness of deep learning to humans \cite{lipton2016mythos, weld2019challenge, ahmad2018interpretable}.

Early research focuses on finding evidence from input data to explain deep model predictions
\cite{zeiler2014visualizing, lundberg2017unified, lundberg2018consistent, simonyan2013deep, shrikumar2016not, springenberg2014striving, sundararajan2017axiomatic, smilkov2017smoothgrad, ribeiro2016should, kim2016examples, brown2017adversarial, athalye2018synthesizing, su2019one}, where the neurons remain unexplained.
More recent efforts have attempted to associate hidden neurons with human-understandable concepts \cite{zhou2014object, zhang2018interpreting, zhang2019interpreting, olah2018feature, olah2018building, carter2019exploring, bau2017network, zhou2018interpreting, bau2018gan, bau2020understanding, ghorbani2020neuron}. Although insightful interpretations of neurons' semantics have been demonstrated, such as identifying the neurons controlling contents of \emph{trees} \cite{bau2020understanding}, existing methods define the concepts in an ad-hoc manner, which heavily rely on human annotations such as manual visual inspection \cite{zhou2014object, olah2018feature, olah2018building, carter2019exploring}, manually labeled classification categories \cite{ghorbani2020neuron}, or hand-crafted guidance images \cite{bau2017network, zhou2018interpreting, bau2018gan, bau2020understanding}.  They thus suffer from heavy costs and scalability issues. Moreover, existing methods often ignore the inherent relationships among different concepts (\eg, \emph{dog} and \emph{cat} both belong to \emph{mammal}), and treat them independently, which therefore loses the chance to discover neurons responsible for implicit higher-level concepts (\eg, \emph{canine}, \emph{mammal}, and \emph{animal}) and explore whether the network can create abstractions of things like our humans do.

\begin{figure*}
    \centering
    \includegraphics[width=17.2cm]{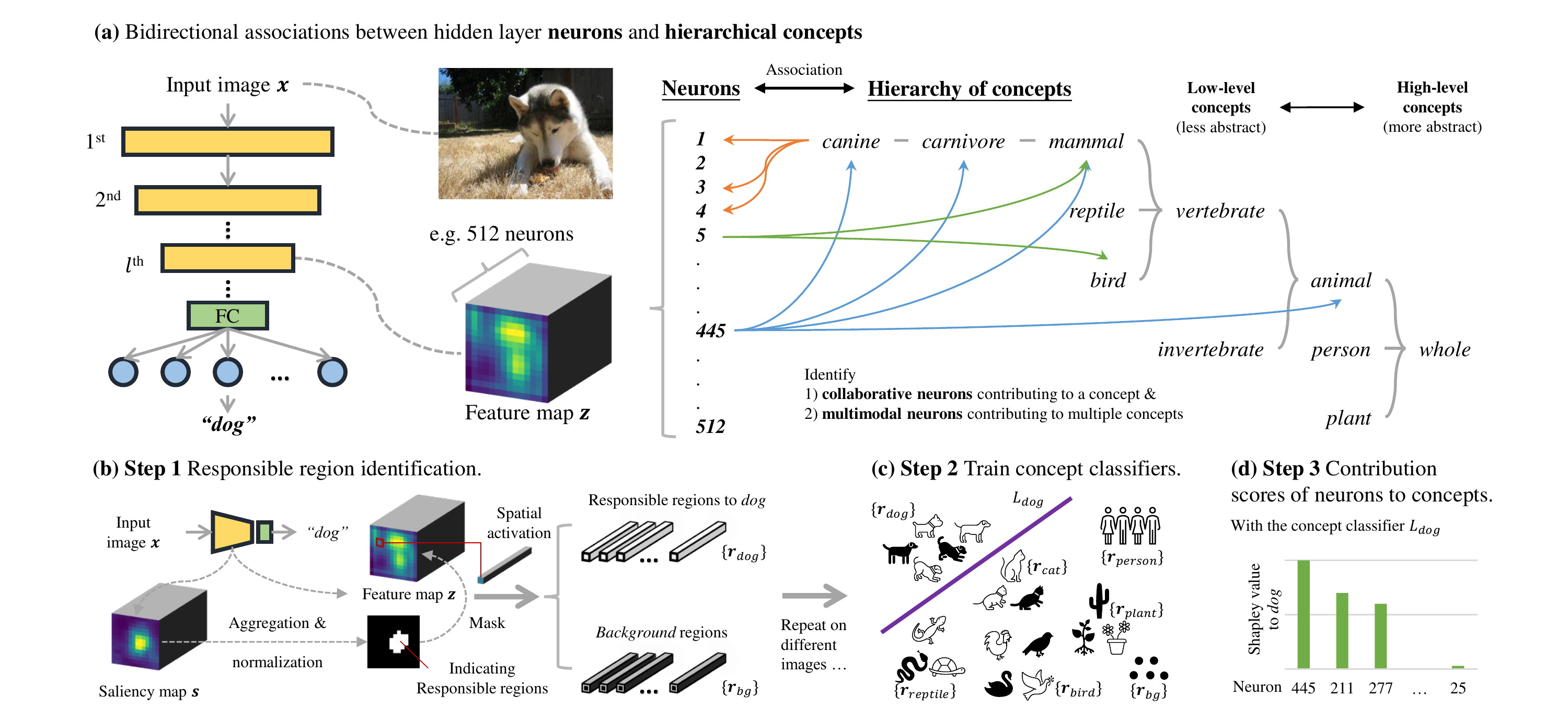}
    \setlength{\abovecaptionskip}{0pt}
    \setlength{\belowcaptionskip}{0pt}
    \caption{Overall illustration of HINT. \textbf{(a)} HINT is able to build bidirectional associations between hidden layer neurons and hierarchical concepts. It can also identify collaborative neurons and multimodal neurons. Further, HINT helps to indicate how the neurons learn the hierarchical relationships of categories. \textbf{(b)-(c)} Main steps. See Section \ref{sec:responsible_region_identification_for_concepts} for Step 1, Section \ref{sec:training_of_concept_classifiers} for Step 2, and Section \ref{sec:method_contribution_scores_of_neurons_to_concepts} for Step 3.}
    \label{fig:fig_central_illustration}
\end{figure*}

The above motivates us to rethink how concepts should be defined to more faithfully reveal the roles of hidden neurons. We draw inspirations from the hierarchical cognition process of human beings-- human tend to organize things from specific to general categories \cite{quillian1968semantic, warrington1975selective, mcclelland2003parallel}-- and propose to explore hierarchical concepts which can be harvested from WordNet \cite{miller1995wordnet} (a lexical database of semantic relations between words). 
We investigate whether deep networks can automatically learn the hierarchical relationships of categories that were not labeled in the training data. More concretely, we aim to identify neurons for both low-level concepts such as \emph{Malamute}, \emph{Husky}, and \emph{Persian cat}, and the implicit higher-level concepts such as \emph{dog} and \emph{animal} as shown in Figure \ref{fig:fig_central_illustration} (a). (Note that we call less abstract concepts low-level and more abstract concepts high-level.)

To this end, we develop \textbf{HIerarchical Neuron concepT explainer} (\textbf{HINT}) which builds a bidirectional association between neurons and hierarchical concepts (see Figure \ref{fig:fig_central_illustration}). First, we develop a saliency-guided approach to identify the high dimensional representations associated with the hierarchical concepts on hidden layers (noted as responsible regions in Figure \ref{fig:fig_central_illustration} (b)), which makes HINT low-cost and scalable as no extra hand-crafted guidance is required. Then, we train classifiers shown in Figure \ref{fig:fig_central_illustration} (c) to separate different concepts' responsible regions where the weights represent the contribution of the corresponding neuron to the classification. 
Based on the classifiers, we design a Shapley value-based scoring method to fairly evaluate neurons' contributions, considering both neurons' individual and collaborative effects.

To our knowledge, HINT presents the first attempt to associate neurons with hierarchical concepts, which enables us to systematically and quantitatively study whether and how hierarchical concepts are embedded into deep network neurons. HINT identifies collaborative neurons contributing to one concept and multimodal neurons contributing to multiple concepts. Especially, HINT finds that, despite being trained with only low-level labels, such as \emph{Husky} and \emph{Persian cat}, deep neural networks automatically embed hierarchical concepts into its neurons. Also, HINT is able to discover responsible neurons to both higher-level concepts, such as \emph{animal}, \emph{person} and \emph{plant}, and lower-level concepts, such as \emph{mammal}, \emph{reptile} and \emph{bird}.

Finally, we verify the faithfulness of neuron-concept associations identified by HINT with a Weakly Supervised Object Localization task. In addition, HINT achieves remarkable performance in a variety of applications,
including saliency method evaluation, adversarial attack explanation, and COVID19 classification model evaluation,
further manifesting the usefulness of HINT.

\section{Related Work}
\label{sec:related_work}

\noindent{\textbf{Neuron-concept Association Methods.}} Neuron-concept association methods aim at directly interpreting the internal computation of CNNs \cite{chen2019looks, bargal2019guidedzoom, gu2019semantics, mu2020compositional}. Early works show that neurons on shallower layers tend to learn simpler concepts, such as lines and curves, while higher layers tend to learn more abstract ones, such as heads or legs \cite{zeiler2014visualizing, yosinski2015understanding}. TCAV \cite{kim2018interpretability} and related studies \cite{ghorbani2019towards, graziani2018regression} quantify the contribution of a given concepts represented by guidance images to a target class on a chosen hidden layer. Object Detector \cite{zhou2014object} visualizes the concept-responsible region of a neuron in the input image by iteratively simplifying the image. After that, Network Dissection \cite{bau2017network, zhou2018interpreting, bau2020understanding} quantifies the roles of neurons by assigning each neuron to a concept with the guidance of extra images. GAN Dissection \cite{bau2018gan, bau2020understanding} illustrates the effect of concept-specific neurons by altering them and observing the emergence and vanishing of concept-related contents in images. Neuron Shapley \cite{ghorbani2020neuron} identifies the most influential neuron over all hidden layers to an image category by sorting Shapley values \cite{shapley201617}. Besides pre-defined concepts, feature visualization methods \cite{olah2018feature, olah2018building, carter2019exploring} generate Deep Dream-style \cite{mordvintsev2015inceptionism} explanations for each neuron and manually interpret their meanings. Additionally, Net2Vec \cite{fong2018net2vec} maps semantic concepts to vectorial embeddings to investigate the relationship between CNN filters and concepts. However, existing methods cannot systematically explain how the network learns the inherent relationships of concepts, and suffer from high cost and scalability issues. HINT is proposed to overcome the limitations and goes beyond exploring each concept individually -- it adopts hierarchical concepts to explore their semantic relationships.


\noindent{\textbf{Saliency Map Methods.}} Saliency map methods are a stream of simple and fast interpretation methods which show the pixel responsibility (\ie saliency score) in the input image for a target model output. There are two main categories of saliency map methods -- backpropagation-based and perturbation-based. Backpropagation-based methods mainly generate saliency maps by gradients; they include Gradient \cite{simonyan2013deep}, Gradient x Input \cite{shrikumar2016not}, Guided Backpropagation \cite{springenberg2014striving}, Integrated Gradient \cite{sundararajan2017axiomatic}, SmoothGrad \cite{smilkov2017smoothgrad}, LRP \cite{bach2015pixel, gu2018understanding}, Deep Taylor \cite{montavon2017explaining}, DeepLIFT \cite{shrikumar2017learning}, and Deep SHAP \cite{chen2021explaining}. 
Perturbation-based saliency methods perturbate input image pixels and 
observe the variations of model outputs; they include Occlusion \cite{zeiler2014visualizing}, RISE \cite{petsiuk2018rise}, Real-time \cite{dabkowski2017real}, Meaningful Perturbation \cite{fong2017interpretable}, and Extremal Perturbation \cite{fong2019understanding}. Inspired by saliency methods, in HINT, we build a saliency-guided approach to identify the responsible regions for each concept on hidden layers.


\section{Method}
\label{sec:method}

\noindent{\textbf{Overview.}} Considering a CNN classification model $f$ and a hierarchy of concepts $\mathcal{E}: \{e\}$ (see Figure \ref{fig:fig_central_illustration} (a)), our goal is to identify bidirectional associations between neurons and hierarchical concepts. 
To this end, we develop \textbf{HIerarchical Neuron concepT explainer} (\textbf{HINT}) to quantify the contribution of each neuron $d$ to each concept $e$ by a contribution score $\phi$ where higher contribution value means stronger association between $d$ and $e$, and vice versa.

The key problem therefore becomes how to estimate the score $\phi$ for any pair of $e$ and $d$. We achieve this by identifying how the network map concept $e$ to a high dimensional space and quantifying the contribution of $d$ for the mapping. First, given a concept $e$ and an image $\mbox{\boldmath$x$}$, on feature map $\mbox{\boldmath$z$}$ of the $l^{th}$ layer, HINT identifies the responsible regions $\mbox{\boldmath$r$}_{e}$ to concept $e$ by developing a saliency-guided approach elaborated in Section \ref{sec:responsible_region_identification_for_concepts}. Then, given the identified regions for all the concepts, HINT further trains concept classifier $L_{e}$ to separate concept $e$'s responsible regions $\mbox{\boldmath$r$}_{e}$ from other regions $\mbox{\boldmath$r$}_{\mathcal{E} \backslash e} \cup \mbox{\boldmath$r$}_{b^{*}}$ where $b^{*}$ represents background (see Section \ref{sec:training_of_concept_classifiers}). Finally, to obtain $\phi$, we design a Shapley value-based approach to fairly evaluate the contribution of each neuron $d$ from the concept classifiers (see Section \ref{sec:method_contribution_scores_of_neurons_to_concepts}).


\begin{algorithm}
\KwInput{A set of images with hierarchical concepts $\{(\mbox{\boldmath$x$}, e)\}$, a set of neurons $\mathcal{D}$ for experiment, modified saliency method $\Lambda$, aggregation method $\zeta$, and threshold $t \in (0, 10$.}
\KwOutput{Score matrix $\Phi$ where every element $\phi$ is the Shapely value of neuron $d$ to concept $e$.}
\KwInit{Responsible region containers $\mbox{\boldmath$r$}_{e} = \{ \ \}$ for each $e$ in $\mathcal{E}$, background region container $\mbox{\boldmath$r$}_{b^{*}} = \{ \ \}$, and score matrix $\Phi = \{0\}^{|\mathcal{D}| \times |\mathcal{E}|}$.}

\For{each $(\mbox{\boldmath$x$}, e)$} {
    feature map $\mbox{\boldmath$z$} = f_{l}(\mbox{\boldmath$x$})$ \;
    saliency map $\mbox{\boldmath$s$} = \Lambda(\mbox{\boldmath$x$}, f_{l} \ | \ e)$ \;
    $\mbox{\boldmath$z$} \leftarrow \mbox{\boldmath$z$}_{\mathcal{D}, :, :}$ \;
    $\mbox{\boldmath$s$} \leftarrow \mbox{\boldmath$s$}_{\mathcal{D}, :, :}$ \;
    $\hat{\mbox{\boldmath$s$}} = Normalization(\zeta(\mbox{\boldmath$s$})) \in [0, 1]^{H_{l} \times W_{l}}$ \;
    $\mbox{\boldmath$z$}_{e} = \mbox{\boldmath$z$} \odot (\hat{\mbox{\boldmath$s$}} \geq t)$, add $\mbox{\boldmath$z$}_{e}$ to $\mbox{\boldmath$r$}_{e}$ \;
    $\mbox{\boldmath$z$}_{b^{*}} = \mbox{\boldmath$z$} \odot (\hat{\mbox{\boldmath$s$}} < t)$, add $\mbox{\boldmath$z$}_{b^{*}}$ to $\mbox{\boldmath$r$}_{b^{*}}$ \;
}
\For{each $e$ in $\mathcal{E}$} {
    Train classifier $L_{e}$ which separates $\mbox{\boldmath$r$}_{e}$ and $\mbox{\boldmath$r$}_{\mathcal{E} \backslash e} \cup \mbox{\boldmath$r$}_{b^{*}}$
}
\For{each $e$ in $\mathcal{E}$}{
    \For {each $d$ in $\mathcal{D}$}{
    $\phi =$ Shapley value of neuron $d$ to concept $e$\;
    Update $\Phi$ with $\phi$\;
    }
}
\caption{HINT}
\label{algo:algorithm_of_HINT}
\end{algorithm}

\subsection{Responsible Region Identification for Concepts}
\label{sec:responsible_region_identification_for_concepts}

In this section, we introduce our saliency-guided approach to collect the responsible regions $\mbox{\boldmath$r$}_{e}$ for a certain concept $e \in \mathcal{E}$ to serve as the training samples of the concept classifier which will be described in Section~\ref{sec:training_of_concept_classifiers}.

Taking an image $\mbox{\boldmath$x$}$ containing a concept $e$ as input, the network $f$ generates a feature map $\mbox{\boldmath$z$} \in \mathbb{R}^{D_{l} \times H_{l} \times W_{l}}$ where there are $D_{l}$ neurons in total. Generally, not all regions of $\mbox{\boldmath$z$}$ are equally related to $e$~\cite{zhang2019interpreting}. In other words, some regions have stronger correlations with $e$ while others are less correlated, as shown in Figure~\ref{fig:fig_central_illustration} (b) ``Step 1". Based on the above observation, we propose a saliency-guided approach to identify the closely related regions $\mbox{\boldmath$r$}_{e}$ to the concept $e$ in feature map $z$. We call them responsible regions.

{First, we obtain the saliency map on the $l^{th}$ layer.} As shown in Figure \ref{fig:fig_central_illustration} (b) ``Step 1", with the feature map $\mbox{\boldmath$z$}$ on the $l^{th}$ layer extracted, we derive the $l^{th}$ layer's saliency map $\mbox{\boldmath$s$}$ with respect to concept $e$ by the saliency map estimation approach $\Lambda$. Note that HINT is compatible with different back-propagation based saliency map estimation methods. We implement five of them~\cite{simonyan2013deep, shrikumar2016not, springenberg2014striving, sundararajan2017axiomatic, smilkov2017smoothgrad}, please refer to the Supplementary Material for more details. Note that different from existing works~\cite{simonyan2013deep, shrikumar2016not, springenberg2014striving, sundararajan2017axiomatic, smilkov2017smoothgrad} that pass the gradients to the input image as saliency scores, we early stop the back-propagation at the $l^{th}$ layer to obtain the saliency map $\mbox{\boldmath$s$}$. Here, we use modified SmoothGrad~\cite{smilkov2017smoothgrad} as an example to demonstrate our approach: $\textstyle \Lambda = \frac {1} {N} \sum_{n=1}^{N} \frac {\partial f^{e}(\mbox{\boldmath$x$}^{'})} {\partial \mbox{\boldmath$z$}^{'}}$ where $\textstyle \mbox{\boldmath$x$}^{'} = \mbox{\boldmath$x$} + \mathcal{N}(\mu, \sigma^{2}_{n})$ and $\mathcal{N}$ indicates normal distribution. It is notable that we may also optimally select part of the neurons $\mathcal{D}$ for analysis.

{Next is to identify the responsible regions on feature map $\mbox{\boldmath$z$}$ with the guidance of the saliency map $\mbox{\boldmath$s$}$.} Specifically, we categorize each entry $z_{\mathcal{D},i,j}$ in $\mbox{\boldmath$z$}$ to be responsible to $e$ or not. To this end, the saliency map $\mbox{\boldmath$s$}$ is first aggregated by an aggregation function $\zeta$ along the channel dimension and then normalized to be within $[0, 1]$. Note that different aggregation functions $\zeta$ can be applied (see five different $\zeta$ shown in Supplementary Material). Here, we aggregate ${\mbox{\boldmath$s$}}$ using Euclidean norm $\textstyle \zeta = \| \mbox{\boldmath$s$} \|$ along its first dimension. After that, we obtain $\hat{\mbox{\boldmath$s$}} \in [0, 1]^{H_{l} \times W_{l}}$ with each element $s_{i,j}$ indicating the relevance of $z_{\mathcal{D},i,j}$ to concept $e$. By setting a threshold $t$ ( we set $t$ as $0.5$ in the paper) and masking ${\mbox{\boldmath$z$}}$ with $\hat{\mbox{\boldmath$s$}} \geq t$ and $\hat{\mbox{\boldmath$s$}} < t$, we finally obtain responsible regions and background regions respectively (see the illustration of the two regions Figure \ref{fig:fig_central_illustration} (b): ``Step 1").

Our saliency-guided approach extends the interpretability of saliency methods, which originally aim to find the ``responsible regions" to a concept on one particular image. However, our approach is able to identify ``responsible regions" to a concept on the high dimensional space of a hidden layer from multiple images, which can more accurately describe how the network represents concept $e$ internally. Therefore, our saliency-guided approach provides better interpretability as it helps us to investigate the internal abstraction of concept $e$ in the network.

\subsection{Training of Concept Classifiers}
\label{sec:training_of_concept_classifiers}

For all images, we identify its responsible regions for each concept $e \in \mathcal{E}$ following the procedures described in \ref{sec:responsible_region_identification_for_concepts} and construct a dataset which contains a collection of responsible regions $\mbox{\boldmath$r$}_{e}$ and a collection of background regions $\mbox{\boldmath$r$}_{b^{*}}$.
Given the dataset, as shown in Figure \ref{fig:fig_central_illustration} (c) ``Step 2", we use the high dimensional CNN hidden layer features to train a concept classifier $L_{e}$  which distinguishes $\mbox{\boldmath$r$}_{e}$ from $\mbox{\boldmath$r$}_{\mathcal{E} \backslash e} \cup \mbox{\boldmath$r$}_{b^{*}}$, \ie, separating concept $e$ from other concepts ${\mathcal{E} \backslash e} \cup {b^{*}}$  (Line 9 and 10 in Algorithm \ref{algo:algorithm_of_HINT}).

$L_{e}$ can have many forms: a linear classifier, a decision tree, a Gaussian Mixture Model, and so on. Here, we use the simplest form, a linear classifier, which is equivalent to a hyperplane separating concept $e$ from others in the high dimensional feature space of CNN.

\begin{equation}
L_{e}(r) = \sigma \left( \mbox{\boldmath$\alpha$}^{T} r \right),
\label{equ:linear_classifier}
\end{equation}
where $r = z_{\mathcal{D},i,j} \in \mathbb{R}^{|\mathcal{D}|}$ represents spatial activation with each element representing a neuron; $\mbox{\boldmath$\alpha$}$ is a vector of weights, $\sigma$ is a sigmoid function, and $L_{e}(r) \in [0, 1]$ represents the confidence of $r$ related to a concept $e$.

It is notable that we can apply the concept classifier $L_{e}$ back to the feature map ${\mbox{\boldmath$z$}}$ to visualize how $L_{e}$ detect concept $e$. Classifiers of more abstract concepts (\eg, \emph{whole}) tend to activate regions of more general features, which helps us to locate the entire extent of the object. On contrary, classifiers of lower-level concepts tend to activate regions of discriminative features such as eyes and heads.

\begin{figure*}
    \centering
    \includegraphics[width=17cm]{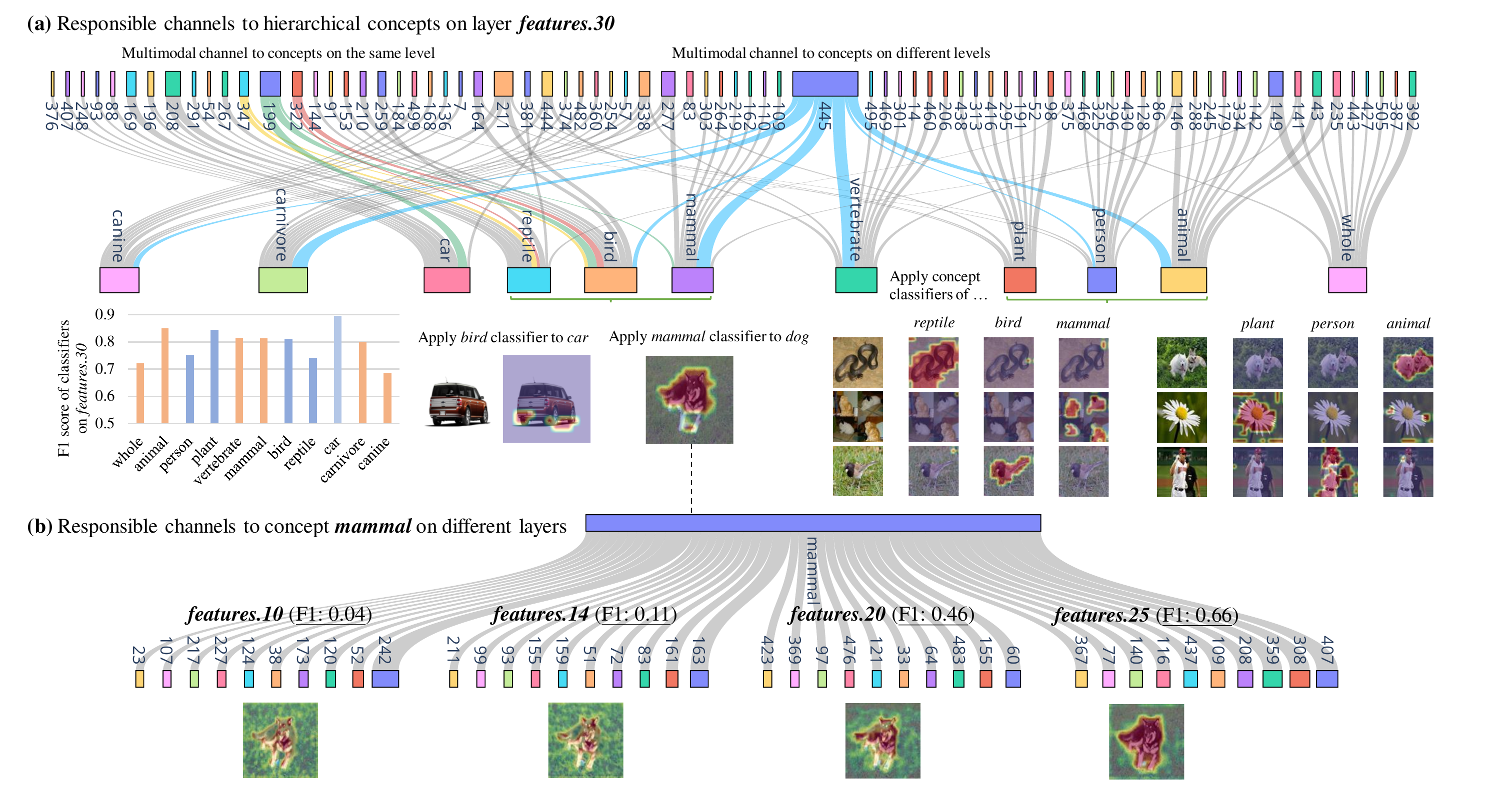}
    \setlength{\abovecaptionskip}{0pt}
    \setlength{\belowcaptionskip}{0pt}
    \caption{Bidirectional associations between neurons and hierarchical concepts. The width of the link indicates the size of the contribution score of a neuron to a concept. \textbf{(a)} Responsible neurons to hierarchical concepts (see the hierarchy in Figure \ref{fig:fig_central_illustration}) on layer features.30 in VGG19. The F1 scores of concept classifiers show their capability of distinguishing the target concepts. The pictures illustrate the results of applying concept classifiers on different images. For most of the cases, the concept classifiers only locate the objects belonging to the target concepts. However, as \emph{bird} and \emph{car} share multimodal neurons, the \emph{bird} classifier responses to the wheels of the car. \textbf{(b)} Responsible neurons to \emph{mammal} on different layers. The pictures and F1 scores indicate the network can more easily distinguish \emph{mammal} from other concepts as the layer goes higher.}
    \label{fig:fig_Responsible_neurons_for_hierarchical_concepts}
\end{figure*}

\subsection{Contribution Scores of Neurons to Concepts}
\label{sec:method_contribution_scores_of_neurons_to_concepts}

Next is to decode the contribution score $\phi$ from the concept classifiers.
A simple method to estimate $\phi$ is to use the learned classifier weights corresponding to each neuron $e$, where a higher value typically means a larger contribution \cite{molnar2020interpretable}.
However, the assumption that $\alpha$ can serve as the contribution score is that the neurons are independent of each other, which is generally not true. 
To achieve a fair evaluation of neurons' contributions to $e$, a Shapley value-based approach is designed to calculate the scores $\phi$, which can take account of neurons' individual effects as well as the contributions coming from the collaboration with others.

Shapely value \cite{shapley201617} is from Game Theory, which evaluates channels’ individual and
collaborative effects. More specifically, if a channel cannot be used for classification independently but can greatly improve classification accuracy when collaborating with other
channels, its Shapley value can still be high. Shapely value satisfies the properties of efficiency, symmetry, dummy, and additivity \cite{molnar2020interpretable}. Monte-Carlo sampling is used to estimate the Shapley values by testing the target neuron's possible coalitions with other neurons. Equation \eqref{equ:shapley_value_calculation} shows how we calculate Shapley value $\phi$ of a neuron $d$ to concept $e$.

\begin{equation}
\phi = \frac{ \sum_{\mbox{\boldmath$r$}} \left| \sum_{i=1}^{M} \left( L_{e}^{\langle \mathcal{S} \cup d \rangle}(r) - L_{e}^{\langle \mathcal{S} \rangle}(r) \right) \right| } {M|\mbox{\boldmath$r$}_{\mathcal{E}} \cup \mbox{\boldmath$r$}_{b^{*}}|}
\label{equ:shapley_value_calculation}
\end{equation}
where $r = z_{\mathcal{D},i,j}$ represents spatial activation from $\mbox{\boldmath$r$}_{\mathcal{E}}$ and $\mbox{\boldmath$r$}_{b^{*}}$; $\textstyle \mathcal{S} \subseteq \mathcal{D} \backslash d$ is the neuron subset randomly selected at each iteration; $\textstyle \langle * \rangle$ is an operator keeping the neurons in the brackets, \ie, $\mathcal{S} \cup d$ or $\mathcal{S}$, unchanged while randomizing others; $M$ is the number of iterations of Monte-Carlo sampling; $L_{e}^{\langle * \rangle}$ means that the classifier is re-trained with neurons in the brackets unchanged and others being randomized.

By repeating the calculation for different $e$ and $d$ (see Line 11 to line 14 in Algorithm \ref{algo:algorithm_of_HINT}), finally, we can get the score matrix $\Phi$.

\subsection{Neuron-Concept Association}

By repeating the score calculations for all pairs of $e$ and $d$, we obtain a score matrix $\Phi$ where each row represents a neuron $d$ and each column represents a concept $e$ in the hierarchy.
By sorting the scores in the column of concept $e$, we can get collaborative neurons all having high contributions to a concept $e$. Also, by sorting the scores in the row of neuron $d$, we can test whether $d$ is multimodal (having high scores to multiple concepts) and observe a hierarchy of concepts that $d$ is responsible for.

Note that the score matrix $\Phi$ cannot tell us the exact number of responsible neurons to concept $e$. For a contribution score $\phi$ which is zero or near zero, the corresponding neuron $d$ can be regarded as irrelevant to the corresponding concept $e$. Therefore, for truncation, we may set a threshold for $\phi$. In our experiment, for a concept, we sort scores and select the top $N$ as responsible neurons.

\section{Experiments}
\label{sec:experiments}

\subsection{Experimental setup}

\begin{figure*}
    \centering
    \includegraphics[width=17cm]{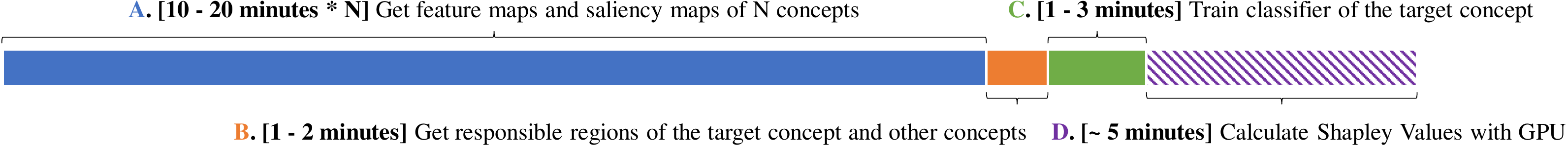}
    \setlength{\abovecaptionskip}{0pt}
    \setlength{\belowcaptionskip}{0pt}
    \caption{Time consumption for different stages of HINT. The most time consuming part is the data preparation process. Shapely value computation takes about $5$ minutes with a single NVIDIA RTX 2080, while linear classifier training takes $1-3$ minutes. Therefore, the time consumption of Shapely value calculation is negligible when considering the whole cycle.}
    \label{fig:fig_time_consumption}
\end{figure*}

HINT is a general framework which can be applied on any CNN architectures. We evaluate HINT on several models trained on ImageNet \cite{deng2009imagenet} with representative CNN backbones including VGG-16 \cite{simonyan2014very}, VGG-19 \cite{simonyan2014very}, ResNet-50 \cite{he2016deep}, and Inception-v3 \cite{szegedy2016rethinking}. In this paper, the layer names are from PyTorch pretrained models (\eg, ``features.30'' is a layer name of VGG19). The hierarchical concept set $\mathcal{E}$ is built upon the $1000$ categories of ImageNet with hierarchical relationship is defined by WordNet \cite{miller1995wordnet} as shown in Figure \ref{fig:fig_central_illustration}. Figure \ref{fig:fig_time_consumption} shows the computational complexity analysis, indicating that Shapely value calculation is negligible when considering the whole cycle.

\subsection{Responsible Neurons to Hierarchical Concepts}
\label{sec:experiments_responsible_neurons_for_hierarchical_concepts}

In this section, we study the responsible neurons for the concepts and show the hierarchical cognitive pattern of CNNs. We adopt the VGG-19 backbone and show the top-10 significant neurons to each concept ($N$=10). The results in Figure \ref{fig:fig_Responsible_neurons_for_hierarchical_concepts} manifest that HINT explicitly reveals the hierarchical learning pattern of the network: some neurons are responsible to concepts with higher semantic levels such as \emph{whole} and \emph{animal}, and others are responsible to more detailed concepts such as \emph{canine}. Besides, HINT shows that there can be multiple neurons contributing to a single concept and HINT identifies multimodal neurons which have high contributions to multiple concepts.

\noindent{\textbf{Concepts of different levels.}} First, we investigate the concepts of different levels in Figure \ref{fig:fig_Responsible_neurons_for_hierarchical_concepts} (a). Among all the concepts, \emph{whole} has the highest semantic level including \emph{animal}, \emph{person}, and \emph{plant}. 
To study how a network recognizes a \emph{Husky} (a subclass of \emph{canine}) image on a given layer, HINT hierarchically identifies the neurons which are responsible for the concept from higher levels (like \emph{whole}, \emph{animal}) to lower ones (like \emph{canine}). Besides, HINT is able to identify multimodal neurons which take responsibility to many concepts at different semantic levels. For example, the $445^{th}$ neuron delivers the most contribution to multiple concepts including \emph{animal}, \emph{vertebrate}, \emph{mammal}, and \emph{carnivore}, and also contributes to \emph{canine}, manifesting that the $445^{th}$ neuron captures the general and specie-specific features which are not labeled in the training data.

\noindent{\textbf{Concepts of the same level.}} Next, we study the responsible neurons for concepts at the same level identified by HINT. For \emph{mamml}, \emph{reptile}, and \emph{bird}, there exist multimodal neurons as the three categories share morphological similarities. For example, the $199^{th}$ and $445^{th}$ neurons contribute to both \emph{mammal} and \emph{bird}, while the $322^{nd}$ and $347^{th}$ neurons are individually responsible for both \emph{reptile} and \emph{bird}. Interestingly, HINT identifies multimodal neurons contributing to concepts which are conceptually far part to humans. For example, the $199^{th}$ neuron contributes to both \emph{bird} and \emph{car}. By applying the \emph{bird} classifier to images of \emph{bird} and \emph{car}, we find that the body of the \emph{bird} and the wheels of the \emph{car} can be both detected.

\noindent{\textbf{Same concept on different layers.}} We also identify responsible neurons on different network layers with HINT. Figure \ref{fig:fig_Responsible_neurons_for_hierarchical_concepts} (b) illustrates the 10 most responsible neurons to \emph{mammal} in other four network layers. On shallow layers, such as on layer features.10, HINT indicates that the concept of \emph{mammal} cannot be recognized by the network (F1 score: 0.04). However, as the network goes deeper, the F1 score of \emph{mammal} classifier increases until around 0.8 on layer features.30, which is consistent with the existing works \cite{zeiler2014visualizing, yosinski2015understanding} that deeper layers capture higher-level and richer semantic meaningful features.

\subsection{Verification of Associations by Weakly Supervised Object Localization}
\label{sec:experiments_localization}

With the associations between neurons and hierarchical concepts obtained by HINT, we further validate the associations using Weakly Supervised Object Localization (WSOL). Specifically, we train a concept classifier $L_{e}$ (see detailed steps in Section \ref{sec:responsible_region_identification_for_concepts} and \ref{sec:training_of_concept_classifiers}) with the top-$N$ significant neurons corresponding to concept $e$ at a certain layer, and locate the responsible regions using $L_{e}$ as the localization results. Good localization performance of $L_{e}$ indicates the $N$ neurons also have high contributions to concept $e$.

\noindent{\textbf{Comparison of localization accuracy.}} Quantitative evaluation in Table \ref{tab:comparison_of_localization_on_CUB-200-2011} and \ref{tab:comparison_of_localization_on_ImageNet} show that HINT achieves comparable performance with existing WSOL approaches, thus validating the associations. 
We train \emph{animal} (Table \ref{tab:comparison_of_localization_on_CUB-200-2011}) and \emph{whole} (Table \ref{tab:comparison_of_localization_on_ImageNet}) classifiers with 10\%, 20\%, 40\%, 80\% neurons sorted and selected by Shapley values on layer ``features.26'' (512 neurons) of VGG16, layer ``layer3.5'' (1024 neurons) of ResNet50, and layer ``Mixed\_6b'' (768 neurons) of Inception v3, respectively. To be consistent with the commonly-used WSOL metric, Localization Accuracy measures the ratio of images with IoU of groundtruth and predicted bounding boxes larger than 50\%. In Table \ref{tab:comparison_of_localization_on_CUB-200-2011}, we compare HINT with the state-of-the-art methods on dataset CUB-200-2011 \cite{wah2011caltech}, which contains images of 200 categories of birds. Note that existing localization methods need to re-train the model on the CUB-200-2011 as they are tailored to the classifier while HINT directly adopts the classifier trained on ImageNet without further finetuning on CUB-200-2011. Even so, HINT still achieves a comparable performance when adopting VGG16 and Inception v3, and performs the best when adopting ResNet50. However, Table \ref{tab:comparison_of_localization_on_ImageNet} shows that HINT outperforms all existing methods on all models on ImageNet.
Besides, the differences of localization accuracy may indicate different models have different learning modes. Precisely, few neurons in VGG16 are responsible for \emph{animal} or \emph{whole} while most neurons in ResNet50 contribute to identifying \emph{animal} or \emph{whole}. In conclusion, the results quantitatively prove that the associations are valid and HINT achieves comparable performance to WSOL. More analysis is included in the supplementary file.


\begin{table}[t]
  \setlength{\abovecaptionskip}{0pt}
  \caption{Comparison of Localization Accuracy on CUB-200-2011. * indicates fine-tuning on CUB-200-2011.}
  \footnotesize
  \centering
  \begin{tabular}{llll}
    \toprule
    & VGG16 & ResNet50 & Inception v3\\
    \cmidrule(lr){2-4}
    CAM* \cite{zhou2016learning} & 34.4\% & 42.7\% & 43.7\% \\
    ACoL* \cite{zhang2018adversarial} & 45.9\% & - & - \\
    SPG* \cite{zhang2018self} & - & - & 46.6\% \\
    ADL* \cite{choe2019attention} & 52.4\% & 62.3\% & 53.0\% \\
    DANet* \cite{xue2019danet} & 52.5\% & - & 49.5\% \\
    EIL* \cite{mai2020erasing} & 57.5\% & - & - \\
    PSOL* \cite{zhang2020rethinking} & 66.3\% & 70.7\% & 65.5\% \\
    GCNet* \cite{lu2020geometry} & 63.2\% & - & - \\
    RCAM* \cite{bae2020rethinking} & 59.0\% & 59.5\% & - \\
    FAM* \cite{meng2021foreground} & \textbf{69.3}\% & \textbf{73.7}\% & \textbf{70.7}\% \\
    \rowcolor{gray!20} \textbf{Ours} (10\%) & \textbf{66.6}\% & 60.2\% & 49.0\% \\
    \rowcolor{gray!20} \textbf{Ours} (20\%) & 65.2\% & 67.1\% & 55.8\% \\
    \rowcolor{gray!20} \textbf{Ours} (40\%) & 61.3\% & 77.3\% & 52.8\% \\
    \rowcolor{gray!20} \textbf{Ours} (80\%) & 64.8\% & \textbf{80.2}\% & \textbf{56.2}\% \\
    \bottomrule
  \end{tabular}
  \label{tab:comparison_of_localization_on_CUB-200-2011}
\end{table}

\begin{table}[t]
  \setlength{\abovecaptionskip}{0pt}
  \caption{Comparison of Localization Accuracy on ImageNet.}
  \footnotesize
  \centering
  \begin{tabular}{llll}
    \toprule
    & VGG16 & ResNet50 & Inception v3\\
    \cmidrule(lr){2-4}
    CAM \cite{zhou2016learning} & 42.8\% & - & - \\
    ACoL \cite{zhang2018adversarial} & 45.8\% & - & - \\
    SPG \cite{zhang2018self} & - & - & 48.6\% \\
    ADL \cite{choe2019attention} & 44.9\% & 48.5\% & 48.7\% \\
    DANet \cite{xue2019danet} & - & - & 48.7\% \\
    EIL \cite{mai2020erasing} & 46.8\% & - & - \\
    PSOL \cite{zhang2020rethinking} & 50.9\% & 54.0\% & 54.8\% \\
    GCNet \cite{lu2020geometry} & - & - & 49.1\% \\
    RCAM \cite{bae2020rethinking} & 44.6\% & 49.4\% & - \\
    FAM \cite{meng2021foreground} & \textbf{52.0}\% & \textbf{54.5}\% & \textbf{55.2}\% \\
    \rowcolor{gray!20} \textbf{Ours} (10\%) & 64.7\% & 59.7\% & 53.1\% \\
    \rowcolor{gray!20} \textbf{Ours} (20\%) & \textbf{66.1}\% & 66.6\% & 54.1\% \\
    \rowcolor{gray!20} \textbf{Ours} (40\%) & 64.4\% & 69.4\% & 54.3\% \\
    \rowcolor{gray!20} \textbf{Ours} (80\%) & 62.6\% & \textbf{70.7}\% & \textbf{58.7}\% \\
    \bottomrule
  \end{tabular}
  \label{tab:comparison_of_localization_on_ImageNet}
\end{table}

\begin{figure*}
    \centering
    \includegraphics[width=17cm]{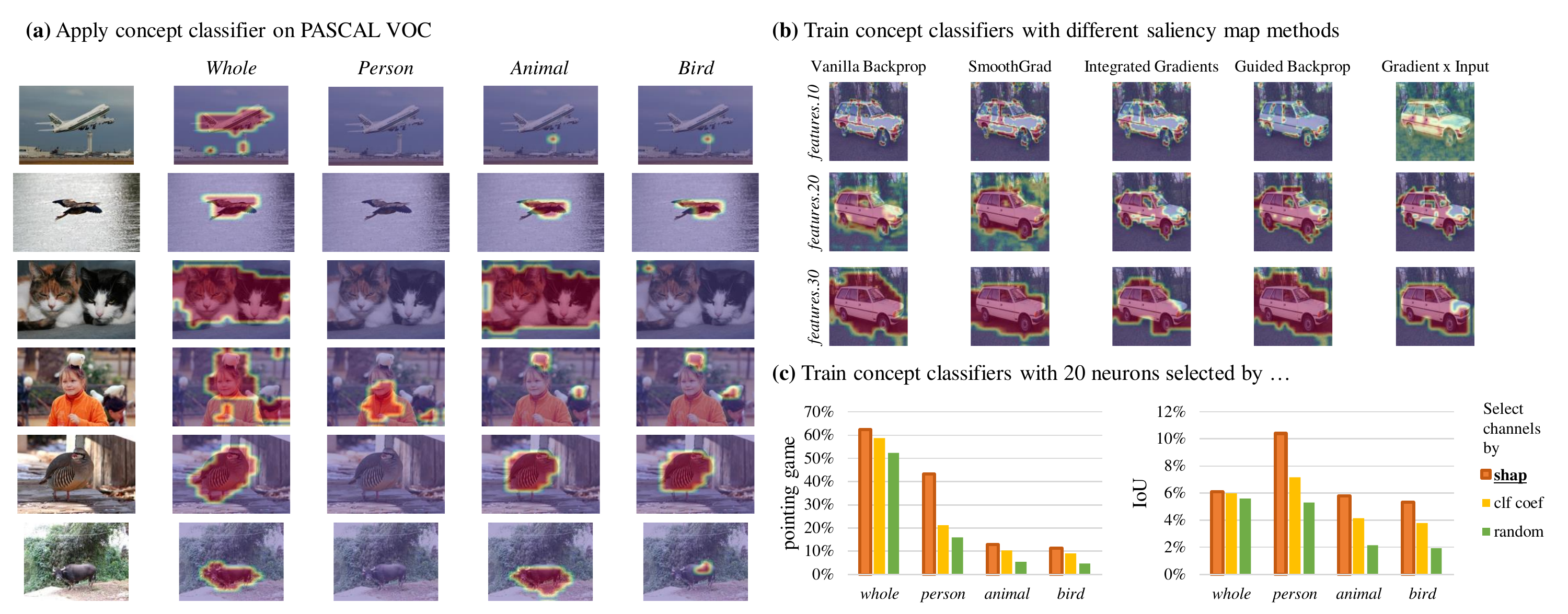}
    \setlength{\abovecaptionskip}{0pt}
    \setlength{\belowcaptionskip}{0pt}
    \caption{Results of Weakly Supervised Object Localization and ablation study. \textbf{(a)} Illustration of applying different concept classifiers on PASCAL VOC 2007, showing that HINT can locate objects of chosen concepts. \textbf{(b)} Ablation study showing the results of different saliency methods. \textbf{(c)} Ablation study showing Shapley values are good measures of neurons' contributions. The concept classifiers are trained with 20 neurons selected by different approaches. The pointing game (mask intersection over the groundtruth mask) and IoU (mask intersection over union of masks) scores show the accuracy of \emph{whole}, \emph{person}, \emph{animal}, and \emph{bird} concept classifiers on PASCAL VOC 2007.}
    \label{fig:fig_localization_and_ablation}
\end{figure*}

\noindent{\textbf{Flexible choice of localization targets.}} When locating objects, HINT has a unique advantage: a flexible choice of localization targets. We can locate objects on different levels in the concept hierarchy (\eg, \emph{bird}, \emph{mammal}, and \emph{animal}). In experiments, we train concept classifiers of \emph{whole}, \emph{person}, \emph{animal}, and \emph{bird} using 20 most important neurons on layer features.30 of VGG19 and apply them on PASCAL VOC 2007 \cite{pascal-voc-2007}. Figure \ref{fig:fig_localization_and_ablation} (a) shows that HINT can accurately locate the objects belonging to different concepts.

\noindent{\textbf{Extension to locate the entire extent of the object.}} Many existing WSOL methods adapt the model architecture and develop training techniques to highlight the entire extent rather than discriminative parts of object \cite{xue2019danet, mai2020erasing, zhang2020rethinking, lu2020geometry, bae2020rethinking, meng2021foreground}. However, can we effectively achieve this goal without model adaptation and retraining? HINT provid
es an approach to utilize the implicit concepts learned by the model. As shown in Figure \ref{fig:fig_localization_and_ablation} (c), classifiers of higher-level concepts (\eg \emph{whole}) tend to draw larger masks on objects than classifiers of lower-level concepts (\eg \emph{bird}). It is because that the responsible regions of \emph{whole} contain all the features of its subcategories. Naturally, the \emph{whole} classifier tends to activate full object regions rather than object parts.

\begin{figure*}
    \centering
    \includegraphics[width=17cm]{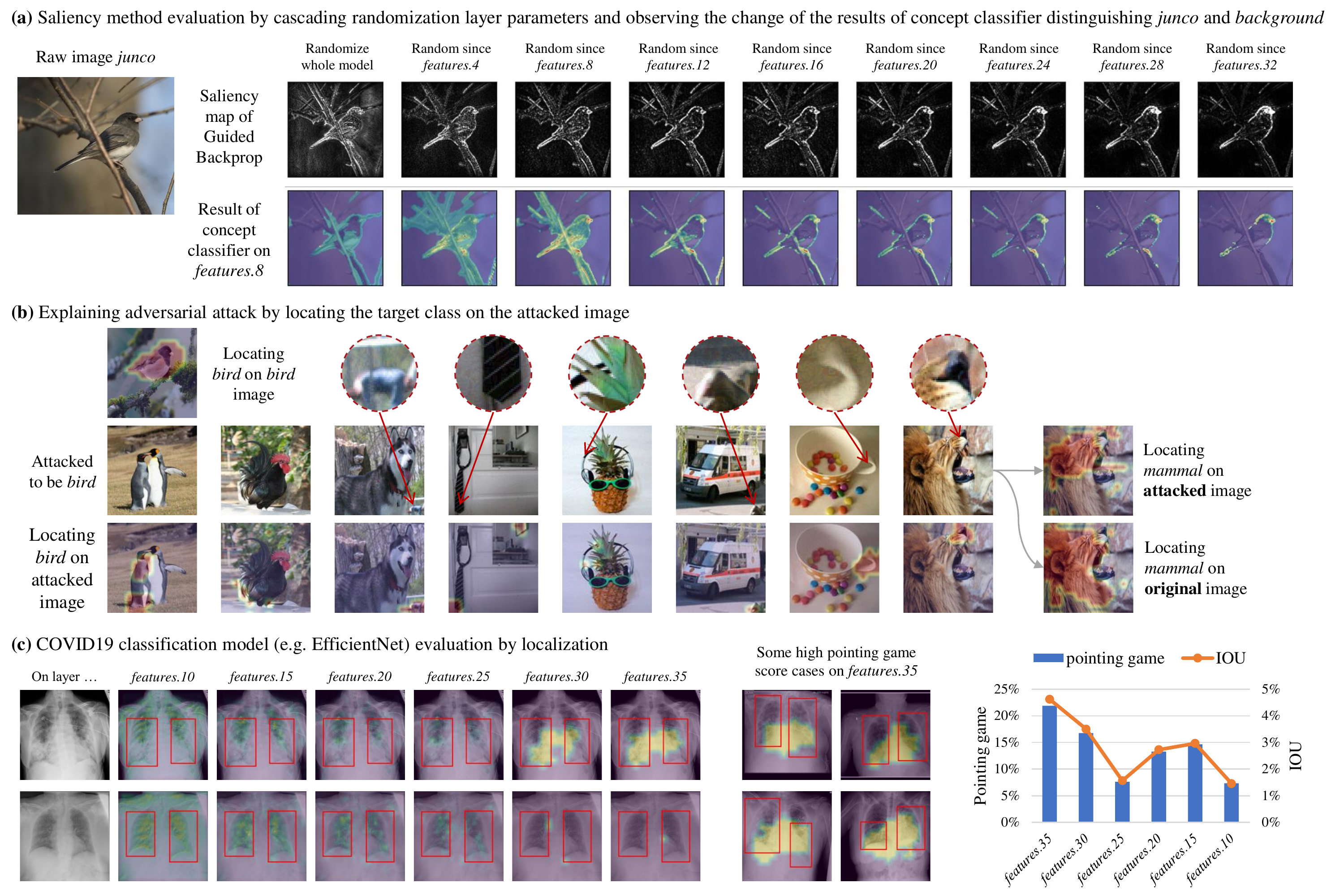}
    \setlength{\abovecaptionskip}{0pt}
    \setlength{\belowcaptionskip}{0pt}
    \caption{Other applications of HINT.
    \textbf{(a)} Saliency method evaluation.
    \textbf{(b)} Explaining adversarial attack.
    \textbf{(c)} COVID19 classification model evaluation.
    }
    \label{fig:fig_applications}
\end{figure*}

\subsection{Ablation Study}

We perform an ablation study to show that HINT is general and can be implemented with different saliency methods, and Shapley values are good measures of neurons' contributions to concepts.

\noindent{\textbf{Implementation with different saliency methods.}} We train concept classifiers with five modified saliency methods (see Supplementary Material). Then, we apply the classifiers to the object localization task. Figure \ref{fig:fig_localization_and_ablation} (b) shows that the five saliency methods all perform well. This shows that HINT is general, and different saliency methods can be integrated into HINT,

\noindent{\textbf{Shapley values.}} To test the effectiveness of Shapley values, we train concept classifiers using 20 neurons on layer features.30 of VGG19 by different selection approaches, including Shapley values (denoted as shap), absolute values of linear classifier coefficients (denoted as clf\_coef), and random selection (denoted as random). We then use the classifiers to perform localization tasks on PASCAL VOC 2007 (see Figure \ref{fig:fig_localization_and_ablation} (c)). Two metrics are used: pointing game (mask intersection over the groundtruth mask, usually used by other interpretation methods) \cite{zhang2018top} and IoU (mask intersection over the union of masks). The results show that ``shap'' outperforms ``clf\_coef'' and ``random'' when locating different targets. This suggests that Shapley value is a good measure of neuron contribution as it considers both the individual and collaborative effects of neurons. On contrary, linear classifier coefficients assume that neurons are independent of each other.

\subsection{More Applications}

We further demonstrate HINT's usefulness and extensibility by saliency method evaluation, adversarial attack explanation, and COVID19 classification model evaluation (Figure \ref{fig:fig_applications}). Please refer to Supplementary Material for detailed descriptions.

\noindent{\textbf{Saliency method evaluation.}} Guided Backpropagation can pass the sanity test in \cite{adebayo2018sanity, khakzar2021towards} if we observe the hidden layer results (see Figure \ref{fig:fig_applications} (a)). On layer features.8, with less randomized layers, the classifier-identified regions are more concentrated on the key features of the bird -- its beak and tail, thereby suggesting that Guided Backpropagation detects the salient regions.

\noindent{\textbf{Explaining adversarial attack.}} We attack images of various classes to be \emph{bird} using PGD \cite{madry2017towards} and apply the \emph{bird} classifier to their feature map.
The responsible regions for concept \emph{bird} highlighted in those fake \emph{bird} images may imply that, for certain images, adversarial attack does not change the whole body of the object to be another class but captures some details of the original image, where there exist shapes similar to \emph{bird} (see Figure \ref{fig:fig_applications} (b)). For example, in the coffee mug image where most shapes are round, adversarial attack catches the only pointed shape and attacks it to be like \emph{bird}. Upon above observations, we design a quantitative evaluation on the faithfulness of our explanations. First, we attack 300 images of other categories excluding \emph{bird} to be \emph{birds} based on VGG19 model.  Then, we use a \emph{bird} classifier to find the regions corresponding to the adversarial features of \emph{bird} on the attacked images. By visual inspection, we find most regions contain point shapes. Based on the regions, we train an adversarial attacked ``\emph{bird}'' classifier (``ad clf''). Finally, we use the ``ad clf'' to perform the WSOL task on real \emph{bird} images. The accuracy is 64.3\% (for true \emph{bird} classifier, it is 70.1\%), indicating HINT captures the adversarial \emph{bird} features and validates the explanation: some kind of adversarial attacks may be caused by attacking the similar shapes of the target class. 

\noindent{\textbf{COVID19 classification model evaluation}} Applying deep learning to the detection of COVID19 in chest radiographs has the potential to provide quick diagnosis and guide management in molecular test resource-limited situations. However, the robustness of those models remains unclear \cite{degrave2021ai}. We do not know whether the model decisions rely on confounding factors or medical pathology in chest radiographs. Object localization with HINT can check whether the identified responsible regions overlap with the lesion regions drawn by doctors (see Figure \ref{fig:fig_applications} (c)). As you can see, the pointing game and IoU are not high. Many cases having low pointing game and IoU values show that the model does not focus on the lesion region, while for the cases with high pointing game and IoU values, further investigation is still required to see whether they capture the medical pathology features or they just accidentally focus on the area of the stomach.

\section{Limitations of Interpretations}

HINT can systematically and quantitatively identify the responsible neurons to implicit high-level concepts. However, our approach cannot handle concepts that are not included in the concept hierarchy. And it is not effective to identify responsible neurons to concepts lower than the bottom level of the hierarchy which are the classification categories. More explorations are needed if we want to build such neuron-concept associations.

\section{Conclusion}
We have presented HIerarchical Neuron concepT explainer (HINT) method which builds bidirectional associations between neurons and hierarchical concepts in a low-cost and scalable manner. HINT systematically and quantitatively explains whether and how the neurons learn the high-level  hierarchical relationships of concepts in an implicit manner. Besides, it is able to identify collaborative neurons contributing to the same concept but also the multimodal neurons contributing to multiple concepts. Extensive experiments and applications manifest the effectiveness and usefulness of HINT. We open source our development package and hope HINT could inspire more investigations in this direction.


\section{Acknowledgments}

This work has been supported in part by Hong Kong Research Grant Council - Early Career Scheme (Grant No. 27209621), HKU Startup Fund, and HKU Seed Fund for Basic Research. Also, we thank Mr. Zhengzhe Liu for his insightful comments and careful editing of this manuscript.

{\small
\bibliographystyle{ieee_fullname}
\bibliography{egbib}
}

\clearpage
\appendix

\setcounter{table}{0}
\renewcommand{\thetable}{S.\arabic{table}}

\setcounter{figure}{0}
\renewcommand{\thefigure}{S.\arabic{figure}}

\setcounter{equation}{0}
\renewcommand{\theequation}{S.\arabic{equation}}

\centerline{\large{\textbf{Supplementary File}}}
\vspace{+0.3cm}

In this supplementary file, first, we show the five modified saliency methods and five aggregation approaches with which HINT can be implemented in Section \ref{sec:modified_saliency_methods} and \ref{sec:aggregation_approaches} respectively. Second, we explain the properties that HINT's Shapley value-based neuron contribution scoring approach satisfies in Section \ref{sec:properties_of_shapley_value-based_neuron_contribution_scoring_approach}. Third, we provide detailed descriptions of applications of HINT -- saliency method evaluation, explaining adversarial attack, and evaluation of COVID19 classification models -- in Section \ref{sec:other_applications}. Next, we demonstrate more neuron-concept associations and the activation maps of multimodal neurons in Section \ref{sec:responsible_neurons_to_hierarchical_concepts}. Then, we show more quantitative analysis and illustrations of the results of applying HINT for Weakly Supervised Object Localization tasks in Section \ref{sec:wsol}. Finally, we provide more illustrations of ablation studies on modified saliency methods and Shapley value-based scoring approach in Section \ref{sec:ablation_study}.

\section{Modified Saliency Methods}
\label{sec:modified_saliency_methods}

\begin{table*}
 \caption{Modified saliency methods and aggregation approaches}
  \centering
  \begin{tabular}{llll}
    \toprule
    \multicolumn{2}{c}{Modified saliency methods $\Lambda$ on the $l^{th}$ layer with respect to concept $e$} & \multicolumn{2}{c}{Aggregation approaches $\zeta$} \\
    \cmidrule(lr){1-2}
    \cmidrule(lr){3-4}
    Vanilla Backpropagation \cite{simonyan2013deep} & \(\textstyle \frac {\partial f^{e}(\mbox{\boldmath$x$})} {\partial \mbox{\boldmath$z$}}\) & Norm & \(\textstyle \| \mbox{\boldmath$s$} \| \) \\[0.3cm]
    Gradient x Input \cite{shrikumar2016not} & \(\textstyle \mbox{\boldmath$z$} \odot \frac {\partial f^{e}(\mbox{\boldmath$x$})} {\partial \mbox{\boldmath$z$}}\) & Filter norm & \(\textstyle \left \| \mbox{\boldmath$s$} > 0 \odot \mbox{\boldmath$s$} \right \| \) \\[0.3cm]
    Guided Backpropagation \cite{springenberg2014striving} & \(\textstyle \left(\frac {\partial f^{e}(\mbox{\boldmath$x$})} {\partial \mbox{\boldmath$z$}}\right)_{l+1} > 0 \odot \frac {\partial f^{e}(\mbox{\boldmath$x$})} {\partial \mbox{\boldmath$z$}}\) & Max & \(\textstyle \max(\mbox{\boldmath$s$}) \) \\[0.3cm]
    Integrated Gradient \cite{sundararajan2017axiomatic} & \(\textstyle f_{l}(\mbox{\boldmath$x$} - \bar{\mbox{\boldmath$x$}}) \odot \int_0^1 \frac {\partial f^{e}(\mbox{\boldmath$x$} + \alpha(\mbox{\boldmath$x$} - \bar{\mbox{\boldmath$x$}}))} {\partial \mbox{\boldmath$z$}} \ d \alpha \) & Abs max & \(\textstyle \max(|\mbox{\boldmath$s$}|) \) \\[0.3cm]
    SmoothGrad \cite{smilkov2017smoothgrad} & \(\textstyle \frac {1} {N} \sum_{n=1}^{N} \frac {\partial f^{e}(\mbox{\boldmath$x$}^{'})} {\partial \mbox{\boldmath$z$}^{'}}, \mbox{\boldmath$x$}^{'} = \mbox{\boldmath$x$} + \mathcal{N}(\mu, \sigma^{2}_{n})\) & Abs sum & \(\textstyle \sum(|\mbox{\boldmath$s$}|) \) \\
    \bottomrule
  \end{tabular}
  \label{tab:modified_saliency_methods_and_aggregation_approaches}
\end{table*}

Inspired by backpropagation-based saliency methods, we develop a saliency-guided approach to identify responsible regions in feature map $\mbox{\boldmath$z$}$. Equation \eqref{equ:org_saliency} shows how the representative backpropagation-based saliency method, Gradient (Vanilla Backpropagation) \cite{simonyan2013deep}, calculates the contribution of pixel $\mbox{\boldmath$x$}_{:,i_{0},j_{0}}$ to a class $C_{k}$.  

\begin{equation}
\frac {\partial f^{C_{k}}(\mbox{\boldmath$x$})} {\partial \mbox{\boldmath$x$}_{:,i_{0},j_{0}}}
\label{equ:org_saliency}
\end{equation}
where $f$ is a deep network, $f^{C_{k}}(\mbox{\boldmath$x$})$ is the logit of $\mbox{\boldmath$x$}$ to class $C_{k}$, and $\mbox{\boldmath$x$}_{:,i_{0},j_{0}}$ is a pixel.

We extend the idea of saliency maps to hidden layers. We take concept $e$ and neurons $\mathcal{D}$ on the $l^{th}$ layer as an example. Given an image $\mbox{\boldmath$x$}$ with label $C_{k}$ where $C_{k}$ is concept $e$ or a subcategory of concept $e$, the contribution of spatial activation $\mbox{\boldmath$z$}_{\mathcal{D},i_{l},j_{l}}$ to class $C_{k}$ (also to concept $e$) is shown in Equation \eqref{equ:saliency_on_hidden_layers}

\begin{equation}
\mbox{\boldmath$s$}_{\mathcal{D},i_{l},j_{l}} = \frac {\partial f^{C_{k}}(\mbox{\boldmath$z$})} {\partial \mbox{\boldmath$z$}_{\mathcal{D},i_{l},j_{l}}}
\label{equ:saliency_on_hidden_layers}
\end{equation}
where $\mbox{\boldmath$s$}_{\mathcal{D},i_{l},j_{l}} \in \mathbb{R}^{|\mathcal{D}|}$ is a vector and $\mbox{\boldmath$s$}_{\mathcal{D},i_{l},j_{l}}$s for each $i_{l}$ and $j_{l}$ form the saliency map $\mbox{\boldmath$s$}$.

As shown in Table \ref{tab:modified_saliency_methods_and_aggregation_approaches}, we modify five backpropagation-based saliency methods. All of them can be used in HINT.

\section{Aggregation Approaches}
\label{sec:aggregation_approaches}

With saliency map $\mbox{\boldmath$s$}$, the next step is to aggregate $\mbox{\boldmath$s$}_{\mathcal{D},i_{l},j_{l}}$, and the aggregated value will be used to decide whether $\mbox{\boldmath$z$}_{\mathcal{D},i_{l},j_{l}}$s belong to responsible foreground regions or irrelevant background regions. We implement five aggregation approaches shown in Table \ref{tab:modified_saliency_methods_and_aggregation_approaches}. All of them can be applied to HINT. Note that the aggregation is only conducted along the first dimension of $\mbox{\boldmath$s$}$.

\section{Properties of HINT's Shapley Value-based Neuron Contribution Scoring Approach}
\label{sec:properties_of_shapley_value-based_neuron_contribution_scoring_approach}

In the main paper, the Shapley value $\phi$ of a neuron $d$ to a concept $e$ is calculated as Equation \eqref{equ:shapley_value_calculation_suppl}.

\begin{equation}
\phi = \frac{ \sum_{\mbox{\boldmath$r$}} \left| \sum_{i=1}^{M} \left( L_{e}^{\langle \mathcal{S} \cup d \rangle}(r) - L_{e}^{\langle \mathcal{S} \rangle}(r) \right) \right| } {M|\mbox{\boldmath$r$}_{\mathcal{E}} \cup \mbox{\boldmath$r$}_{b^{*}}|}
\label{equ:shapley_value_calculation_suppl}
\end{equation}
where $\mathcal{D}$ is the set of neurons; $L_{e}$ is the classifier for concept $e$; $r = z_{\mathcal{D},i,j}$ represents spatial activation; $\mbox{\boldmath$r$}_{\mathcal{E}}$ and $\mbox{\boldmath$r$}_{b^{*}}$ are responsible regions of all concept $e \in \mathcal{E}$ and background regions; $\textstyle \mathcal{S} \subseteq \mathcal{D} \backslash d$ is the neuron subset randomly selected at each iteration; $\textstyle \langle * \rangle$ is an operator keeping the neurons in the brackets, \ie, $\mathcal{S} \cup d$ or $\mathcal{S}$, unchanged while randomizing others; $M$ is the number of iterations of Monte-Carlo sampling; $L_{e}^{\langle * \rangle}$ means that the classifier is re-trained with neurons in the brackets unchanged and others being randomized.

The following explains the properties of efficiency, symmetry, dummy, and additivity that Shapley values satisfy \cite{molnar2020interpretable}, \ie, our Shapley value-based scoring approach satisfies.

\paragraph{Efficiency.} The sum of neuron contributions should be equal to the difference between the prediction for $r$ and its expectation as shown in Equation \eqref{equ:shap_efficiency}.

\begin{equation}
\sum_{\mathcal{D}} \phi = \frac { \sum_{\mbox{\boldmath$r$}} \left( L_{e}(r) - E(L_{e}(r)) \right) } {|\mbox{\boldmath$r$}_{\mathcal{E}} \cup \mbox{\boldmath$r$}_{b^{*}}|}
\label{equ:shap_efficiency}
\end{equation}

\paragraph{Symmetry.} The contribution scores of neuron $d_{n}$ and $d_{m}$ should be the same if they contribute equally to concept $e$.

If
\begin{equation}
L_{e}^{\langle \mathcal{S} \cup d_{n} \rangle}(r) = L_{e}^{\langle \mathcal{S} \cup d_{m} \rangle}(r), \forall \mathcal{S} \subseteq \mathcal{D} \backslash \{d_{n}, d_{m}\}
\label{equ:shap_symmetry_1}
\end{equation}

Then
\begin{equation}
\phi_{d_{n}} = \phi_{d_{m}}
\label{equ:shap_symmetry_2}
\end{equation}
where $\textstyle \langle * \rangle$ is an operator keeping the neurons in the brackets, \ie, $\mathcal{S} \cup d_{n}$ or $\mathcal{S} \cup d_{m}$, unchanged while randomizing others.

\paragraph{Dummy.} If a neuron $d$ has no contribution to concept $e$, which means $d$'s individual contribution is zero and $d$ also has no contribution when it collaborates with other neurons, $d$'s contribution score should be zero.

If
\begin{equation}
L_{e}^{\langle \mathcal{S} \cup d \rangle}(r) = L_{e}^{\langle \mathcal{S} \rangle}(r), \forall \mathcal{S} \subseteq \mathcal{D} \backslash d
\label{equ:shap_dummy_1}
\end{equation}

Then
\begin{equation}
\phi_{d} = 0
\label{equ:shap_dummy_2}
\end{equation}

\paragraph{Additivity.} If $L_{e}$ is a random forest including different decision trees, the Shapley value of neuron $d$ of the random forest is the sum of the Shapley value of neuron $d$ of each decision tree.

\begin{equation}
\phi_{d} = \sum_{t=1}^{T} \phi_{d}^{t}
\label{equ:shap_additivity}
\end{equation}
where there are $T$ decision trees.

\section{Other Applications}
\label{sec:other_applications}

We demonstrate more applications of HINT as follows.

\begin{figure*}
  \centering
  \includegraphics[width=17cm]{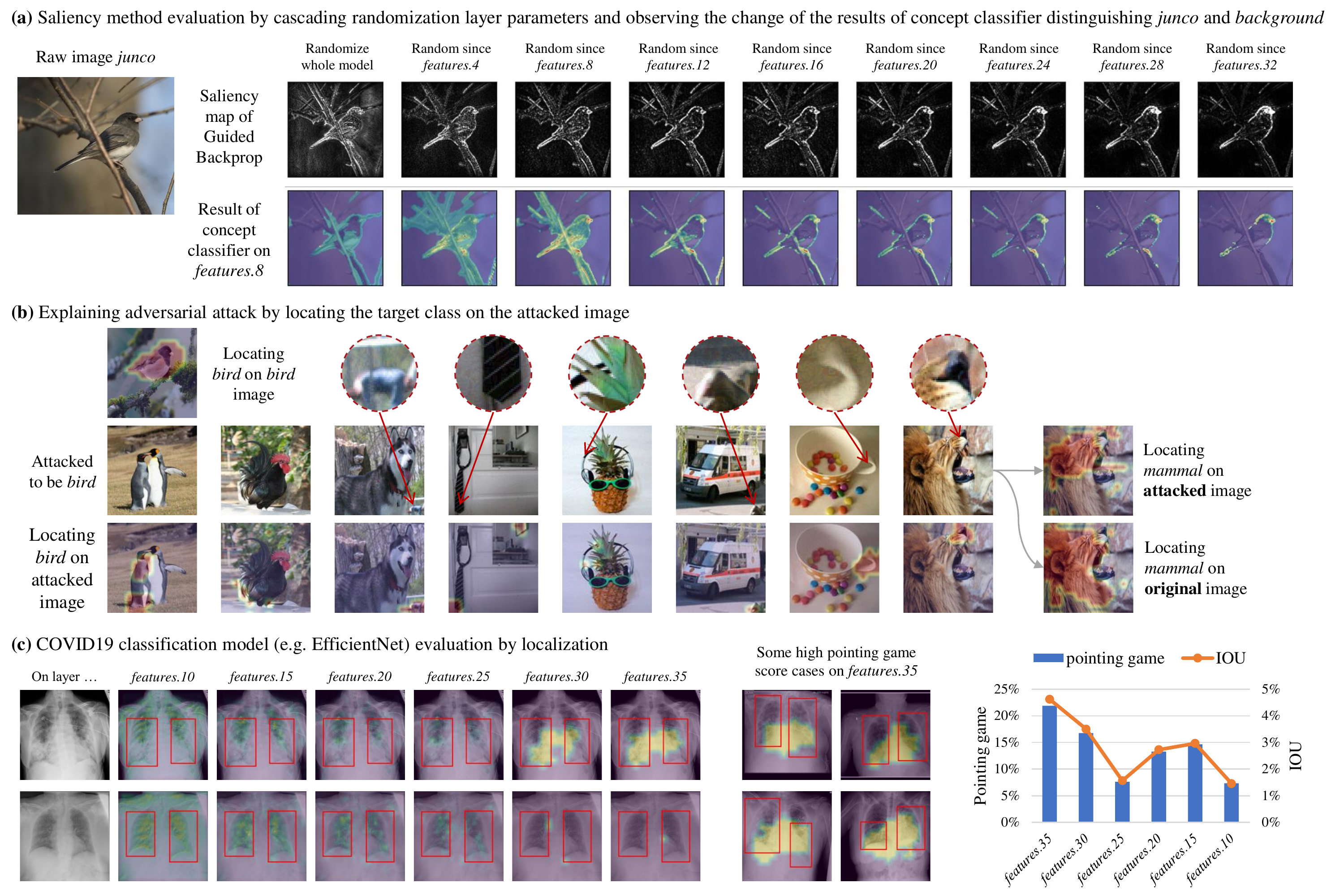}
    \caption{Other applications of HINT. (a) Saliency method evaluation. See Section \ref{sec:saliency_method_evaluation}. (b) Explaining adversarial attack. See Section \ref{sec:explaining_adversarial_attack}. (c) Evaluation of COVID19 classification model. See Section \ref{sec:COVID19_classification_model_evaluation}.}
    \label{fig:other_applications}
\end{figure*}

\subsection{Saliency Method Evaluation}
\label{sec:saliency_method_evaluation}

With the emergence of various saliency methods, different sanity evaluation approaches have been proposed \cite{adebayo2018sanity, yeh2019fidelity, kindermans2019reliability}. However, as most saliency methods only show responsible pixels on the input images, feature maps on hidden layers are not considered, which makes the sanity evaluation not comprehensive enough. For example, \cite{adebayo2018sanity} proposed a sanity test by comparing the saliency map before and after cascading randomization of model parameters from the top to the bottom layers. Guided Backpropagation failed the test because its results remained invariant.

We propose to apply the concept classifier implemented with the target saliency method to identify the responsible regions on hidden layer feature maps for the sanity test. The target saliency method passes the sanity test if meaningful responsible regions can be observed. As shown in Figure \ref{fig:other_applications} (a), on the hidden layer features.8, when fewer layers are randomized, the responsible regions are more focused on the key features of the bird -- its beak and tail, which means that Guided Backpropagation does reveal the salient region and Guided Backpropagation could pass the sanity test if hidden layer results are considered.

\subsection{Explaining Adversarial Attack}
\label{sec:explaining_adversarial_attack}

Concept classifiers can also be applied to explain how the object in an adversarial attacked image is shifted to be another class. As shown in Figure \ref{fig:other_applications} (b), we attack images of various classes to be \emph{bird} using PGD \cite{madry2017towards} and apply the \emph{bird} classifier to the attacked images' feature maps. The responsible regions for concept \emph{bird} highlighted in those fake \emph{bird} images imply that adversarial attack does not change all the content of the original object to be another class but captures some details of the original image where there exist shapes similar to \emph{bird}. For example, in the image of a coffee mug where most shapes are round, adversarial attack catches the only pointed shape and attacks it to be like \emph{bird}. Additionally, we find the attacked image still preserves features of the original class. In Figure \ref{fig:other_applications} (b), the result of applying \emph{mammal} classifier on the attacked lion image shows the most parts of the lion face are highlighted, while the result of applying \emph{mammal} classifier on the original lion image shows a similar pattern.

\subsection{COVID19 Classification Model Evaluation}
\label{sec:COVID19_classification_model_evaluation}

Applying deep learning to the detection of COVID19 in chest radiographs has the potential to provide quick diagnosis and guide management in molecular test resource-limited situations. However, the robustness of those models remains unclear \cite{degrave2021ai, khakzar2021towards}. We do not know whether the model decisions rely on confounding factors or medical pathology in chest radiographs. To tackle the challenge, object localization by HINT can be used to see whether the identified responsible regions overlap with the lesion regions drawn by doctors. With the COVID19 dataset from SIIM-FISABIO-RSNA COVID-19 Detection competition \cite{lakhani20212021}, we trained models used by high-ranking teams and other baseline models for classification. The localization results of COVID19 cases with typical symptoms by EfficientNet \cite{tan2019efficientnet} are shown in Figure \ref{fig:other_applications} (c). As you can see, the pointing game and IoU are not high. Many cases having low pointing game and IoU values show that the model does not focus on the lesion region, while for the cases with high pointing game and IoU values, further investigation is still required to see whether they capture the medical pathology features or they just accidentally focus on the area of the stomach.

\begin{table}[ht]
 \caption{Pointing game (pointing) and IoU of the localization results of different models on the chest radiographs of COVID19 cases with typical symptoms.}
  \centering
  \begin{tabular}{llll}
    \toprule
    Model & Layer & pointing & IoU\\
    \cmidrule(lr){1-2} \cmidrule(lr){3-4}
    \rowcolor{gray!20} EfficientNet \cite{tan2019efficientnet} & features.35 & 21.8\% & 4.6\% \\
    DenseNet161 \cite{huang2017densely} & denseblock4 & 94.1\% & 18.2\% \\
    \rowcolor{gray!20} Inception v3 \cite{szegedy2016rethinking} & Mixed\_6c & 17.3\% & 3.2\% \\
    ResNet50 \cite{he2016deep} & layer3.3 & 15.7\% & 2.9\% \\
    \rowcolor{gray!20} ShuffleNet v2 \cite{ma2018shufflenet} & stage3.5 & 22.2\% & 3.8\% \\
    SqueezeNet1 \cite{iandola2016squeezenet} & features.9 & 0\% & 0\% \\
    \rowcolor{gray!20} VGG19 \cite{simonyan2014very} & features.40 & 9.9\% & 1.6\% \\
    \bottomrule
  \end{tabular}
  \label{tab:pointing game and iou of different covid19 classification models}
\end{table}

\begin{figure*}
  \centering
  \includegraphics[width=16cm]{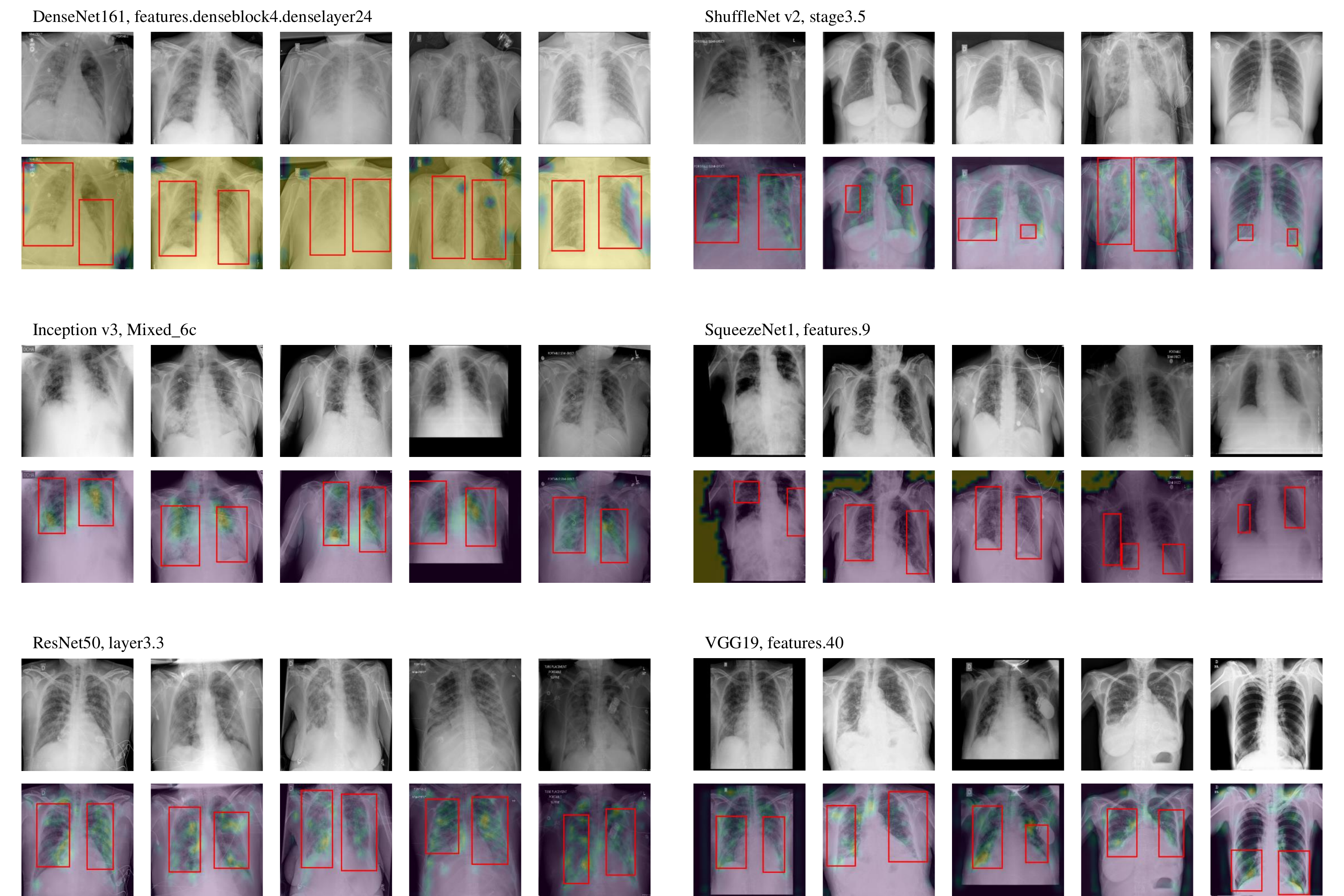}
    \caption{Localization results of different models on the radiographs of COVID19 cases with typical symptoms. The red bounding boxes are the lesion regions drawn by doctors.}
    \label{fig:covid19_dif_models}
\end{figure*}

Figure \ref{fig:covid19_dif_models} illustrates results of other models and Table \ref{tab:pointing game and iou of different covid19 classification models} quantitatively compares the different models by metrics of pointing game (pointing) and IoU. The accuracy values indicate that the hidden layers of SqueezeNet1 may fail to learn the concept of COVID19 pulmonary lesion. This can also be observed from Figure \ref{fig:covid19_dif_models} that SqueezeNet1 locates background regions. Note that although the pointing game score and IoU of DenseNet161 are very high, it is still possible that DenseNet161 fails to learn the concept of COVID19 pulmonary lesion as it highlights all the regions (see Figure \ref{fig:covid19_dif_models}).

\section{Identification of Responsible Neurons to Hierarchical Concepts}
\label{sec:responsible_neurons_to_hierarchical_concepts}

\begin{figure}[ht]
  \centering
  \includegraphics[width=7cm]{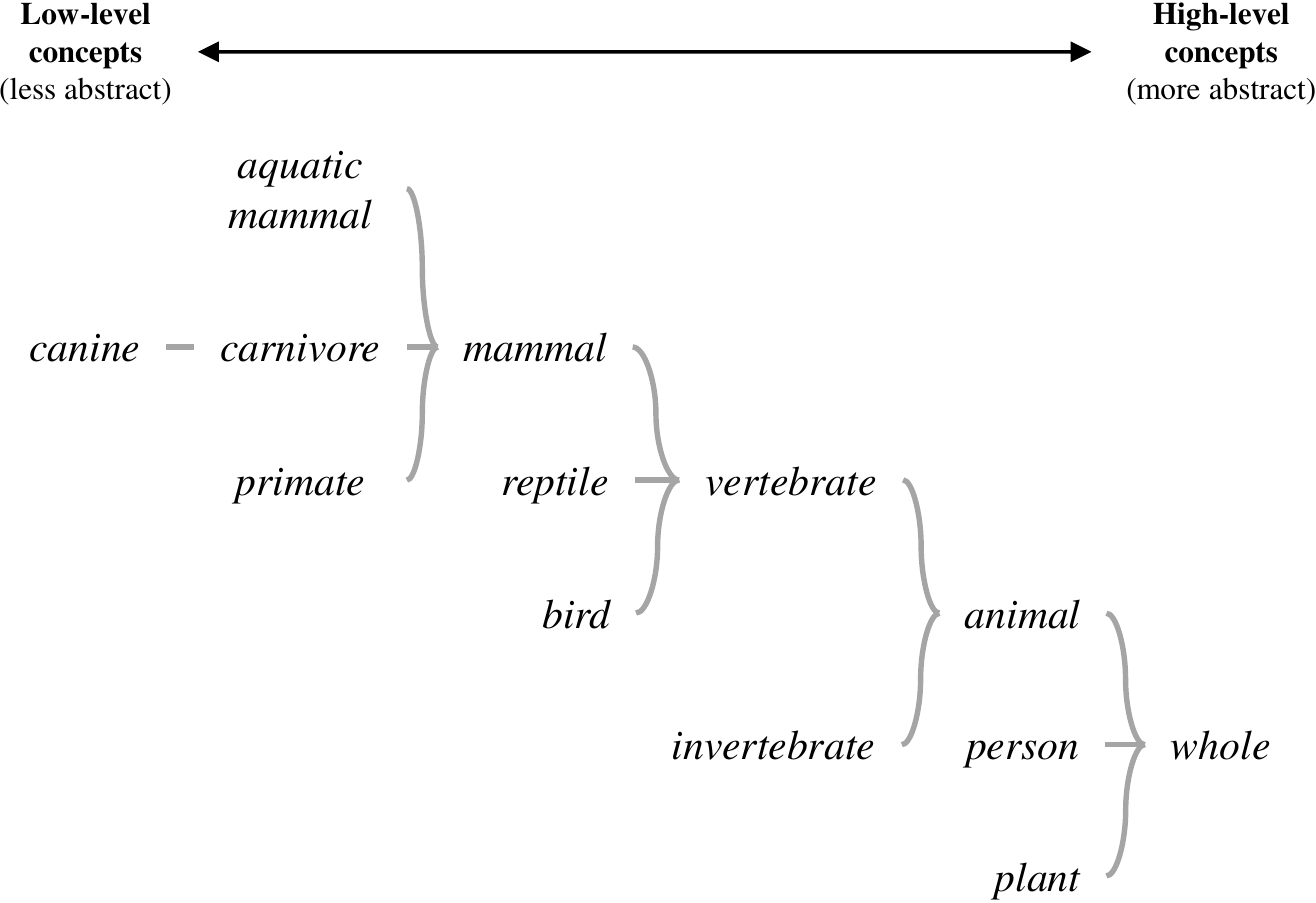}
    \caption{A hierarchy of concepts.}
    \label{fig:fig_supplementary_hierarchy}
\end{figure}

\subsection{More Neuron-concept Associations}

\begin{figure*}
  \centering
  \includegraphics[width=16cm]{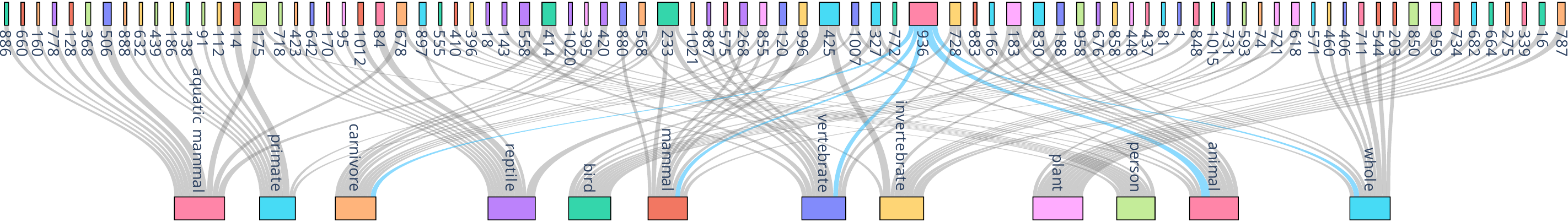}
    \caption{Top-10 responsible neurons to concepts on layer ``layer3.5'' of ResNet50.}
    \label{fig:responsible_neurons_to_concepts_ResNet50 layer3.5}
\end{figure*}

This section illustrates more associations between neurons and concepts. The Sankey diagram in Figure \ref{fig:responsible_neurons_to_concepts_vgg19 features.30 complete} shows the top-10 responsible neurons on layer features.30 of VGG19 to each concept in the hierarchy (see Figure \ref{fig:fig_supplementary_hierarchy}). And the Sankey diagram in Figure \ref{fig:responsible_neurons_to_concepts_ResNet50 layer3.5} shows the case on layer layer3.5 of ResNet50.

\begin{figure*}
  \centering
  \includegraphics[width=16cm]{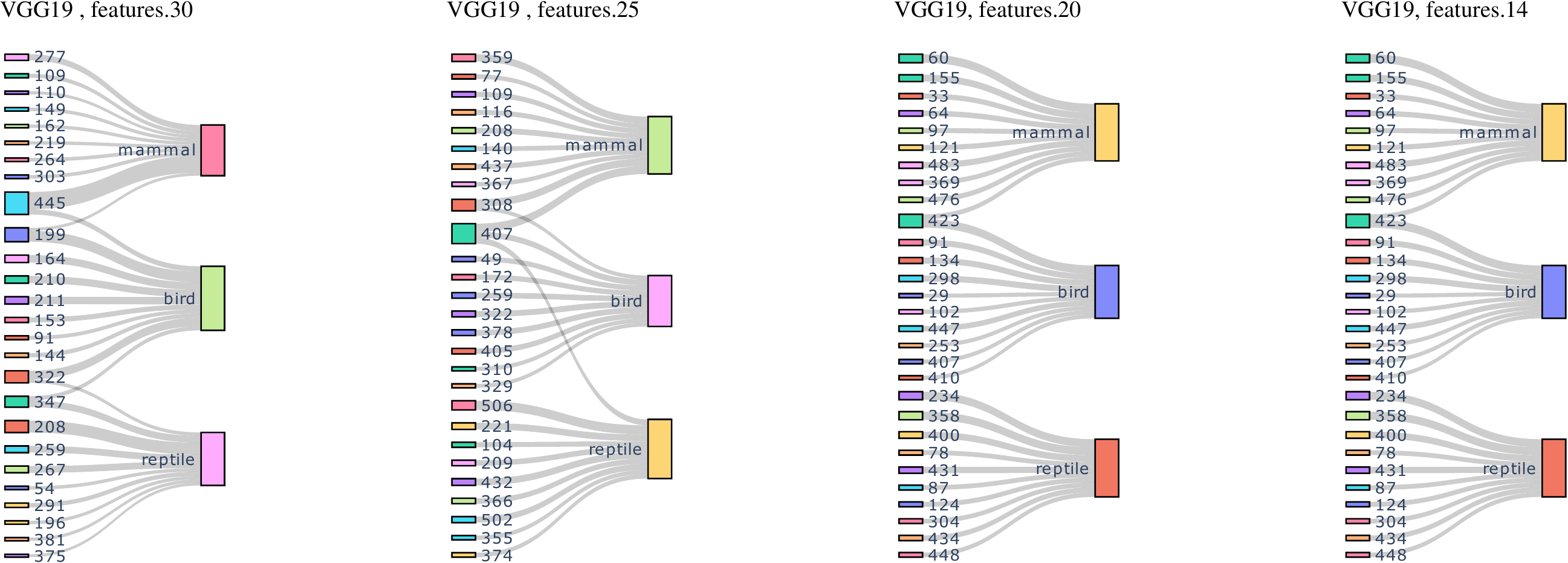}
    \caption{Top-10 responsible neurons to concepts of \emph{mammal}, \emph{bird}, and \emph{reptile} on different layer of VGG19.}
    \label{fig:responsible_neurons_to_concepts_vgg19 dif layer}
\end{figure*}

\paragraph{Different layers.} Figure \ref{fig:responsible_neurons_to_concepts_vgg19 dif layer} shows the top-10 responsible neurons on different layers on VGG19 to concepts of \emph{mammal}, \emph{bird}, and \emph{reptile}.

\begin{figure*}
  \centering
  \includegraphics[width=16cm]{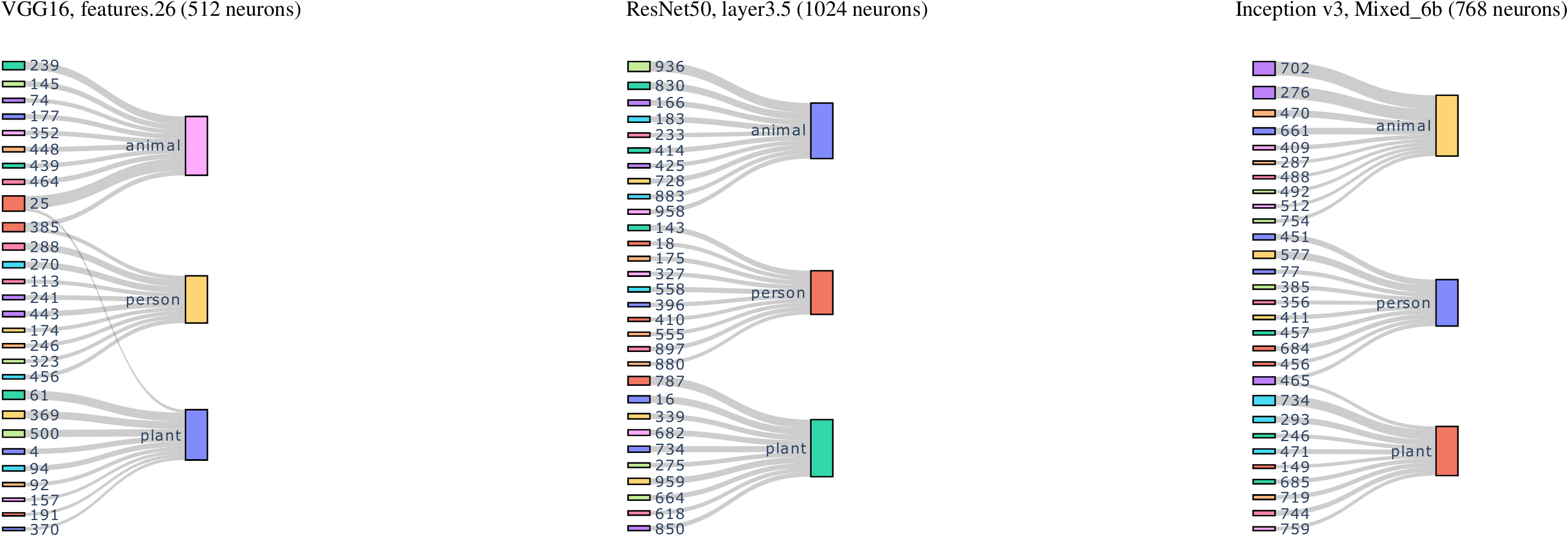}
    \caption{Top-10 responsible neurons to concepts of \emph{animal}, \emph{person}, and \emph{plant} of other models.}
    \label{fig:responsible_neurons_to_concepts_other models}
\end{figure*}

\paragraph{Different models.} Figure \ref{fig:responsible_neurons_to_concepts_other models} shows top-10 responsible neurons on layer features.26 of VGG16, layer3.5 of ResNet50, and Mixed\_6b of Inception v3 to concepts of \emph{animal}, \emph{person}, and \emph{plant}.

\subsection{Contribution Scores (Shapley Values) of Neurons to Concepts.}

\begin{figure*}
  \centering
  \includegraphics[width=16cm]{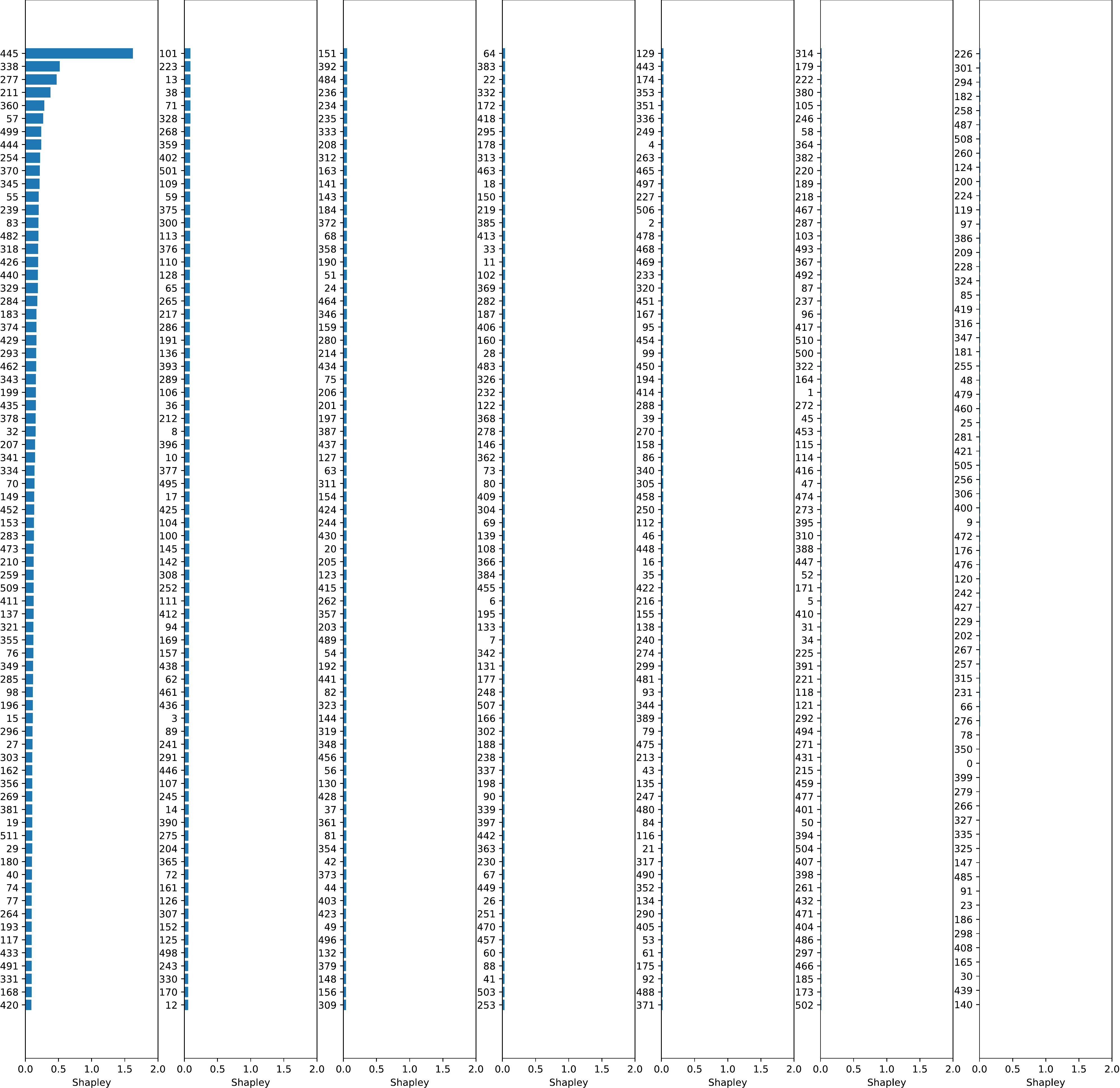}
    \caption{Contribution scores (Shapley values) of neurons on layer features.30 of VGG19 to the concept of \emph{carnivore}.}
    \label{fig:shap_bar_plot_vgg19_features.30_carnivore_classifier}
\end{figure*}

\paragraph{Concepts of different levels.} The bar charts in Figure \ref{fig:shap_bar_plot_vgg19_features.30_animal, animate being, beast, brute, creature, fauna_classifier}, \ref{fig:shap_bar_plot_vgg19_features.30_vertebrate, craniate_classifier}, \ref{fig:shap_bar_plot_vgg19_features.30_mammal, mammalian_classifier}, and \ref{fig:shap_bar_plot_vgg19_features.30_carnivore_classifier} show the contribution scores (Shapley values) of neurons on layer features.30 of VGG19 to concepts of \emph{animal}, \emph{vertebrate}, \emph{mammal}, and \emph{carnivore} respectively. As we can see, the $445^{th}$ neuron has the highest contribution to all the concepts.

\begin{figure*}
  \centering
  \includegraphics[width=16cm]{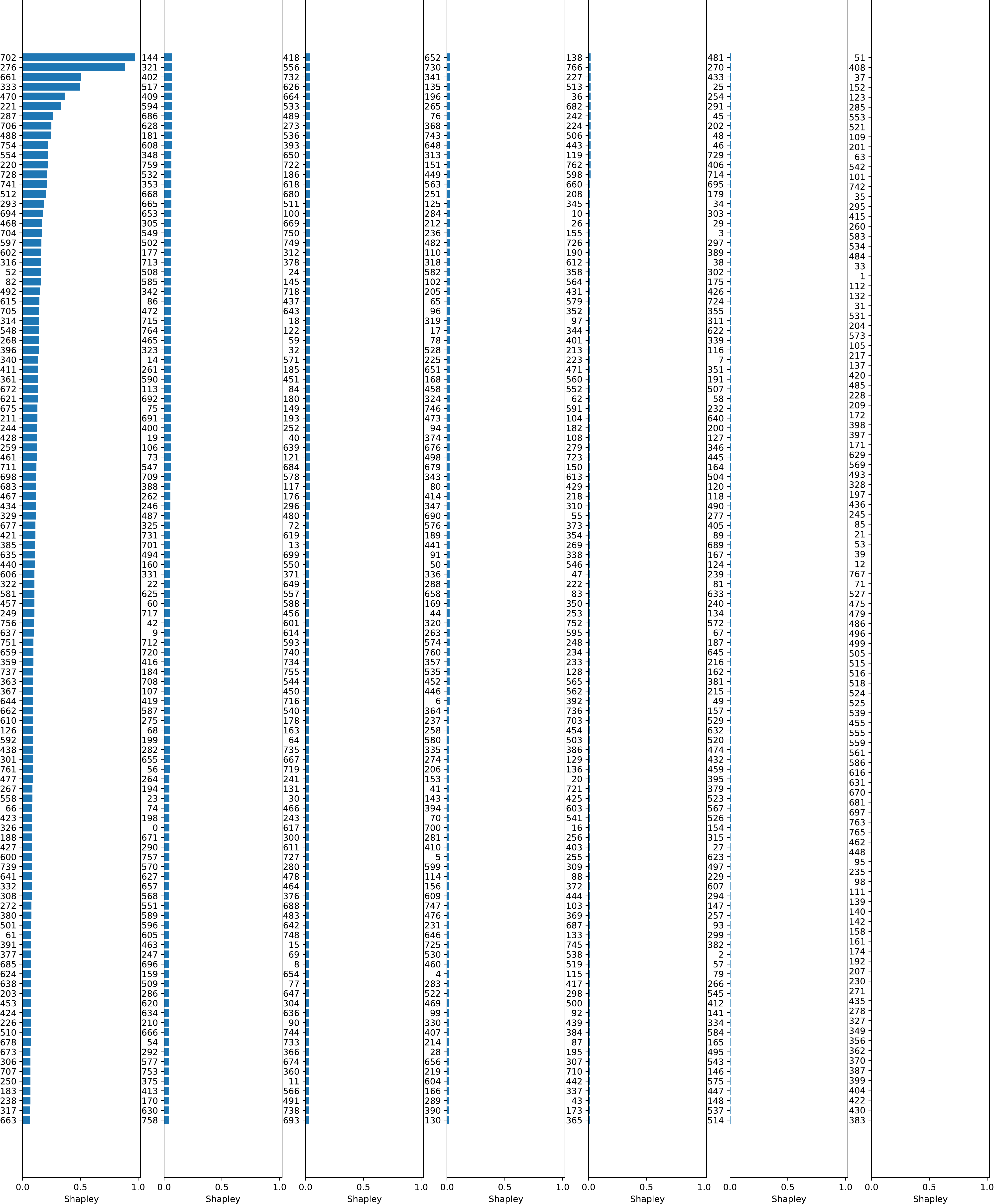}
    \caption{Contribution scores (Shapley values) of neurons on layer Mixed\_6b of Inception v3 to the concept of \emph{animal}.}
    \label{fig:shap_bar_plot_inception_v3_Mixed_6b_animal_classifier}
\end{figure*}

\begin{figure*}
  \centering
  \includegraphics[width=16cm]{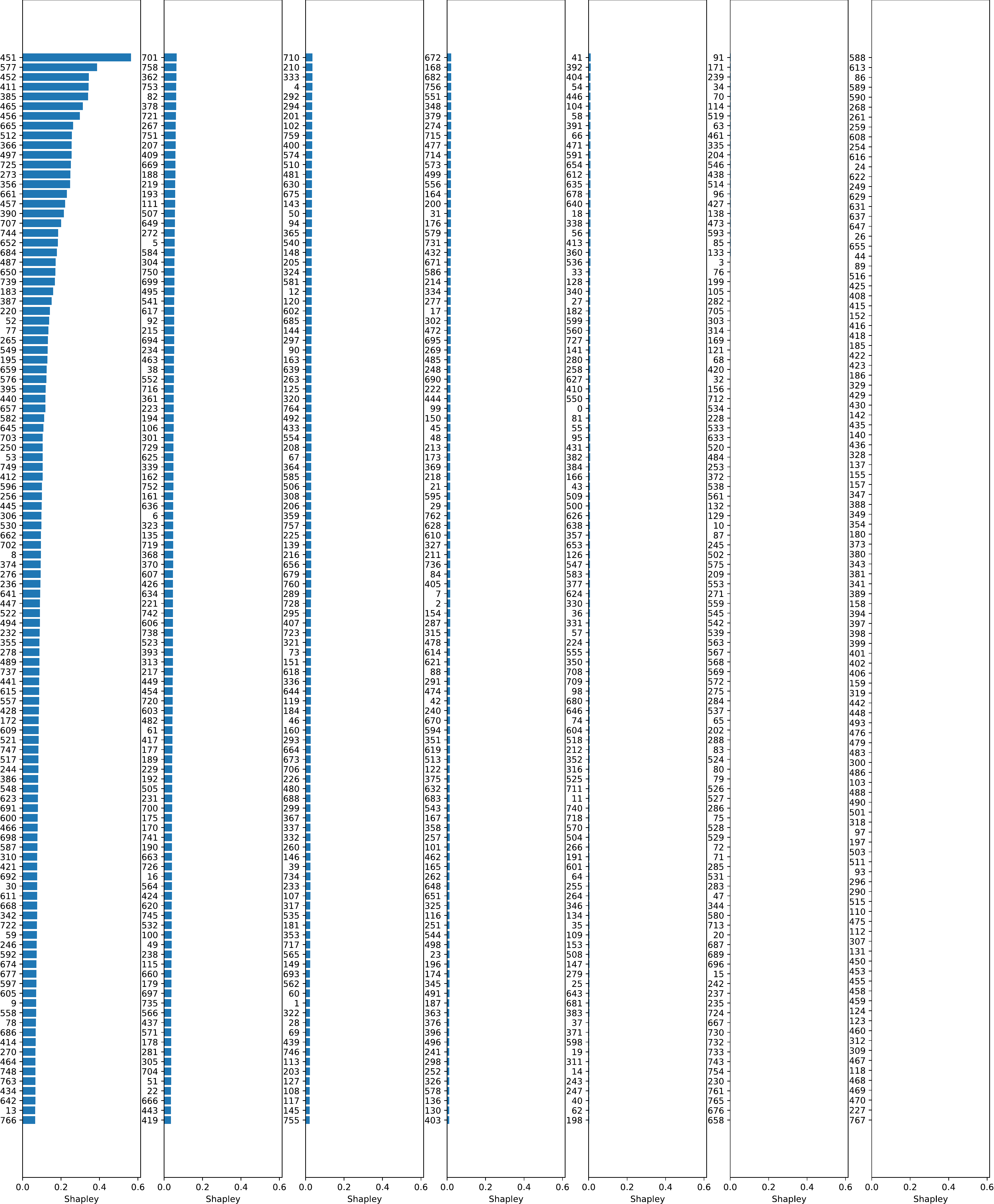}
    \caption{Contribution scores (Shapley values) of neurons on layer Mixed\_6b of Inception v3 to the concept of \emph{person}.}
    \label{fig:shap_bar_plot_inception_v3_Mixed_6b_person_classifier}
\end{figure*}

\begin{figure*}
  \centering
  \includegraphics[width=16cm]{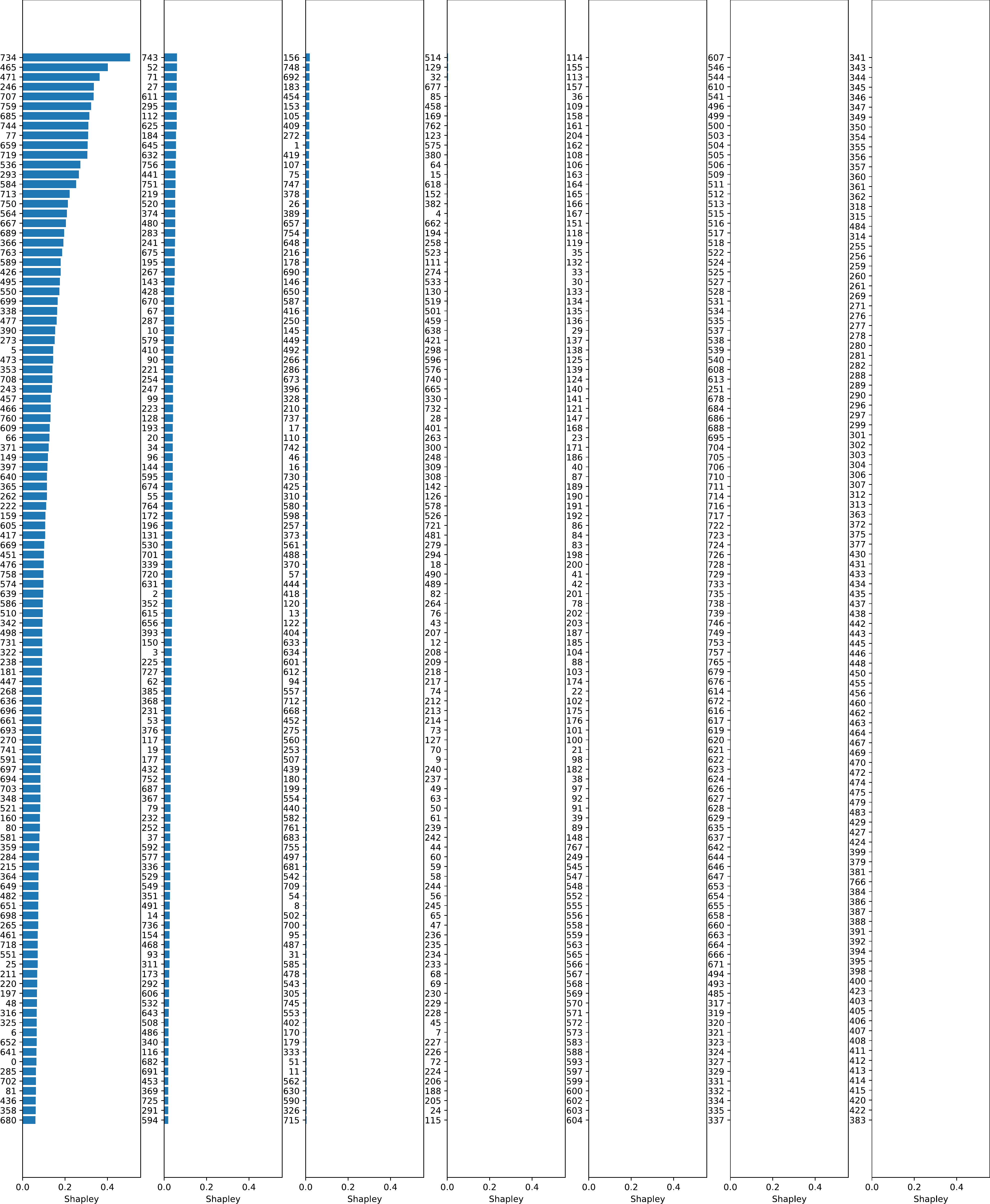}
    \caption{Contribution scores (Shapley values) of neurons on layer Mixed\_6b of Inception v3 to the concept of \emph{plant}.}
    \label{fig:shap_bar_plot_inception_v3_Mixed_6b_plant_classifier}
\end{figure*}

\paragraph{Concepts of the same level.} The bar charts in Figure \ref{fig:shap_bar_plot_inception_v3_Mixed_6b_animal_classifier}, \ref{fig:shap_bar_plot_inception_v3_Mixed_6b_person_classifier}, \ref{fig:shap_bar_plot_inception_v3_Mixed_6b_plant_classifier} show the contribution scores (Shapley values) of neurons on layer Mixed\_6b of Inception v3 to concepts of \emph{animal}, \emph{person}, and \emph{plant} respectively. There are 768 neurons on layer Mixed\_6b in total. For \emph{animal}, there are 711 neurons with contribution scores larger than zero. For \emph{person}, the number is 615. And for \emph{plant}, the number is 387. This indicates that there are less neurons responsible for \emph{plant}, which may reflect the bias of the training data that only few categories of \emph{plants} were included and \emph{plant} images only take a small percentage of the whole dataset.

\begin{figure*}
  \centering
  \includegraphics[width=16cm]{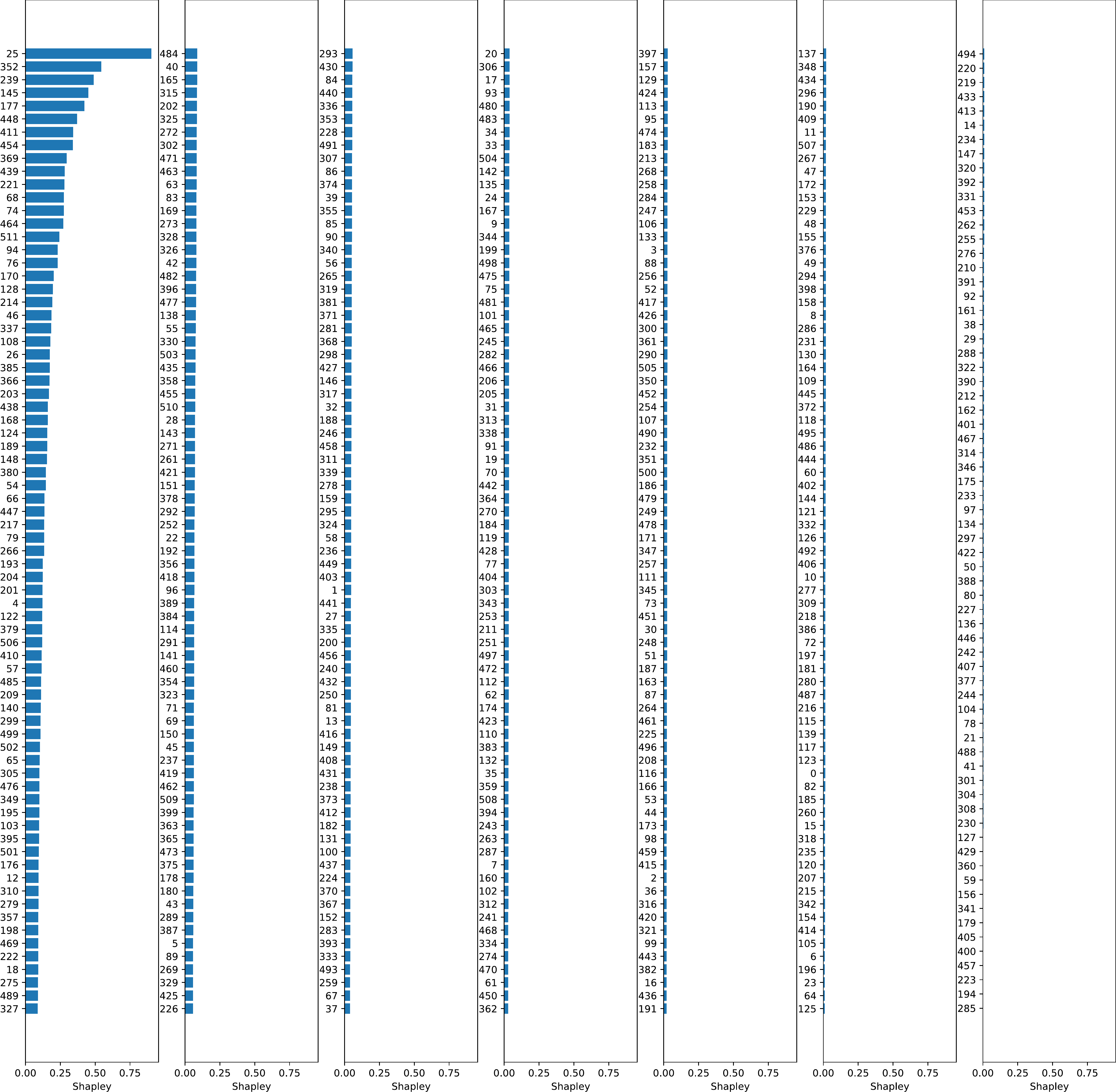}
    \caption{Contribution scores (Shapley values) of neurons on layer features.26 of VGG16 to the concept of \emph{animal}.}
    \label{fig:shap_bar_plot_vgg16_features.26_animal_classifier}
\end{figure*}

\begin{figure*}
  \centering
  \includegraphics[width=12cm]{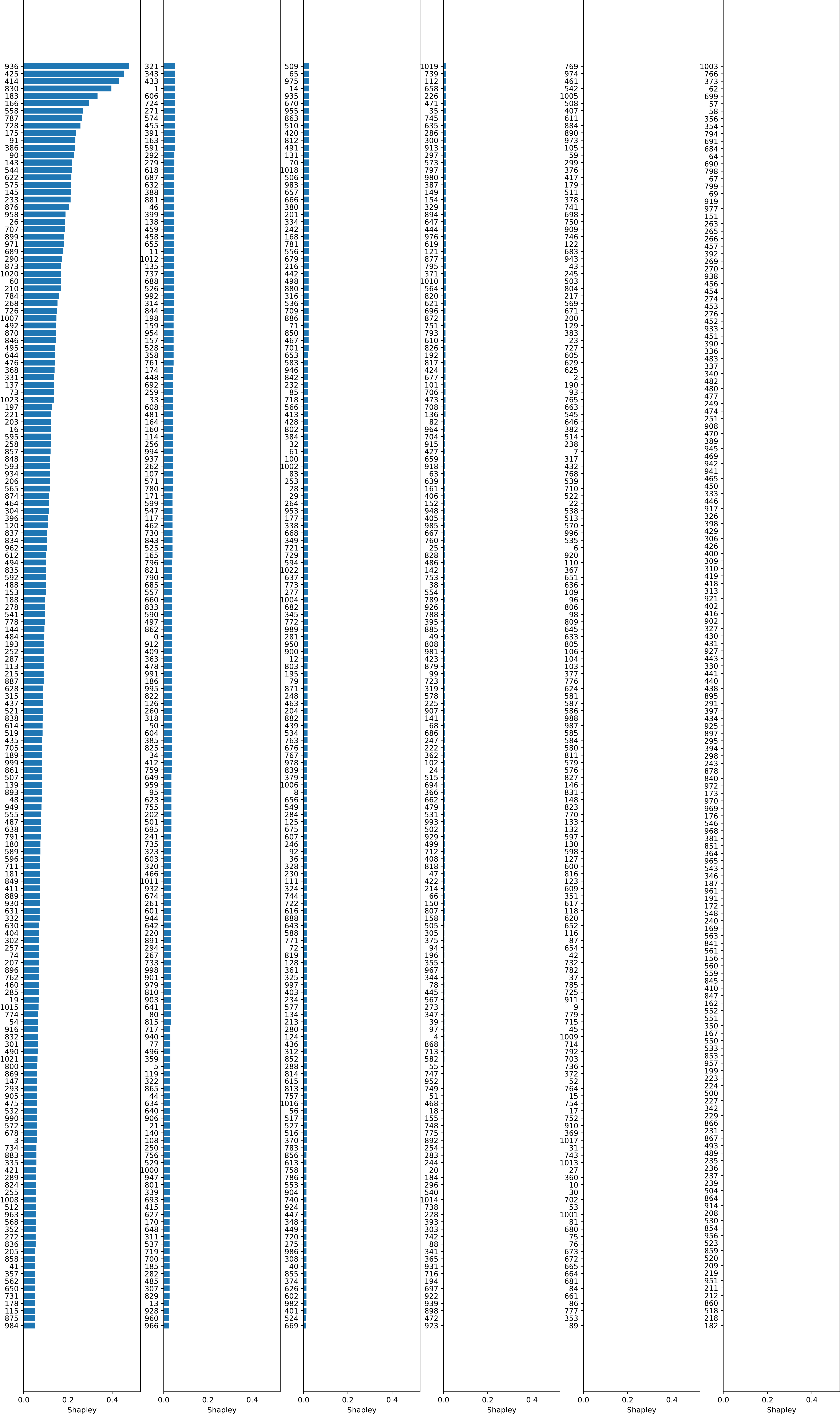}
    \caption{Contribution scores (Shapley values) of neurons on layer layer3.5 of ResNet50 to the concept of \emph{animal}.}
    \label{fig:shap_bar_plot_resnet50_layer3.5_animal_classifier}
\end{figure*}

\paragraph{Different models.} The bar charts in Figure \ref{fig:shap_bar_plot_vgg19_features.30_animal, animate being, beast, brute, creature, fauna_classifier}, \ref{fig:shap_bar_plot_vgg16_features.26_animal_classifier}, \ref{fig:shap_bar_plot_resnet50_layer3.5_animal_classifier}, and \ref{fig:shap_bar_plot_inception_v3_Mixed_6b_animal_classifier} show the the contribution scores (Shapley values) of neurons on different layers of VGG19, VGG16, ResNet50, and Inception v3 to the concept of \emph{animal} respectively. As we can see, the drop of the neurons' contribution scores of ResNet50 is less sharp compared with VGG16 and Inception v3, which means that the neurons of ResNet50 more rely on collaboration to detect \emph{animal}.

\subsection{Activation Maps of Multimodal Neurons}

\begin{figure*}
  \centering
  \includegraphics[width=16cm]{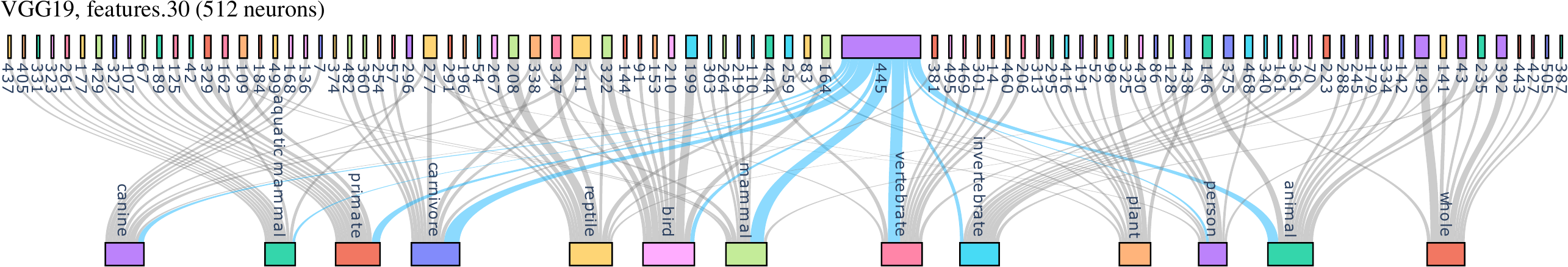}
    \caption{Top-10 responsible neurons to concepts on layer ``features.30'' of VGG19.}
    \label{fig:responsible_neurons_to_concepts_vgg19 features.30 complete}
\end{figure*}

As shown in \ref{fig:responsible_neurons_to_concepts_vgg19 features.30 complete}, the $445^{th}$ neuron on layer features.30 of VGG19 contribute strongly to multiple concepts, indicating it is multimodal. We show the activation maps of the $445^{th}$ neuron on images of \emph{animal} (see Figure \ref{fig:channel_activation_445th_channel_images_vgg19_animal, animate being, beast, brute, creature, fauna}), \emph{mammal} (see Figure \ref{fig:channel_activation_445th_channel_images_vgg19_mammal, mammalian}), and \emph{canine} (see Figure \ref{fig:channel_activation_445th_channel_images_vgg19_canine, canid}) respectively.

\begin{figure*}
  \centering
  \includegraphics[width=16cm]{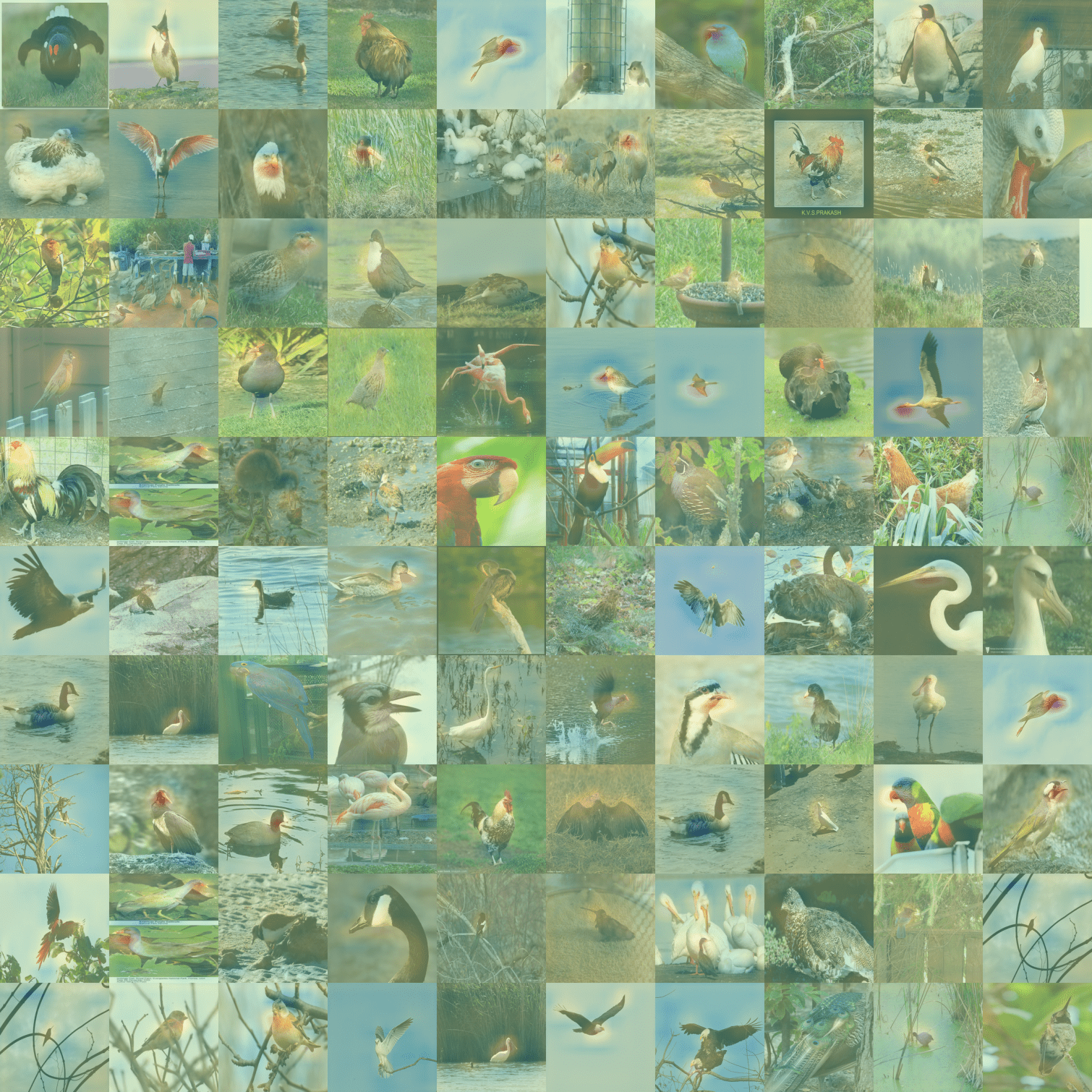}
    \caption{Activation map of the $199^{th}$ neuron on \emph{bird} images.}
    \label{fig:channel_activation_199th_channel_images_vgg19_bird}
\end{figure*}

Also, we show the activation maps of the $199^{th}$ neuron on layer features.30 of VGG19 which contributes strongly to both \emph{bird} and \emph{car} in Figure \ref{fig:channel_activation_199th_channel_images_vgg19_bird} and \ref{fig:channel_activation_199th_channel_images_vgg19_beach wagon, station wagon, wagon, estate car, beach waggon, station waggon, waggon}. The results indicate the $199^{th}$ neuron activates the head of \emph{bird} while deactivating the wheels of \emph{car}. Therefore, it is multimodal and can detect both \emph{bird} and \emph{car}.

\section{Weakly Supervised Object Localization}
\label{sec:wsol}

\subsection{Localization Accuracy on CUB-200-2011}

\begin{table}
 \caption{Comparison of Localization Accuracy on CUB-200-2011. * indicates fine-tuning on CUB-200-2011. "rand" indicates the neurons are randomly selected.}
  \centering
  \begin{tabular}{llll}
    \toprule
    & VGG16 & ResNet50 & Inception v3\\
    \cmidrule(lr){2-4}
    CAM* \cite{zhou2016learning} & 34.4\% & 42.7\% & 43.7\% \\
    ACoL* \cite{zhang2018adversarial} & 45.9\% & - & - \\
    SPG* \cite{zhang2018self} & - & - & 46.6\% \\
    ADL* \cite{choe2019attention} & 52.4\% & 62.3\% & 53.0\% \\
    DANet* \cite{xue2019danet} & 52.5\% & - & 49.5\% \\
    EIL* \cite{mai2020erasing} & 57.5\% & - & - \\
    PSOL* \cite{zhang2020rethinking} & 66.3\% & 70.7\% & 65.5\% \\
    GCNet* \cite{lu2020geometry} & 63.2\% & - & - \\
    RCAM* \cite{bae2020rethinking} & 59.0\% & 59.5\% & - \\
    FAM* \cite{meng2021foreground} & \textbf{69.3}\% & \textbf{73.7}\% & \textbf{70.7}\% \\
    \rowcolor{gray!20} \textbf{Ours} (10\%) & \textbf{66.6}\% & 60.2\% & 49.0\% \\
    \textbf{Ours} (10\%, rand) & 56.2\% & 4.7\% & 14.2\% \\
    \rowcolor{gray!20} \textbf{Ours} (20\%) & 65.2\% & 67.1\% & 55.8\% \\
    \textbf{Ours} (20\%, rand) & 58.4\% & 35.9\% & 34.2\% \\
    \rowcolor{gray!20} \textbf{Ours} (40\%) & 61.3\% & 77.3\% & 52.8\% \\
    \textbf{Ours} (40\%, rand) & 60.5\% & 68.6\% & 48.1\% \\
    \rowcolor{gray!20} \textbf{Ours} (80\%) & 64.8\% & \textbf{80.2}\% & \textbf{56.2}\% \\
    \textbf{Ours} (80\%, rand) & 61.5\% & 76.5\% & 53.0\% \\
    \bottomrule
  \end{tabular}
  \label{tab:comparison_of_localization_on_CUB-200-2011_suppl}
\end{table}

As shown in Table \ref{tab:comparison_of_localization_on_CUB-200-2011_suppl}, the localization accuracy of HINT is compared with existing methods on the CUB-200-2011 \cite{wah2011caltech} dataset. We train \emph{animal} classifiers with 10\%, 20\%, 40\%, 80\% neurons sorted and selected by Shapley values using different models. Besides, we add a baseline tests of HINT where the neurons are randomly chosen. The results verify that Shapley values are good measurements of neuron contributions and show that different models might have different learning modes: ResNet50 and Inception v3 rely more on neurons' collaboration while neurons in VGG16 work more independently. This can be observed from the Localization Accuracy values. The Localization Accuracy of ResNet50 and Inception v3 increase steadily when more neurons are included in the concept classifier while the Localization Accuracy of VGG16 only has minor increase when more neurons are added.

\subsection{Quantitative Results of Applying Concept Classifiers on ImageNet}

In this section, because many images in ImageNet only have classification labels, we use the hidden layer saliency map as the mask of the target object. And we apply metrics of pointing game (pointing) \cite{zhang2018top}, Spearman's correlation (spearman cor), and structure similarity index (SSMI) \cite{wang2009mean} to evaluate concept classifiers' performances on ImageNet. VGG19 is used for testing.

\begin{table}[ht]
 \caption{Apply \emph{whole} classifier trained on layer features.30 to images of different concepts.}
  \centering
  \begin{tabular}{llll}
    \toprule
    Images of & pointing & spearman cor & SSMI\\
    \cmidrule(lr){2-4}
    \rowcolor{gray!20} \emph{whole} & 88.0\% & 52.2\% & 34.4\% \\
    \emph{person} & 34.0\% & 32.0\% & 26.5\% \\
    \rowcolor{gray!20} \emph{plant} & 60.4\% & 37.9\% & 24.6\% \\
    \emph{animal} & 81.9\% & 62.8\% & 38.1\% \\
    \rowcolor{gray!20} \emph{mammal} & 77.7\% & 63.4\% & 43.5\% \\
    \emph{bird} & 86.7\% & 60.3\% & 44.1\% \\
    \rowcolor{gray!20} \emph{reptile} & 68.5\% & 56.3\% & 35.8\% \\
    \emph{carnivore} & 82.2\% & 68.3\% & 42.4\% \\
    \rowcolor{gray!20} \emph{primate} & 82.6\% & 53.7\% & 36.9\% \\
    \emph{aquatic mammal} & 56.9\% & 57.0\% & 43.5\% \\
    \bottomrule
  \end{tabular}
  \label{tab:apply whole classifier trained on layer features.30 to images of different concepts}
\end{table}

\paragraph{Images of different concepts.} As shown in Table \ref{tab:apply whole classifier trained on layer features.30 to images of different concepts}, we apply \emph{whole} classifier trained on layer features.30 to images of different concepts. The results indicate that the \emph{whole} classifier can locate all the target objects as the concepts are all subcategories of \emph{whole}. Also, we test the \emph{mammal} classifier to images of other concepts which have no intersection with \emph{mammal}, showing that the \emph{mammal} classifier only responses to image contents of \emph{mammal} (see Table \ref{tab:apply mammal classifier trained on layer features.30 to person and plant images}).

\begin{table}[ht]
 \caption{Apply \textbf{\emph{mammal} classifier} trained on layer features.30 to \textbf{\emph{person}} and \textbf{\emph{plant}} images.}
  \centering
  \begin{tabular}{llll}
    \toprule
    Images of & pointing & spearman cor & SSMI\\
    \cmidrule(lr){2-4}
    \rowcolor{gray!20} \emph{person} & 8.8\% & 6.4\% & 8.6\% \\
    \emph{plant} & 3.6\% & 9.3\% & 0.9\% \\
    \bottomrule
  \end{tabular}
  \label{tab:apply mammal classifier trained on layer features.30 to person and plant images}
\end{table}

\paragraph{Different layers.} As shown in Table \ref{tab:apply mammal classifier trained on different layers to mammal images}, we apply \emph{mammal} classifier trained on different layers to \emph{mammal} images. The accuracy values increase as the layer goes higher, indicating the network can learn abstract concepts such as \emph{mammal} on high layers.

\begin{table}[ht]
 \caption{Apply \textbf{\emph{mammal} classifier} trained on different layers to \textbf{\emph{mammal}} images.}
  \centering
  \begin{tabular}{llll}
    \toprule
    Layer & pointing & spearman cor & SSMI\\
    \cmidrule(lr){2-4}
    \rowcolor{gray!20} features.2 & 11.7\% & 4.9\% & 3.7\% \\
    features.7 & 13.0\% & 13.7\% & 6.1\% \\
    \rowcolor{gray!20} features.10 & 28.7\% & 30.5\% & 8.9\% \\
    features.14 & 35.1\% & 34.5\% & 9.7\% \\
    \rowcolor{gray!20} features.20 & 58.4\% & 45.3\% & 15.4\% \\
    features.25 & 67.8\% & 51.7\% & 25.3\% \\
    \rowcolor{gray!20} features.30 & 76.4\% & 59.8\% & 37.7\% \\
    \bottomrule
  \end{tabular}
  \label{tab:apply mammal classifier trained on different layers to mammal images}
\end{table}

\subsection{Visualizations of Localization Results on ImageNet, CUB-200-2011, and PASCAL VOC}

\begin{figure*}
  \centering
  \includegraphics[width=16cm]{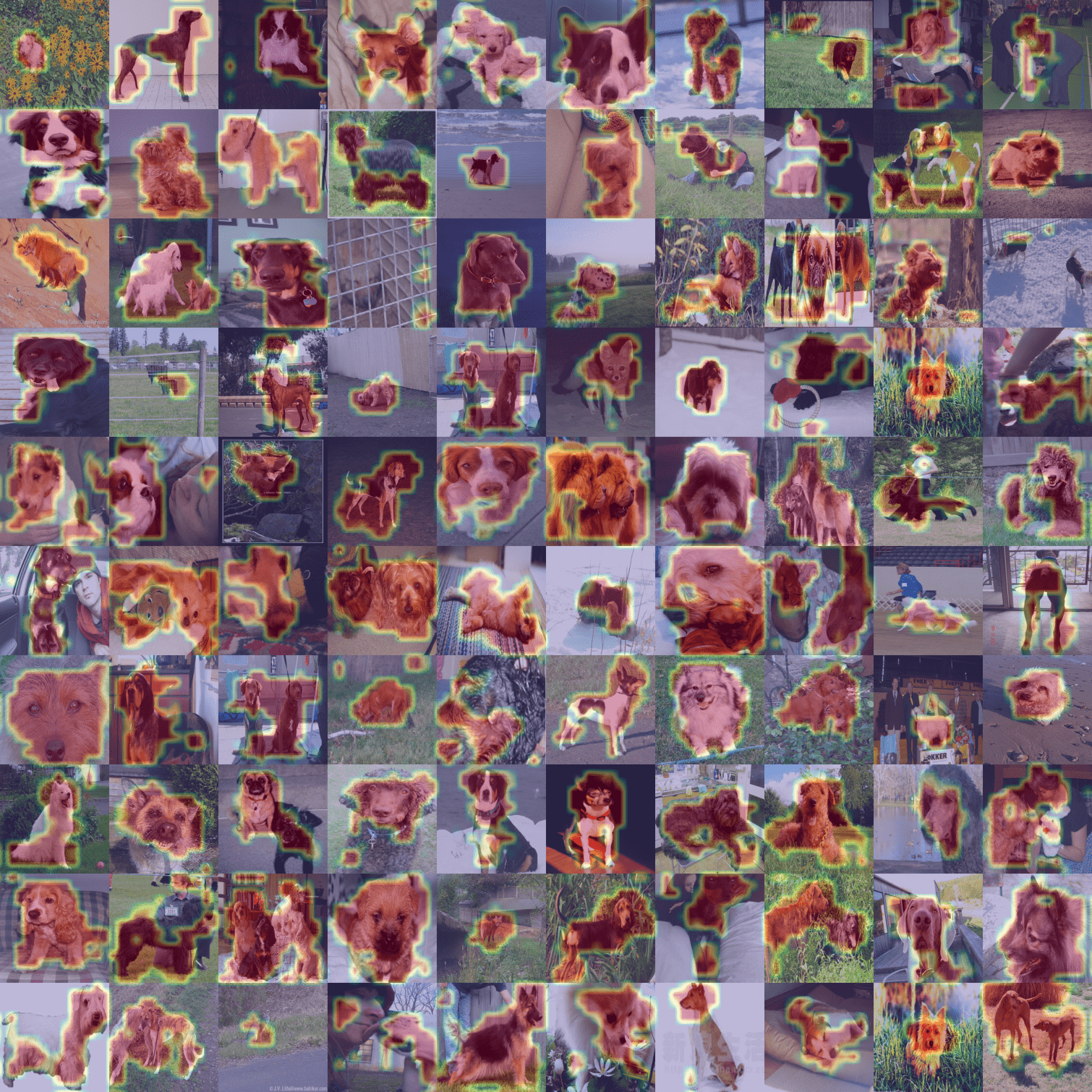}
    \caption{Localization results of applying \emph{mammal} classifier on the images containing the concept of \emph{canine} from ImageNet.}
    \label{fig:responsible_region_on_images_vgg19_features.30_mammal_mammalian_classifier_on_canine_canid}
\end{figure*}

\begin{figure*}
  \centering
  \includegraphics[width=16cm]{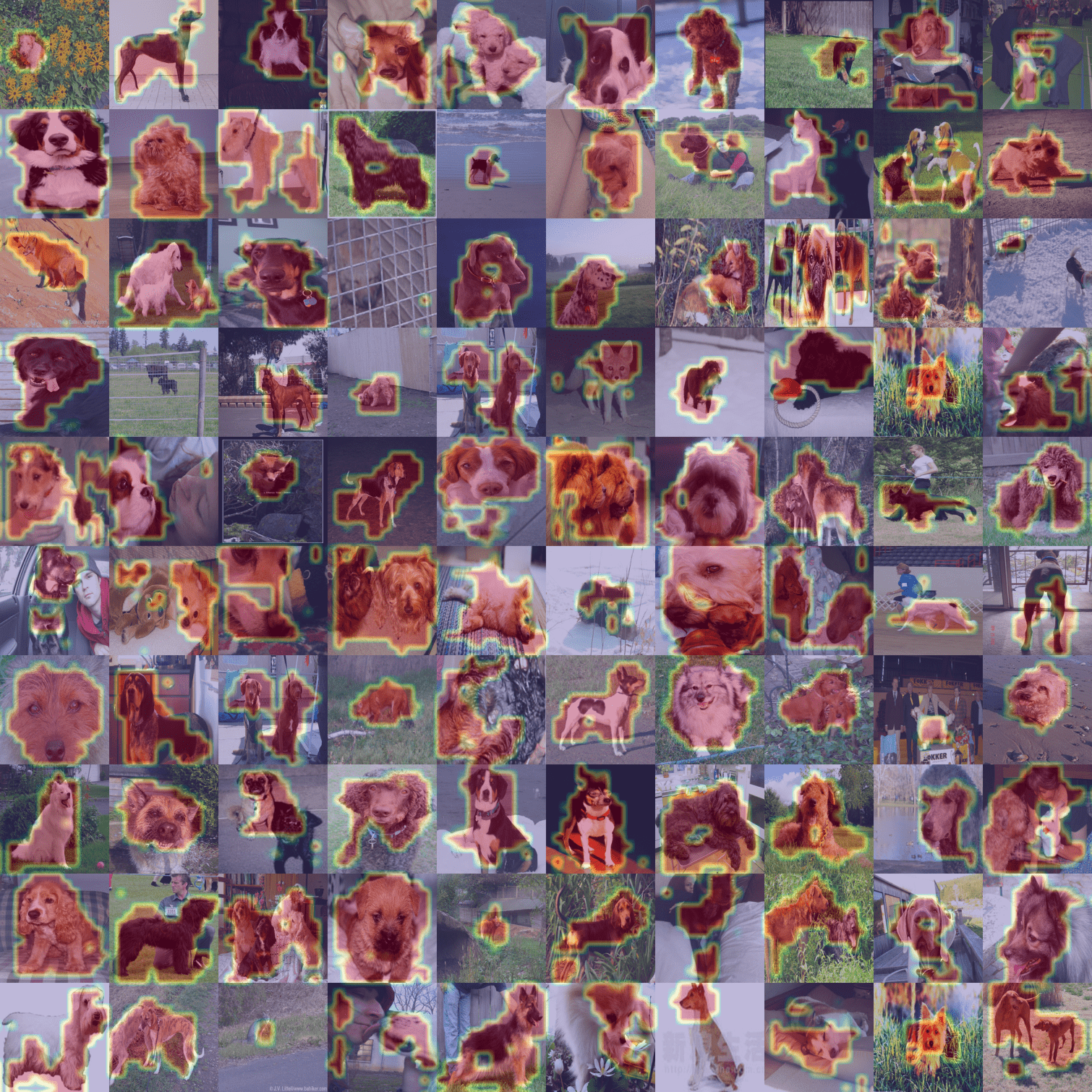}
    \caption{Localization results of applying \emph{carnivore} classifier on the images containing the concept of \emph{canine} from ImageNet.}
    \label{fig:responsible_region_on_images_vgg19_features.30_carnivore_classifier_on_canine_canid}
\end{figure*}

\paragraph{ImageNet.} Figure \ref{fig:responsible_region_on_images_vgg19_features.30_whole, unit_classifier_on_whole, unit}, \ref{fig:responsible_region_on_images_vgg19_features.30_whole, unit_classifier_on_plant, flora, plant life}, \ref{fig:responsible_region_on_images_vgg19_features.30_whole, unit_classifier_on_animal, animate being, beast, brute, creature, fauna}, \ref{fig:responsible_region_on_images_vgg19_features.30_whole, unit_classifier_on_bird}, and \ref{fig:responsible_region_on_images_vgg19_features.30_whole, unit_classifier_on_canine, canid} illustrate the localization results of applying \emph{whole} classifier on images containing contents of \emph{whole}, \emph{plant}, \emph{animal}, \emph{bird}, and \emph{canine} respectively. Figure \ref{fig:responsible_region_on_images_vgg19_features.30_mammal, mammalian_classifier_on_animal, animate being, beast, brute, creature, fauna}, \ref{fig:responsible_region_on_images_vgg19_features.30_mammal, mammalian_classifier_on_mammal, mammalian}, and \ref{fig:responsible_region_on_images_vgg19_features.30_mammal_mammalian_classifier_on_canine_canid} illustrate the localization results of applying \emph{mammal} classifier on images containing contents of \emph{animal}, \emph{mammal}, and \emph{canine} respectively. Note that some \emph{animals} are not \emph{mammals} and cannot be located. Figure \ref{fig:responsible_region_on_images_vgg19_features.30_whole, unit_classifier_on_canine, canid}, \ref{fig:responsible_region_on_images_vgg19_features.30_mammal_mammalian_classifier_on_canine_canid}, and \ref{fig:responsible_region_on_images_vgg19_features.30_carnivore_classifier_on_canine_canid} illustrate the localization results of applying \emph{whole}, \emph{mammal}, and \emph{carnivore} classifiers on images containing contents of \emph{canine} respectively.

\begin{figure*}
  \centering
  \includegraphics[width=16cm]{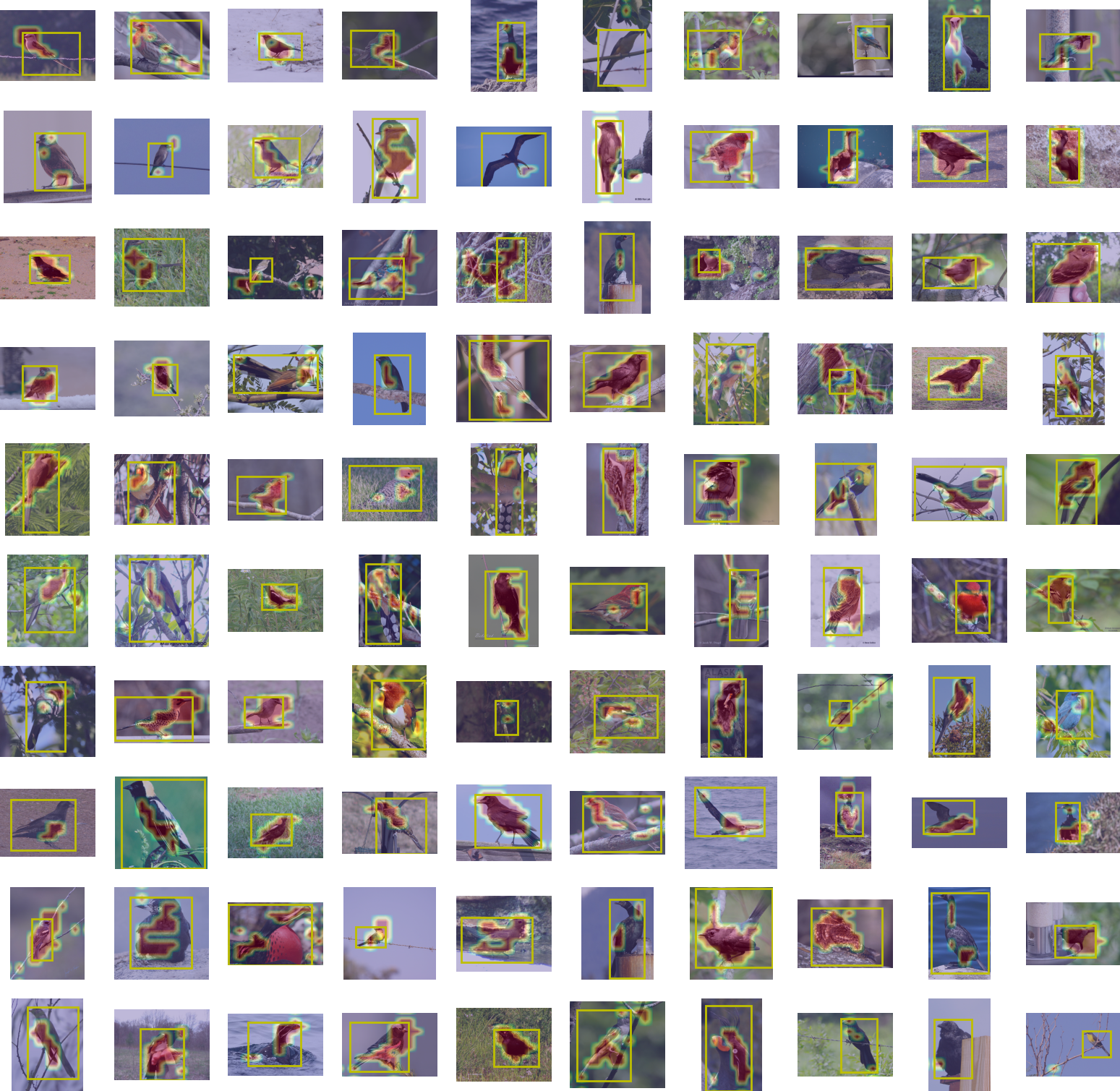}
    \caption{Localization results of applying \emph{animal} classifier trained on layer Mixed\_6b of Inception v3 on the images from CUB-200-2011. The yellow bounding boxes are the groundtruth.}
    \label{fig:Localization_CUB-200-2011_inception_v3_614_channels}
\end{figure*}

\begin{figure*}
  \centering
  \includegraphics[width=16cm]{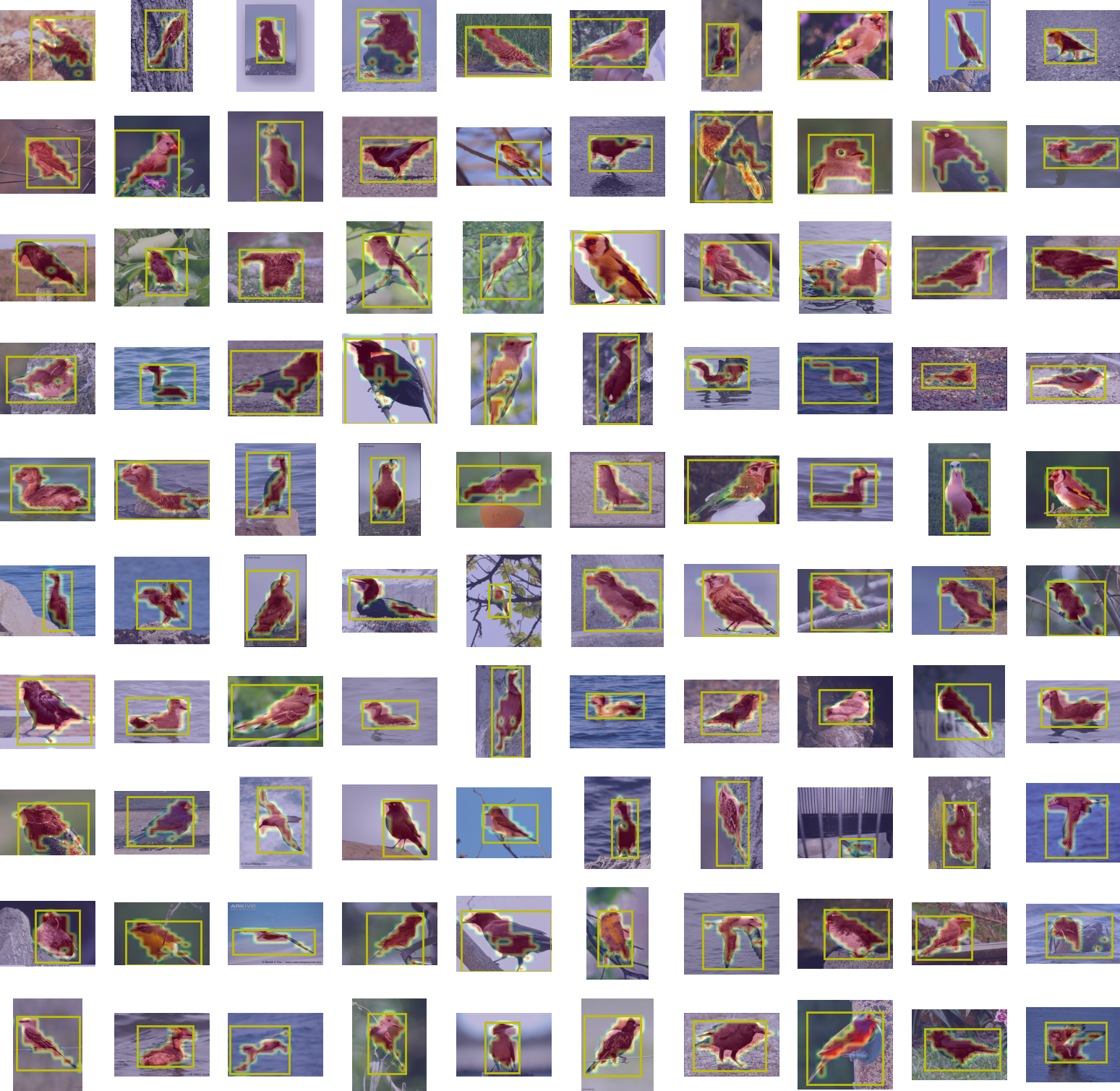}
    \caption{Localization results of applying \emph{animal} classifier trained on layer layer3.5 of ResNet50 on the images from CUB-200-2011. The yellow bounding boxes are the groundtruth.}
    \label{fig:Localization_CUB-200-2011_resnet50_with_819_channels}
\end{figure*}

\begin{figure*}
  \centering
  \includegraphics[width=16cm]{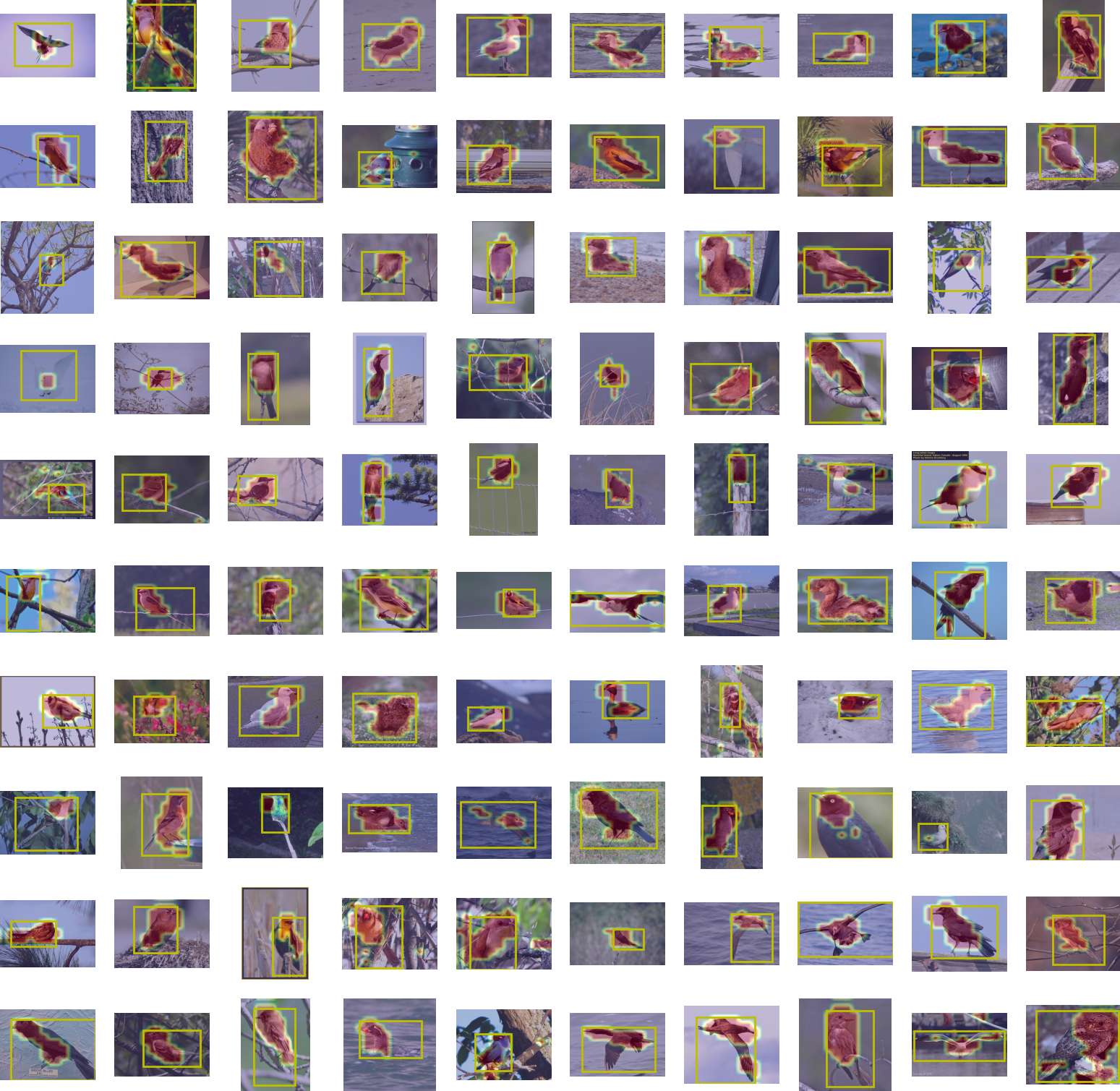}
    \caption{Localization results of applying \emph{animal} classifier trained on layer features.26 of VGG16 on the images from CUB-200-2011. The yellow bounding boxes are the groundtruth.}
    \label{fig:Localization_CUB-200-2011_vgg16_with_51_channels}
\end{figure*}

\paragraph{CUB-200-2011.} Figure \ref{fig:Localization_CUB-200-2011_inception_v3_614_channels}, \ref{fig:Localization_CUB-200-2011_resnet50_with_819_channels}, and \ref{fig:Localization_CUB-200-2011_vgg16_with_51_channels} illustrate the localization results of applying \emph{animal} classifier trained on layer Mixed\_6b of Inception v3, layer3.5 of ResNet50, and features.26 of VGG16 on the images from CUB-200-2011 respectively.

\begin{figure*}
  \centering
  \includegraphics[width=16cm]{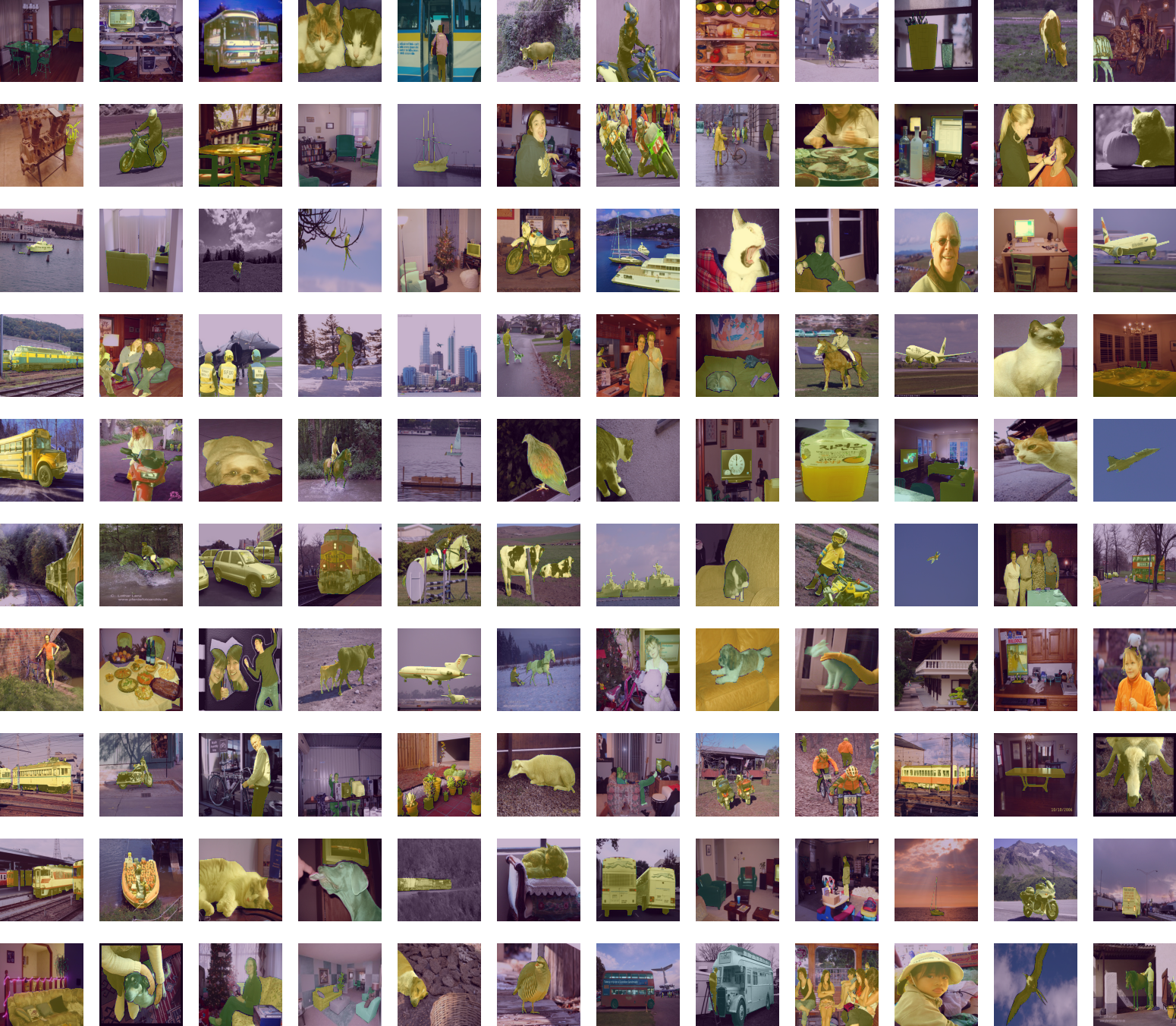}
    \caption{Sample images from PASCAL VOC with masks indicating the target objects.}
    \label{fig:Localization_PASCAL_VOC_img}
\end{figure*}

\begin{figure*}
  \centering
  \includegraphics[width=16cm]{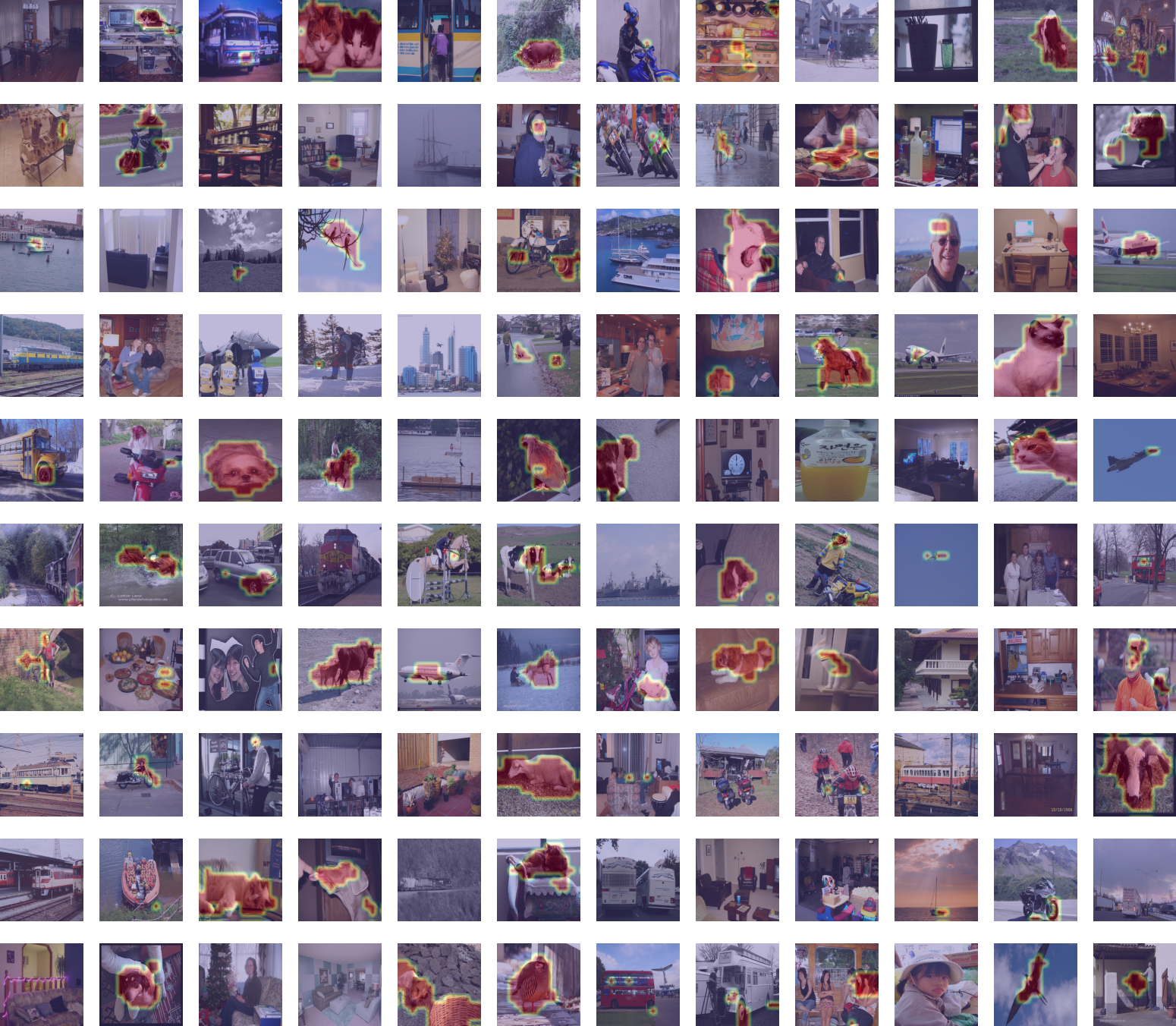}
    \caption{Localization results of applying \emph{animal} classifier on the sample images from PASCAL VOC. The classifier is trained on layer features.30 of VGG19 with 20 neurons selected by Shapley values.}
    \label{fig:Localization_PASCAL_VOC_animal}
\end{figure*}

\begin{figure*}
  \centering
  \includegraphics[width=16cm]{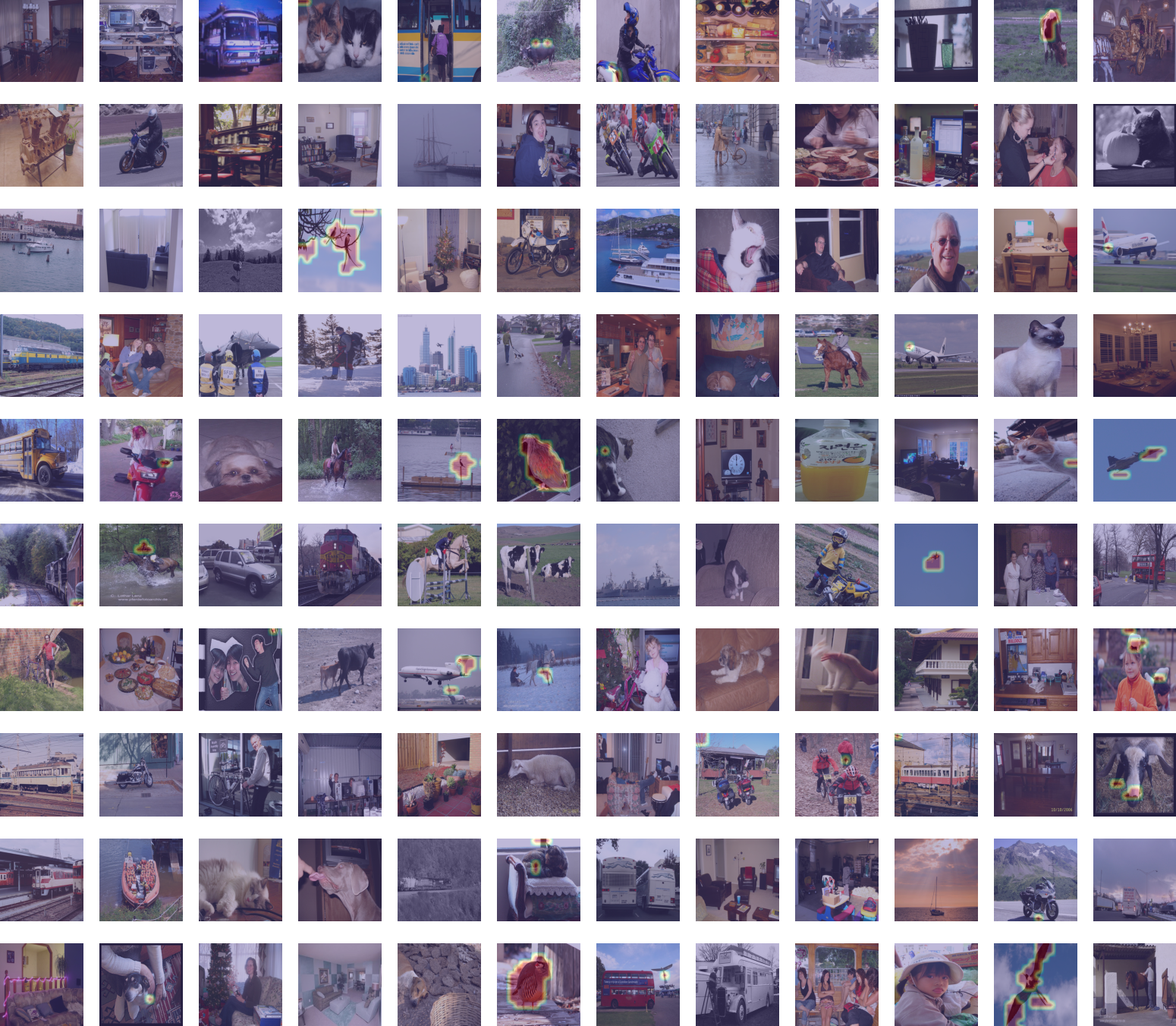}
    \caption{Localization results of applying \emph{bird} classifier on the sample images from PASCAL VOC. The classifier is trained on layer features.30 of VGG19 with 20 neurons selected by Shapley values.}
    \label{fig:Localization_PASCAL_VOC_bird}
\end{figure*}

\paragraph{PASCAL VOC.} Figure \ref{fig:Localization_PASCAL_VOC_img} shows the sample images from PASCAL VOC used for test with masks indicating the target objects. Figure \ref{fig:Localization_PASCAL_VOC_whole}, \ref{fig:Localization_PASCAL_VOC_animal}, and \ref{fig:Localization_PASCAL_VOC_bird} illustrate the localization results of applying \emph{whole}, \emph{animal}, and \emph{bird} classifiers on the sample images. The classifiers are all trained on layer features.30 of VGG19 with 20 neurons selected by Shapley values. The results indicate the unique advantage of HINT for object localization: a flexible choice of localization targets.

\section{Ablation study}
\label{sec:ablation_study}

\begin{figure*}
  \centering
  \includegraphics[width=10cm]{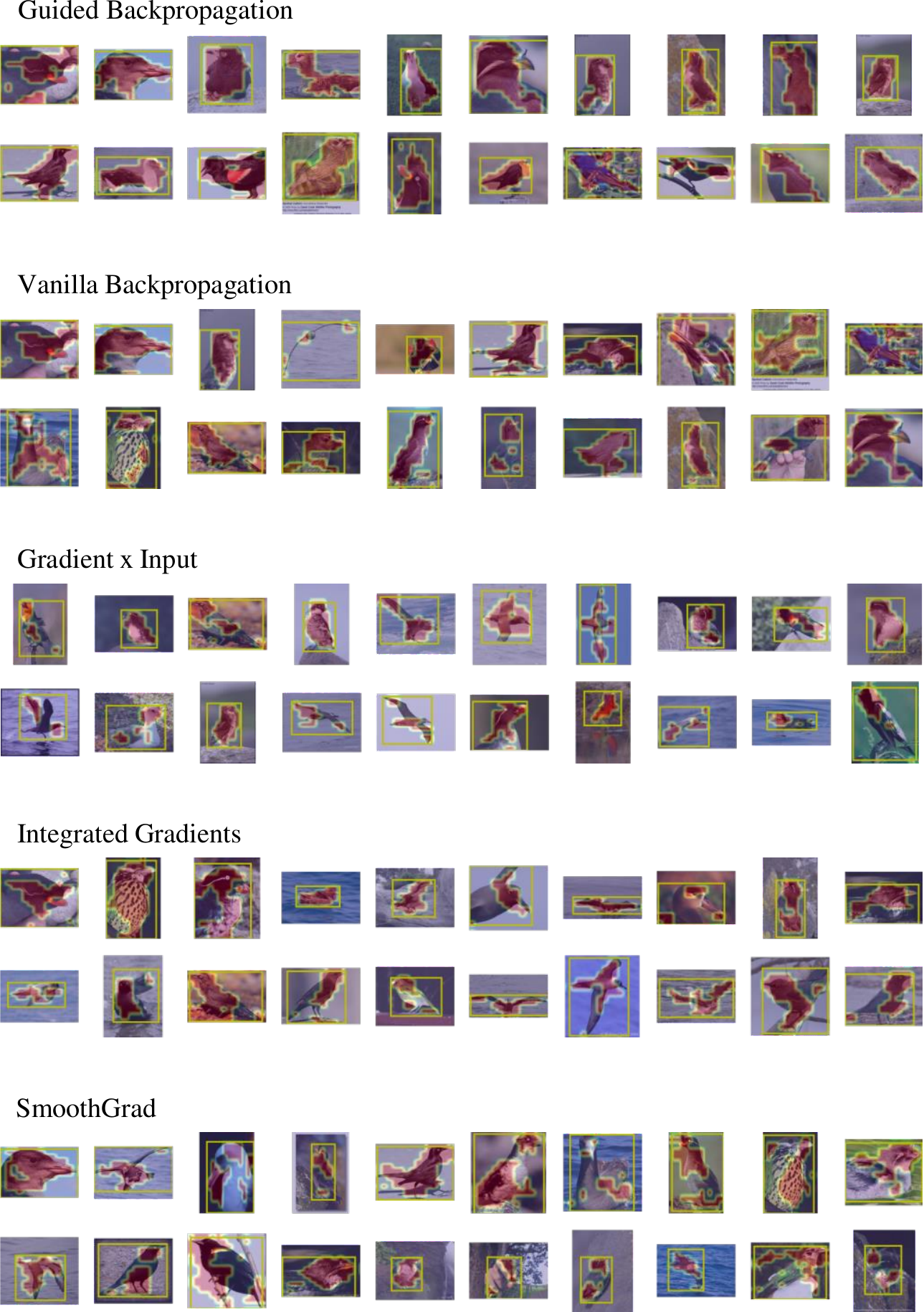}
    \caption{Localization results of \emph{animal} classifiers implemented with different modified saliency methods.}
    \label{fig:fig_dif_saliency_methods}
\end{figure*}

\paragraph{Illustration of the localization results of concept classifiers implemented with different saliency methods.} Figure \ref{fig:fig_dif_saliency_methods} shows the localization results of concept classifiers using Guided Backpropagation, Vanilla Backpropgation, Gradient x Input, Integrated Gradients, and SmoothGrad on dataset CUB-200-2011. The illustration indicates that HINT is general and can be implemented with different saliency methods.

\begin{figure*}
  \centering
  \includegraphics[width=16cm]{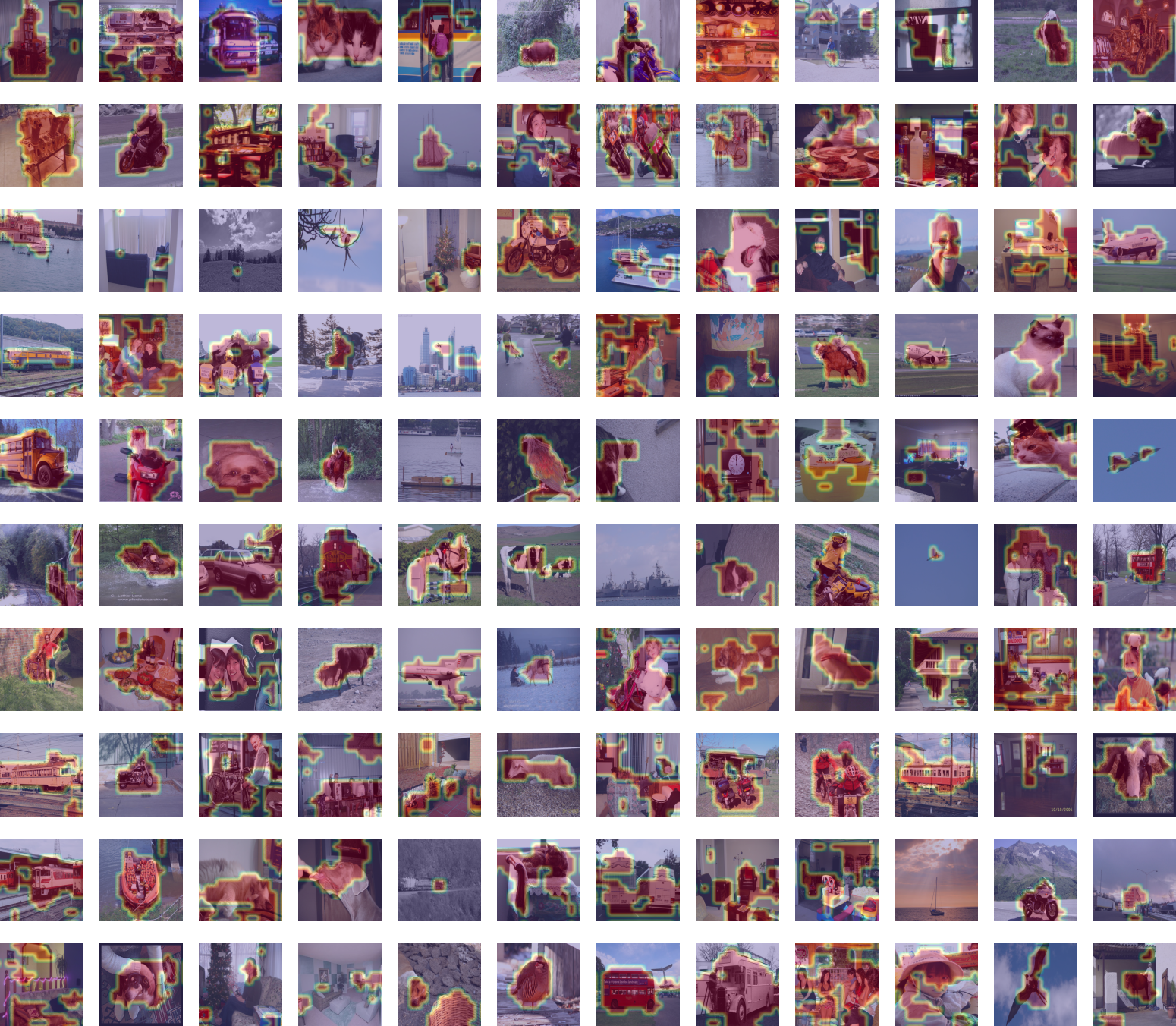}
    \caption{Localization results of applying \emph{whole} classifier on the sample images from PASCAL VOC. The classifier is trained on layer features.30 of VGG19 with 20 neurons selected by Shapley values.}
    \label{fig:Localization_PASCAL_VOC_whole}
\end{figure*}

\begin{figure*}
  \centering
  \includegraphics[width=16cm]{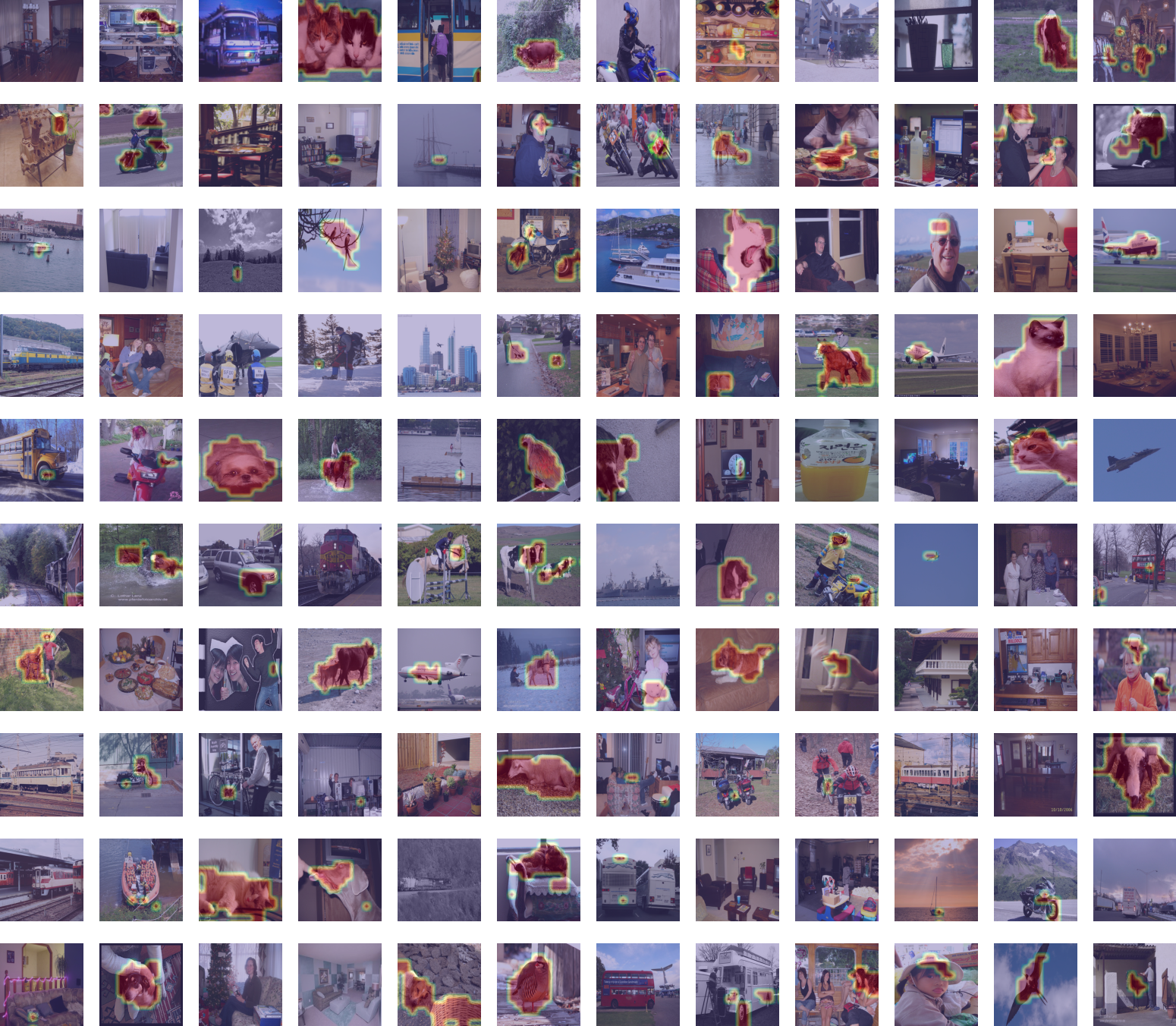}
    \caption{Localization results of applying \emph{whole} classifier on the sample images from PASCAL VOC. The classifier is trained on layer features.30 of VGG19 with 20 neurons selected by the coefficients of the linear classifier.}
    \label{fig:Localization_PASCAL_VOC_whole_clf_coef}
\end{figure*}

\begin{figure*}
  \centering
  \includegraphics[width=16cm]{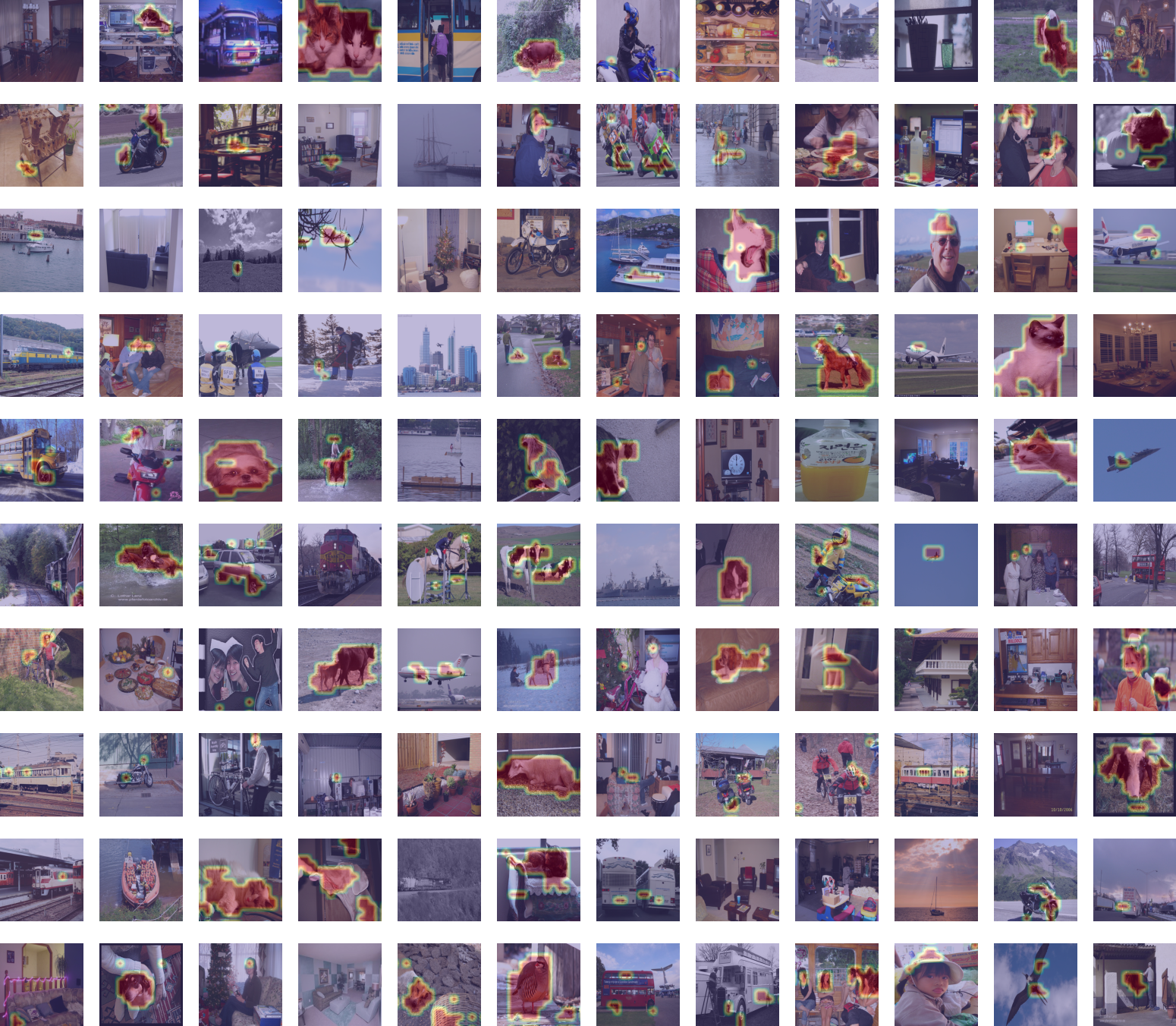}
    \caption{Localization results of applying \emph{whole} classifier on the sample images from PASCAL VOC. The classifier is trained on layer features.30 of VGG19 with 20 neurons randomly selected.}
    \label{fig:Localization_PASCAL_VOC_whole_random}
\end{figure*}

\paragraph{Illustration of the localization results of concept classifiers trained with neurons chosen by shap, clf\_coef, and random} Figure \ref{fig:Localization_PASCAL_VOC_whole}, \ref{fig:Localization_PASCAL_VOC_whole_clf_coef}, and \ref{fig:Localization_PASCAL_VOC_whole_random} show the localization results of applying \emph{whole} classifiers on the sample images from PASCAL VOC, where the classifiers are trained on layer features.30 of VGG19 with 20 neurons selected by Shapley values (shap), selected by the coefficients of the linear classifier (clf\_coef), and randomly selected (random) respectively. From observation, "shap" locates more \emph{whole} objects and larger object contents, indicating that Shapley values are good measures of neurons' contributions to concepts.

\clearpage

\begin{figure*}
  \centering
  \includegraphics[width=16cm]{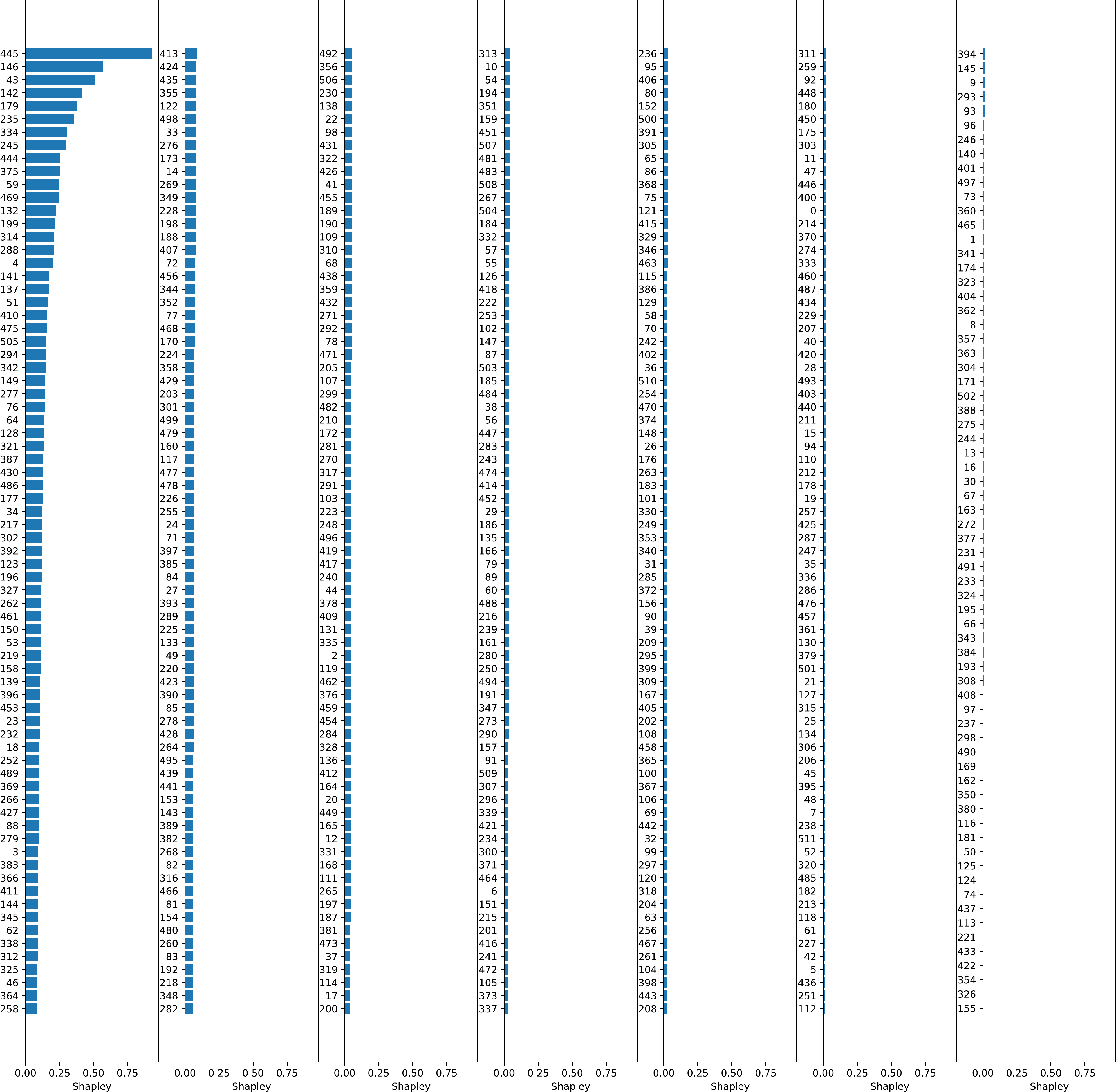}
    \caption{Contribution scores (Shapley values) of neurons on layer features.30 of VGG19 to the concept of \emph{animal}.}
    \label{fig:shap_bar_plot_vgg19_features.30_animal, animate being, beast, brute, creature, fauna_classifier}
\end{figure*}

\begin{figure*}
  \centering
  \includegraphics[width=16cm]{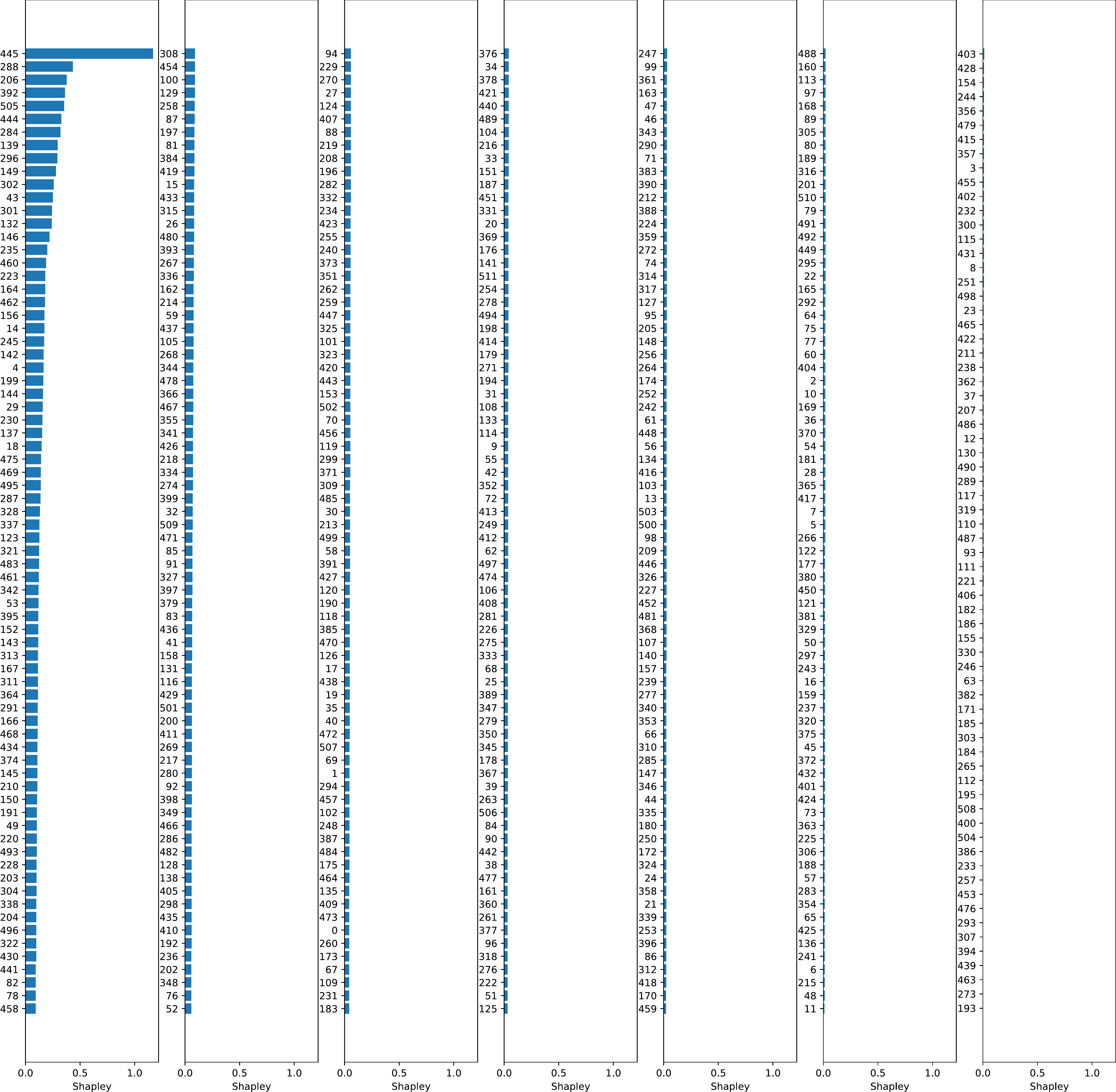}
    \caption{Contribution scores (Shapley values) of neurons on layer features.30 of VGG19 to the concept of \emph{vertebrate}.}
    \label{fig:shap_bar_plot_vgg19_features.30_vertebrate, craniate_classifier}
\end{figure*}

\begin{figure*}
  \centering
  \includegraphics[width=16cm]{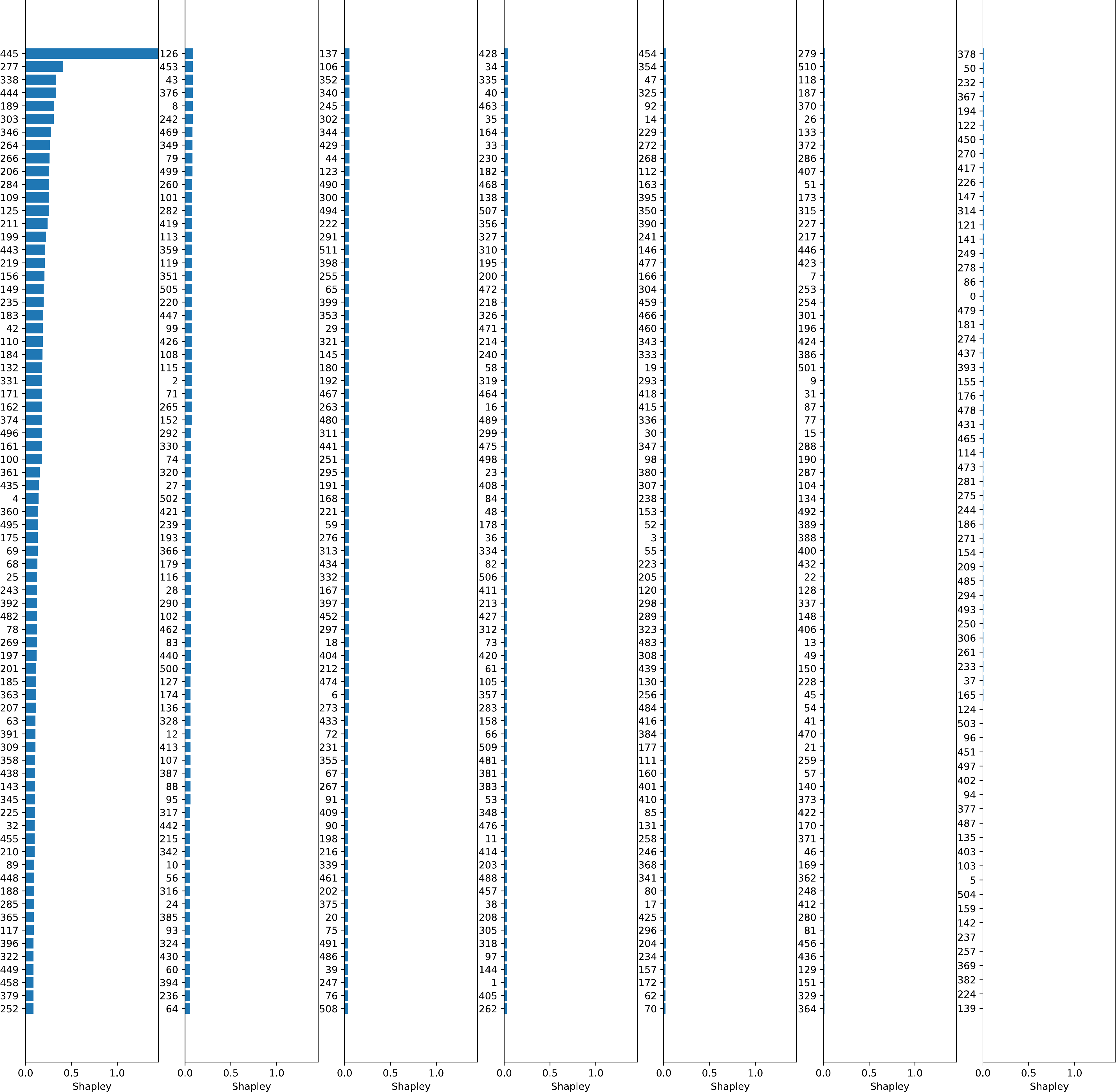}
    \caption{Contribution scores (Shapley values) of neurons on layer features.30 of VGG19 to the concept of \emph{mammal}.}
    \label{fig:shap_bar_plot_vgg19_features.30_mammal, mammalian_classifier}
\end{figure*}

\begin{figure*}
  \centering
  \includegraphics[width=16cm]{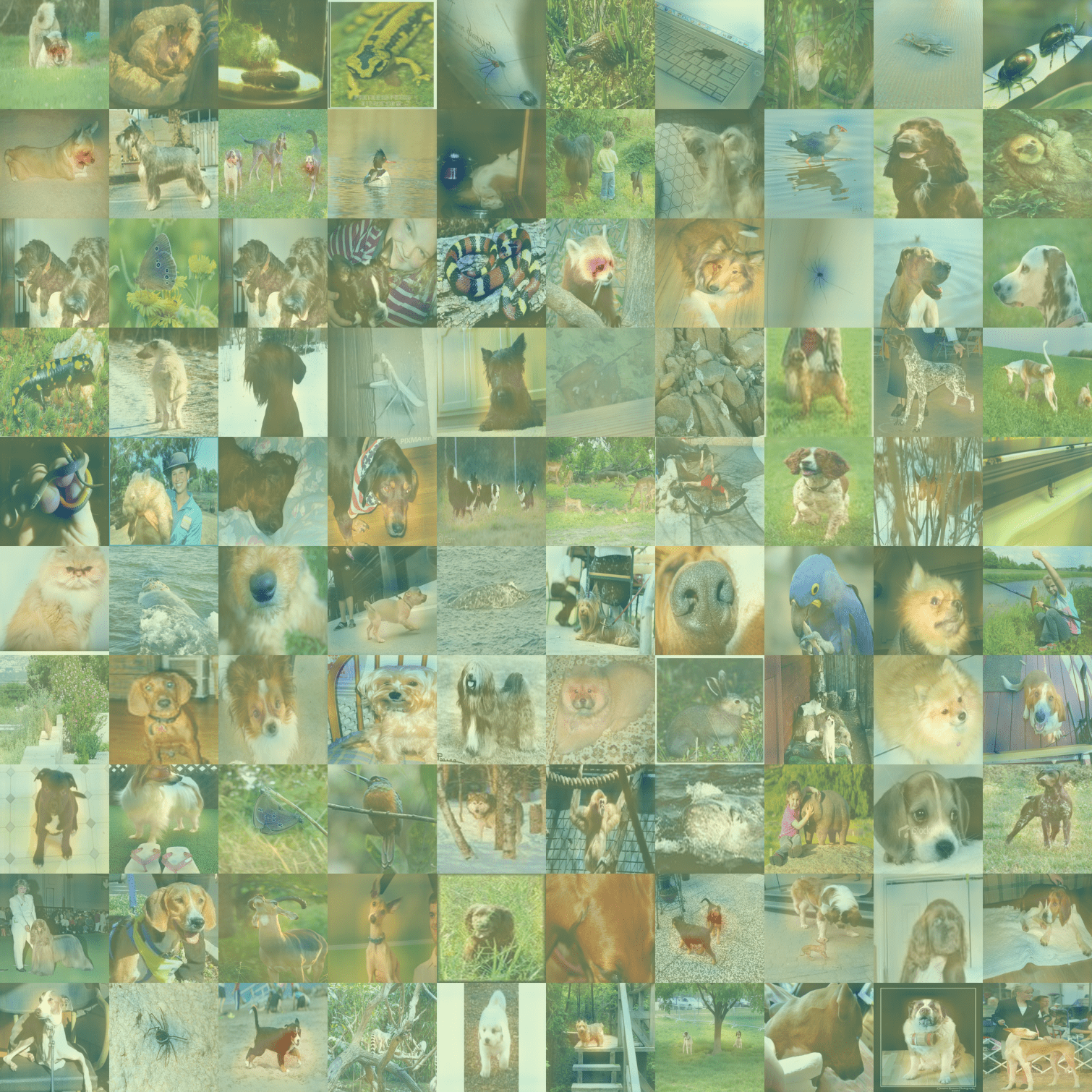}
    \caption{Activation map of the $445^{th}$ neuron on \emph{animal} images.}
    \label{fig:channel_activation_445th_channel_images_vgg19_animal, animate being, beast, brute, creature, fauna}
\end{figure*}

\begin{figure*}
  \centering
  \includegraphics[width=16cm]{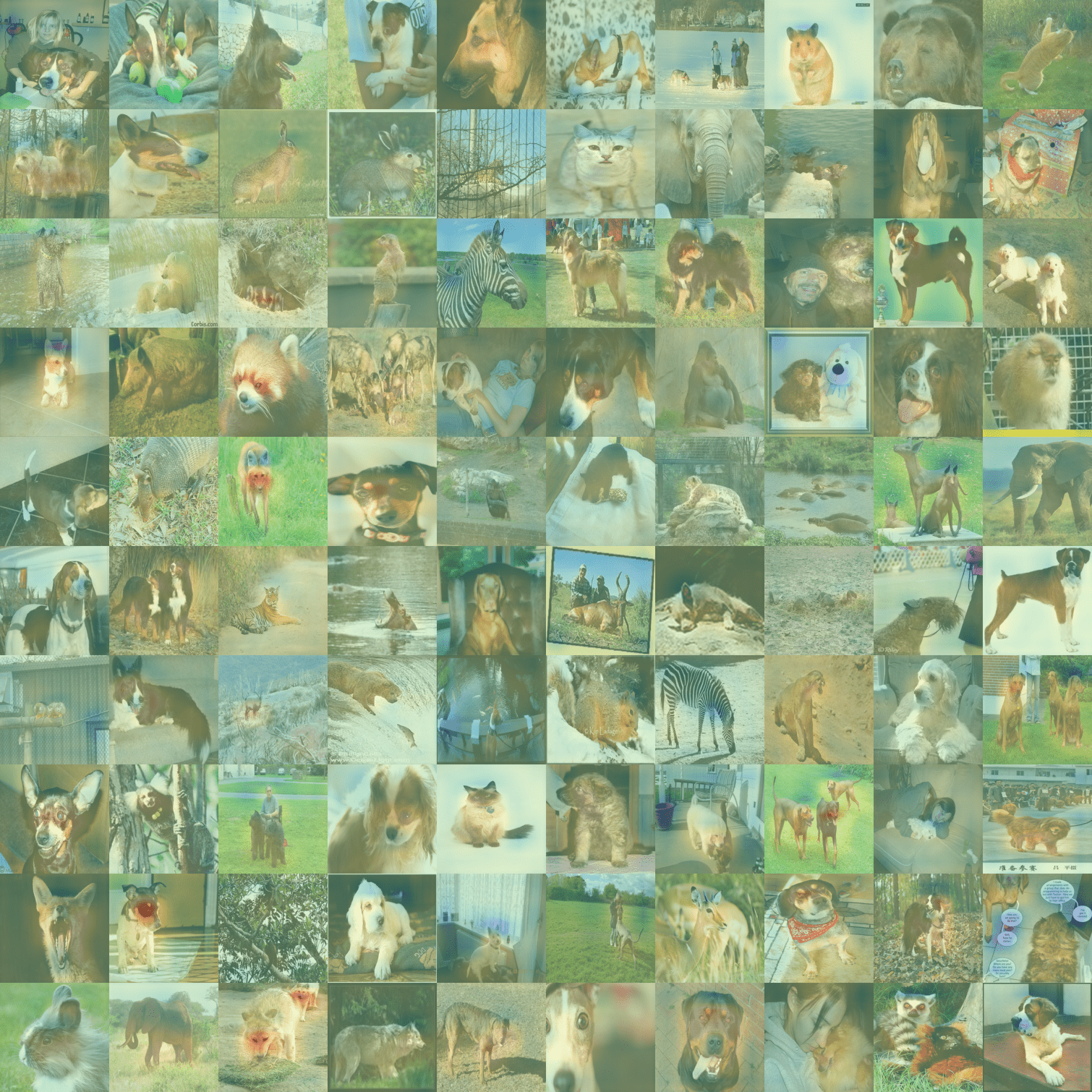}
    \caption{Activation map of the $445^{th}$ neuron on \emph{mammal} images.}
    \label{fig:channel_activation_445th_channel_images_vgg19_mammal, mammalian}
\end{figure*}

\begin{figure*}
  \centering
  \includegraphics[width=16cm]{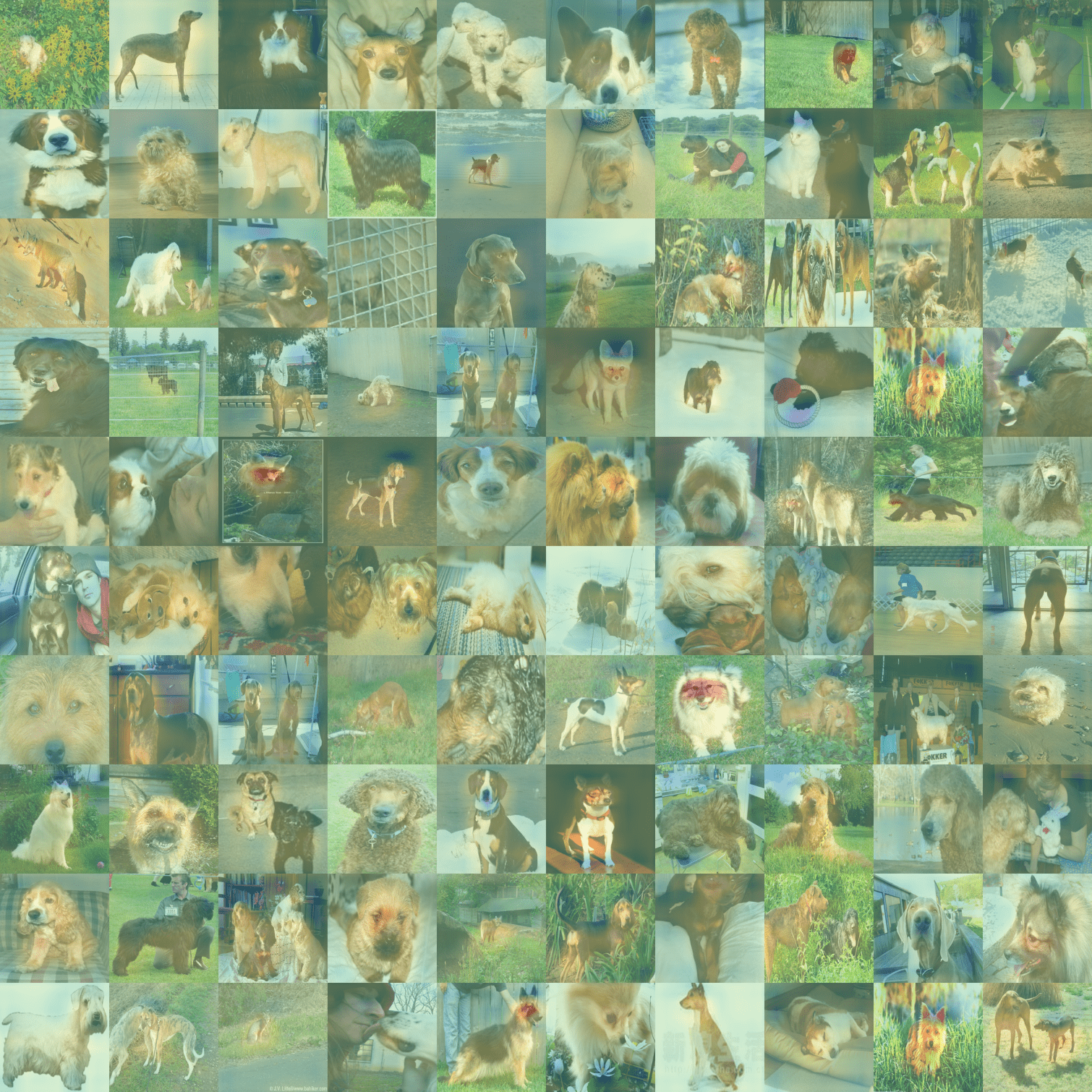}
    \caption{Activation map of the $445^{th}$ neuron on \emph{canine} images.}
    \label{fig:channel_activation_445th_channel_images_vgg19_canine, canid}
\end{figure*}

\begin{figure*}
  \centering
  \includegraphics[width=16cm]{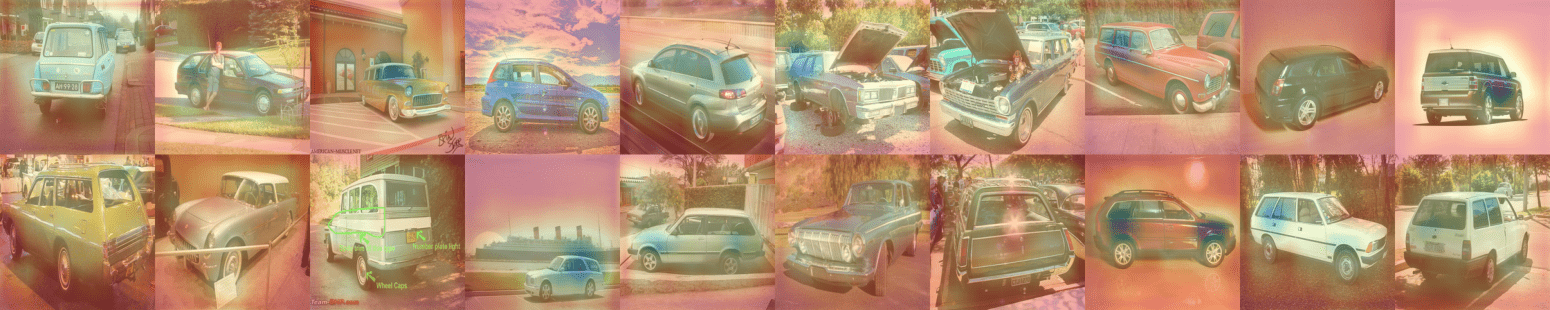}
    \caption{Activation map of the $199^{th}$ neuron on \emph{car} images.}
    \label{fig:channel_activation_199th_channel_images_vgg19_beach wagon, station wagon, wagon, estate car, beach waggon, station waggon, waggon}
\end{figure*}

\begin{figure*}
  \centering
  \includegraphics[width=16cm]{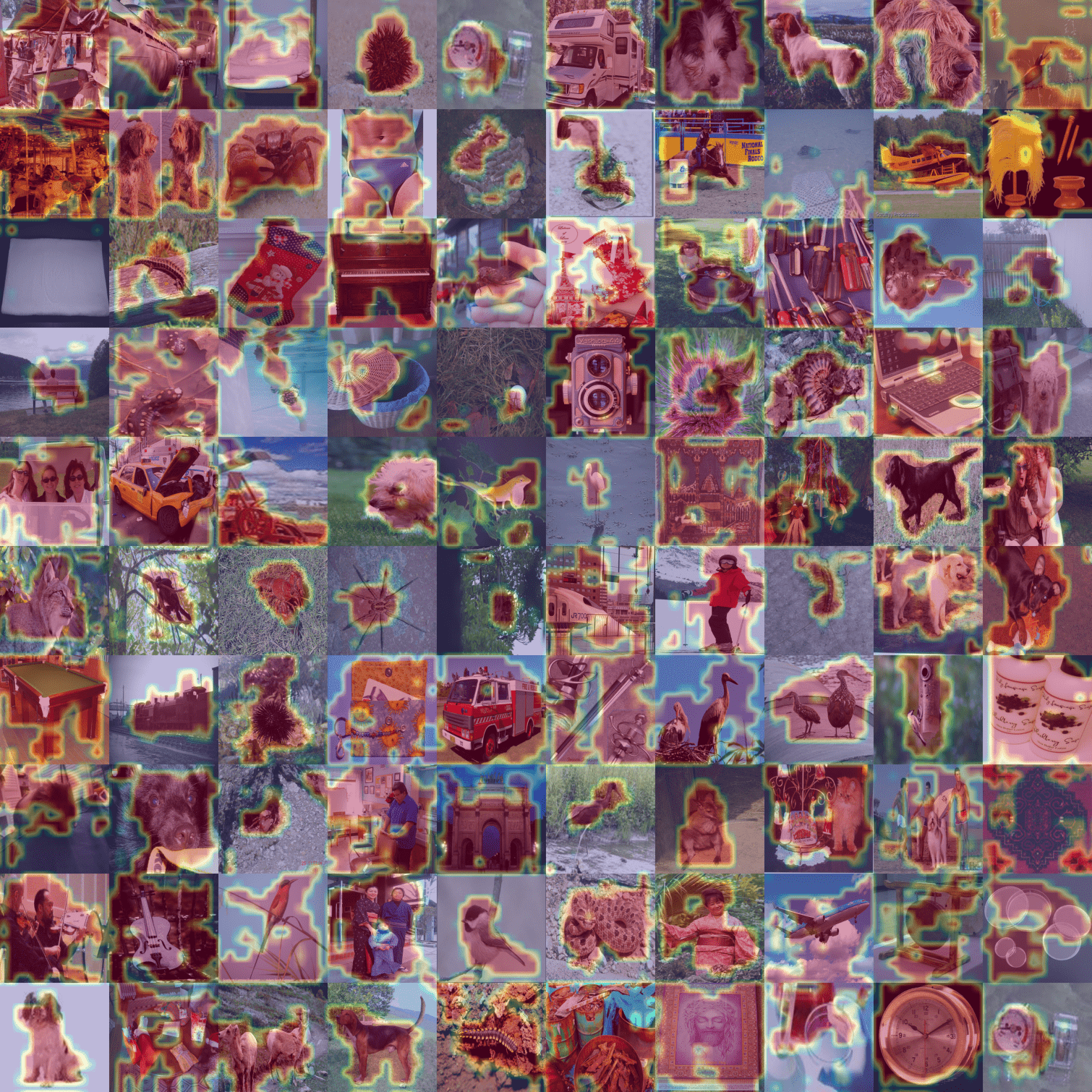}
    \caption{Localization results of applying \emph{whole} classifier on the images containing the concept of \emph{whole} from ImageNet.}
    \label{fig:responsible_region_on_images_vgg19_features.30_whole, unit_classifier_on_whole, unit}
\end{figure*}

\begin{figure*}
  \centering
  \includegraphics[width=16cm]{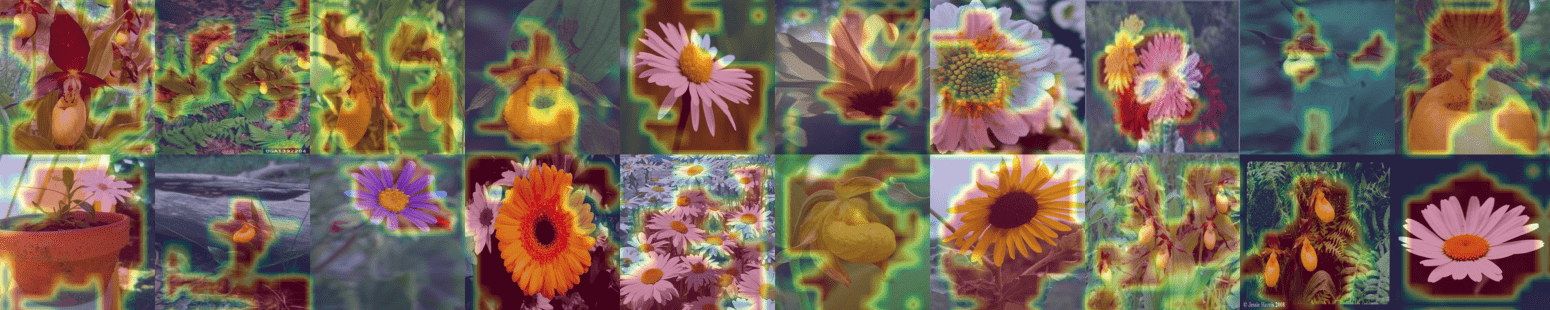}
    \caption{Localization results of applying \emph{whole} classifier on the images containing the concept of \emph{plant} from ImageNet.}
    \label{fig:responsible_region_on_images_vgg19_features.30_whole, unit_classifier_on_plant, flora, plant life}
\end{figure*}

\begin{figure*}
  \centering
  \includegraphics[width=16cm]{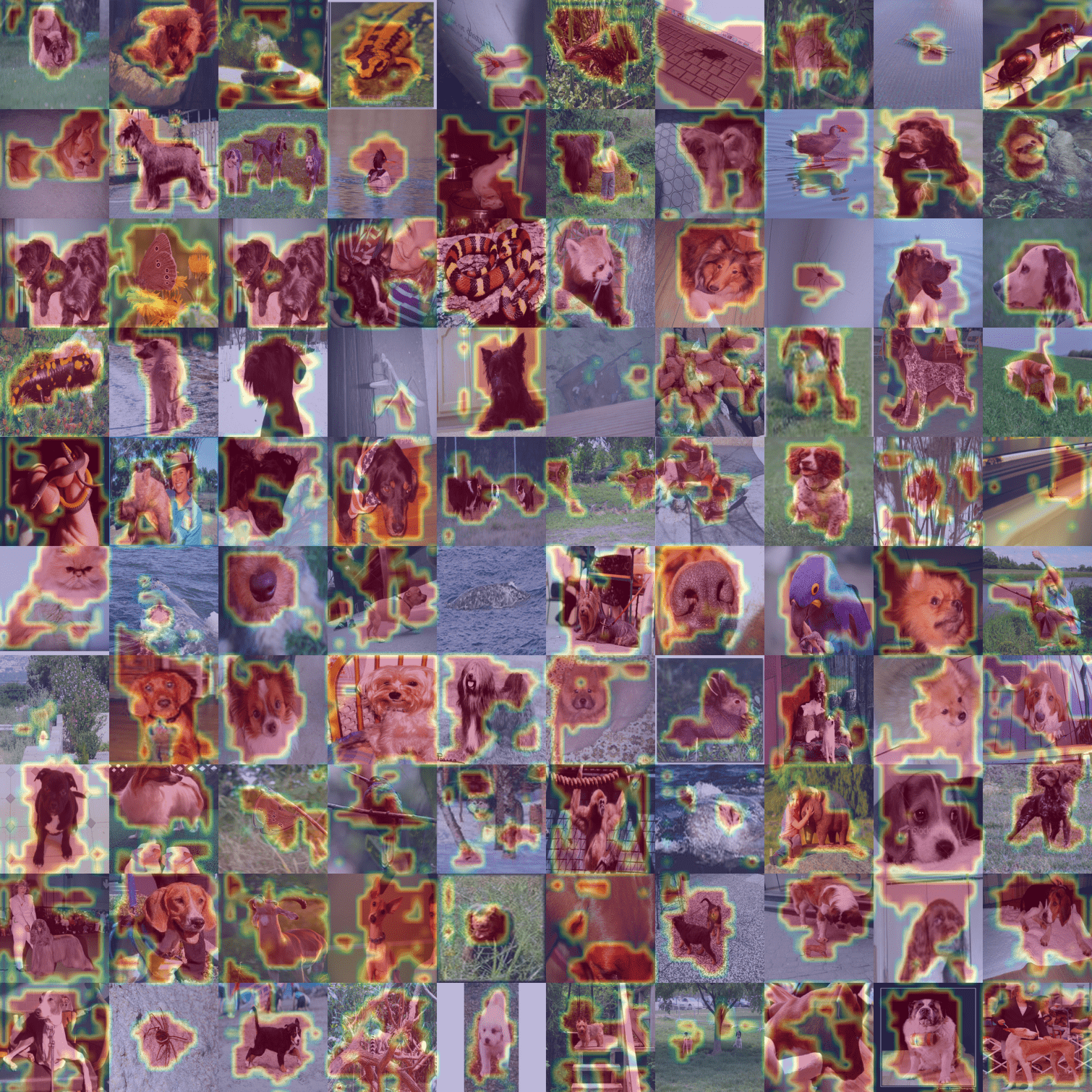}
    \caption{Localization results of applying \emph{whole} classifier on the images containing the concept of \emph{animal} from ImageNet.}
    \label{fig:responsible_region_on_images_vgg19_features.30_whole, unit_classifier_on_animal, animate being, beast, brute, creature, fauna}
\end{figure*}

\begin{figure*}
  \centering
  \includegraphics[width=16cm]{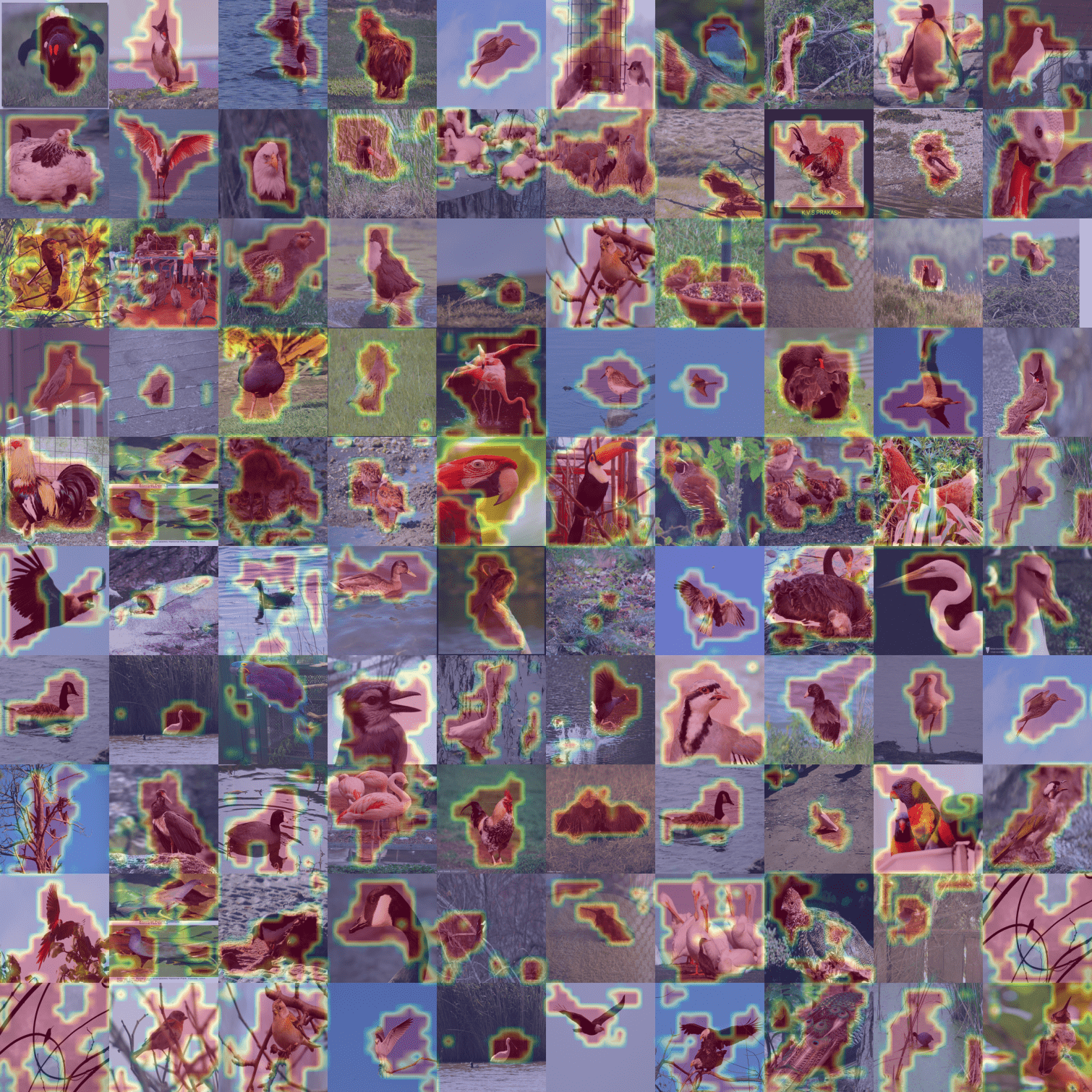}
    \caption{Localization results of applying \emph{whole} classifier on the images containing the concept of \emph{bird} from ImageNet.}
    \label{fig:responsible_region_on_images_vgg19_features.30_whole, unit_classifier_on_bird}
\end{figure*}

\begin{figure*}
  \centering
  \includegraphics[width=16cm]{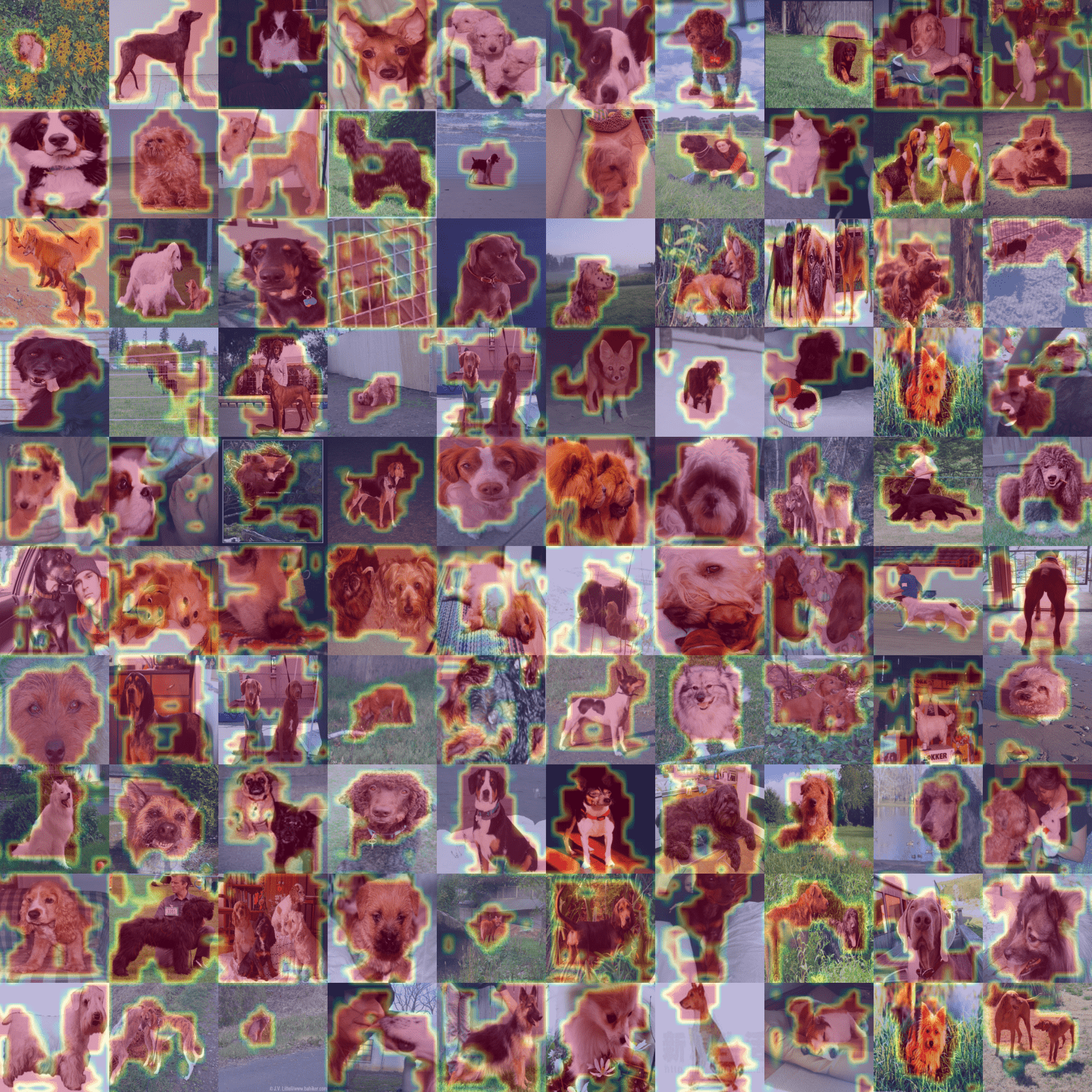}
    \caption{Localization results of applying \emph{whole} classifier on the images containing the concept of \emph{canine} from ImageNet.}
    \label{fig:responsible_region_on_images_vgg19_features.30_whole, unit_classifier_on_canine, canid}
\end{figure*}

\begin{figure*}
  \centering
  \includegraphics[width=16cm]{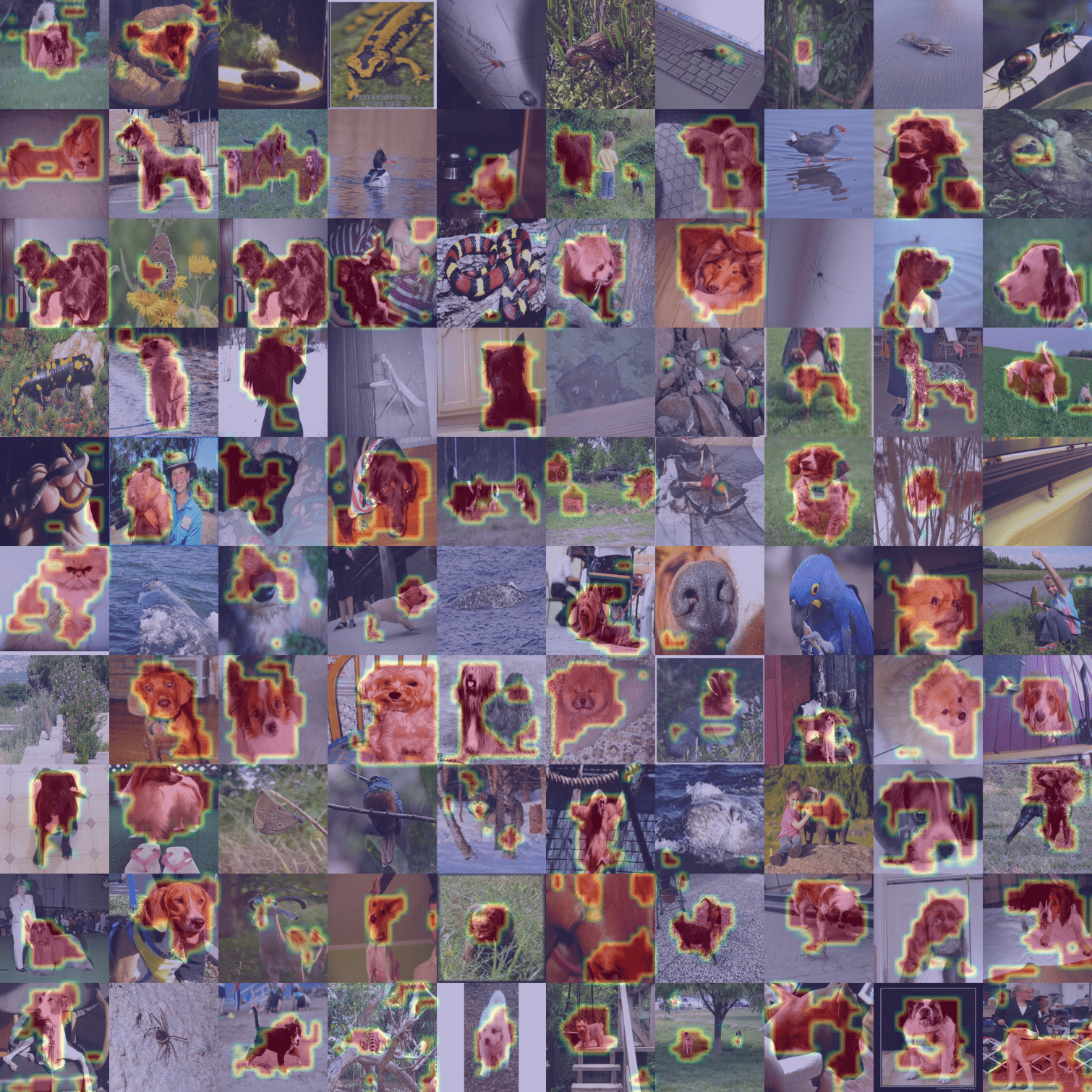}
    \caption{Localization results of applying \emph{mammal} classifier on the images containing the concept of \emph{animal} from ImageNet. Note that some \emph{animals} are not \emph{mammals} and cannot be located.}
    \label{fig:responsible_region_on_images_vgg19_features.30_mammal, mammalian_classifier_on_animal, animate being, beast, brute, creature, fauna}
\end{figure*}

\begin{figure*}
  \centering
  \includegraphics[width=16cm]{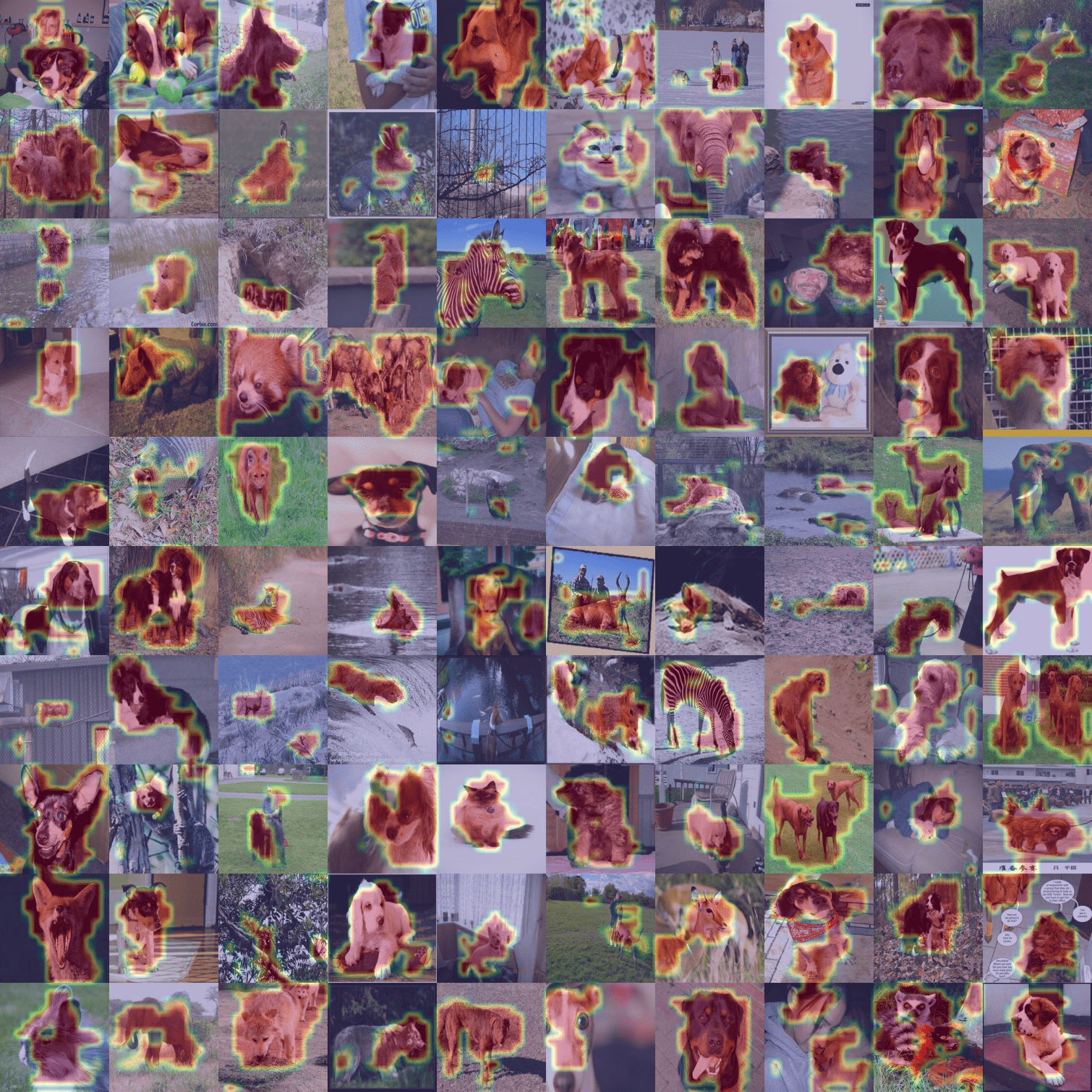}
    \caption{Localization results of applying \emph{mammal} classifier on the images containing the concept of \emph{mammal} from ImageNet.}
    \label{fig:responsible_region_on_images_vgg19_features.30_mammal, mammalian_classifier_on_mammal, mammalian}
\end{figure*}

\clearpage


\end{document}